\documentclass[rgb,table]{taotian}

\usepackage[utf8]{inputenc} 
\usepackage[T1]{fontenc} 
\usepackage{microtype} 

\usepackage[toc,page,header]{appendix}
\usepackage{multicol}
\usepackage{wrapfig}
\usepackage{lipsum}

\usepackage{amsmath}
\usepackage{mathtools}
\usepackage{amsthm}
\usepackage{amsfonts} 
\usepackage{amssymb}
\usepackage{bm}
\usepackage{bbm}
\usepackage{nicefrac}
\usepackage{textcomp}
\usepackage{bbding}
\usepackage{pifont}
\usepackage{xspace}
\usepackage{gradient-text}

\usepackage{array} 
\usepackage{booktabs}
\usepackage{multirow}
\usepackage{makecell}
\usepackage{colortbl}
\usepackage{tabularx}
\usepackage{xltabular}

\usepackage{graphicx}
\usepackage{caption}
\usepackage{subcaption}
\usepackage{adjustbox}
\usepackage[breakable]{tcolorbox}

\usepackage{enumitem}

\usepackage{hyperref} 
\usepackage{url}

\definecolor{myyellow}{RGB}{255,192,0}
\definecolor{mygreen}{RGB}{107,170,64}
\definecolor{mywrite}{RGB}{255,227,132}

\usepackage{booktabs} 
\usepackage{multirow}
\usepackage{amssymb}
\usepackage{gradient-text}
\usepackage{subcaption}
\usepackage{pifont}
\usepackage[breakable]{tcolorbox}
\usepackage{array}
\usepackage{xltabular}
\usepackage{amsmath}

\usepackage{pifont}

\newcommand{\methodgradient}{\gradientRGB{DeepPHY:}{255,100,0}{255,200,50}}
\newcommand\cmark{\textcolor[RGB]{13, 120, 24}{\ding{52}}}
\newcommand\doublecmark{\textcolor[RGB]{0,150,255}{\ding{80}}}
\newcommand\xmark{\textcolor[RGB]{173, 21, 14}{\ding{55}}}
\newcommand\goldstar{\textcolor[RGB]{204, 153, 51}{\ding{72}}}

\newcommand\pooltoollogo{\raisebox{-0.5ex}{\includegraphics[height=3ex]{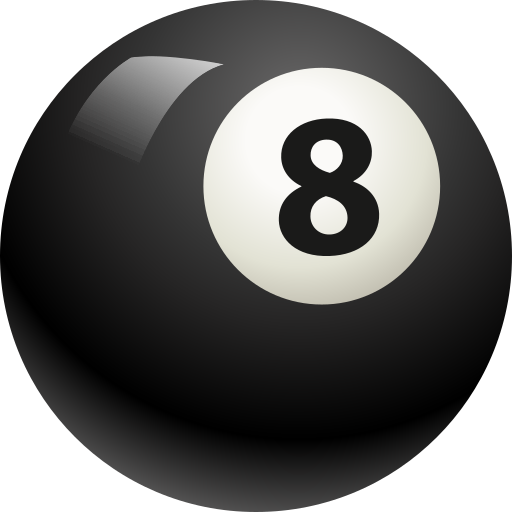}} \textbf{Pooltool}}
\newcommand\angrybirdlogo{\raisebox{-0.5ex}{\includegraphics[height=3ex]{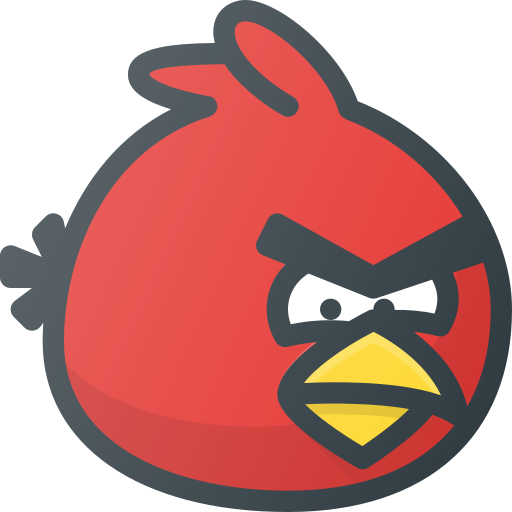}} \textbf{Angry Birds}}
\newcommand\cutrepologo{\raisebox{-0.5ex}{\includegraphics[height=3ex]{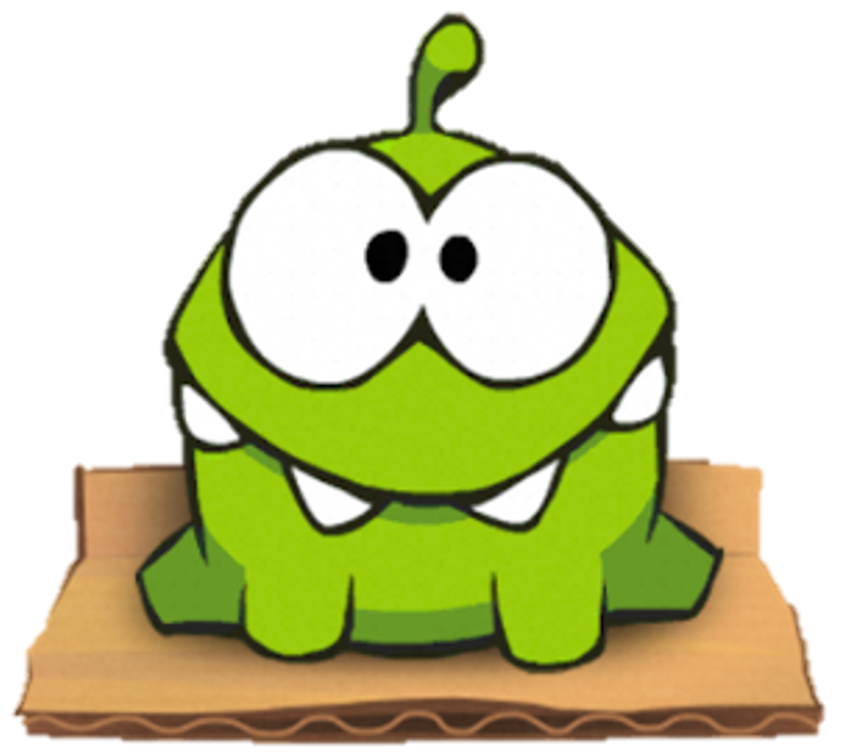}} \textbf{Cut the Rope}}
\newcommand\firstlogo{\raisebox{-0.5ex}{\includegraphics[height=2.5ex]{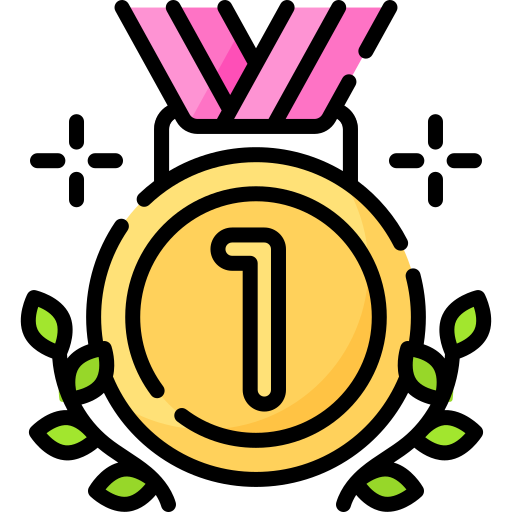}}}
\newcommand\secondlogo{\raisebox{-0.5ex}{\includegraphics[height=2.5ex]{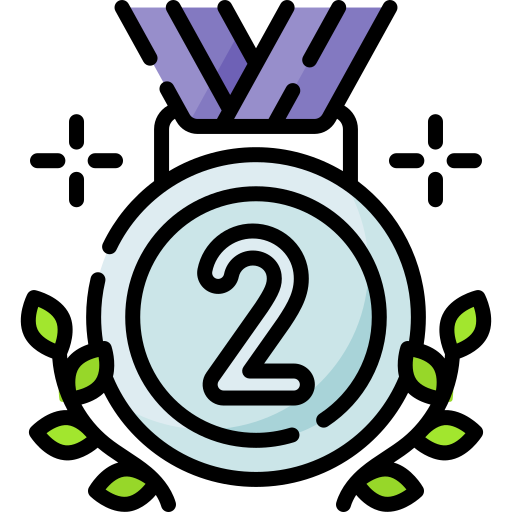}}}

\definecolor{lightblue}{HTML}{C7E0EB}
\definecolor{midblue}{HTML}{7E9CD2}
\definecolor{darkblue}{HTML}{82B9D3}
\definecolor{lightred}{HTML}{F0C8C6}
\definecolor{midred}{HTML}{EAB0AB}
\definecolor{darkred}{HTML}{DB776F}
\definecolor{lightyellow}{HTML}{FDEEC9}
\definecolor{midyellow}{HTML}{EBCB8B}
\definecolor{darkyellow}{HTML}{B5830F}

\definecolor{promptBlueBack}{HTML}{EBF5FF}
\definecolor{promptBlueFrame}{HTML}{003366}
\definecolor{promptGreenBack}{HTML}{F0FFF0}
\definecolor{promptGreenFrame}{HTML}{004D00}
\definecolor{promptOrangeBack}{HTML}{FFF5E6}
\definecolor{promptOrangeFrame}{HTML}{D35400} 
\definecolor{mylightgray}{gray}{0.9}

\newtcolorbox{systempromptbox}[1]{
    colback=promptBlueBack,   
    colframe=promptBlueFrame, 
    title=#1,                 
    fonttitle=\bfseries, 
    boxrule=0.5mm, 
    arc=2mm, 
    left=2mm,
    right=2mm, 
    top=2mm,
    bottom=2mm,
    breakable,
}

\newtcolorbox{userpromptbox}[1]{
    colback=promptGreenBack,    
    colframe=promptGreenFrame,  
    title=#1,             
    fonttitle=\bfseries, 
    boxrule=0.5mm, 
    arc=2mm, 
    left=2mm,
    right=2mm, 
    top=2mm,
    bottom=2mm,
    breakable,
}

\newtcolorbox{examplebox}[1]{
    colback=promptOrangeBack,    
    colframe=promptOrangeFrame,  
    title=#1,             
    fonttitle=\bfseries, 
    boxrule=0.5mm, 
    arc=2mm, 
    left=2mm,
    right=2mm, 
    top=2mm,
    bottom=2mm,
    breakable,
}

\definecolor{BlockC}{gray}{0.98}  
\definecolor{BlockA}{RGB}{191,211,230}
\definecolor{BlockB}{RGB}{199,233,192}

\title{\methodgradient\ Benchmarking Agentic VLMs on Physical Reasoning}

\author{{\bfseries 
Xinrun Xu$^{1,2,3}$, Pi Bu$^{1}$, Ye Wang$^{4}$, Börje F. Karlsson$^{5}$, Ziming Wang$^{1}$, Tengtao Song$^{1}$, Qi Zhu$^{1}$, Jun Song$^{1,*}$, Zhiming Ding$^{2,*}$, Bo Zheng$^{1}$}}

\affiliation{
$^{1}$Taobao \& Tmall Group of Alibaba,\\
$^{2}$Institute of Software, Chinese Academy of Science,
$^{3}$University of Chinese Academy of Sciences,\\
$^{4}$Renmin University of China,
$^{5}$Informatics Department, PUC-Rio
}

\webpage{\url{https://github.com/XinrunXu/DeepPHY}}

\firstfig[width=\linewidth][\textwidth]{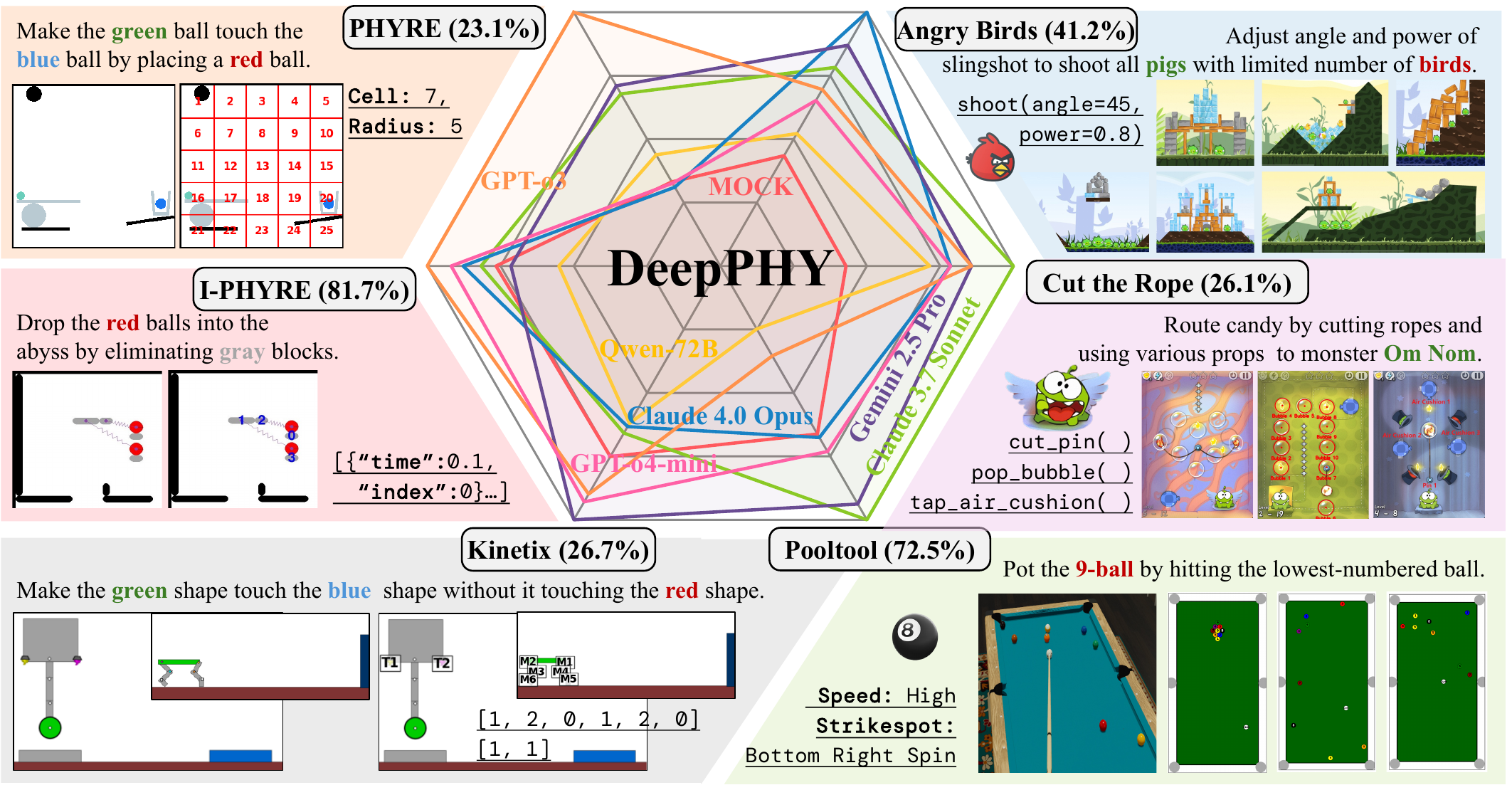}
{\textbf{DeepPHY Benchmark Suite.} Six diverse and challenging environments for evaluating interactive physical reasoning in agentic VLMs. For each environment, a representative task and its goal are displayed. The values in parentheses show the success rate of the best performing VLMs (VLA Prompt Format), and the \underline{underlined text} provides examples of the structured action spaces. These results indicate that even state-of-the-art models have significant performance gaps to address.}
{fig:001}

\abstract{
Although Vision Language Models (VLMs) exhibit strong perceptual abilities and impressive visual reasoning, they struggle with attention to detail and precise action planning in complex, dynamic environments, leading to subpar performance.
Real-world tasks typically require complex interactions, advanced spatial reasoning, long-term planning, and continuous strategy refinement, usually necessitating understanding the physics rules of the target scenario. 
However, evaluating these capabilities in real-world scenarios is often prohibitively expensive.
To bridge this gap, we introduce DeepPHY, a novel benchmark framework designed to systematically evaluate VLMs' understanding and reasoning about fundamental physical principles through a series of challenging simulated environments. 
DeepPHY integrates multiple physical reasoning environments of varying difficulty levels and incorporates fine-grained evaluation metrics. 
Our evaluation finds that even state-of-the-art VLMs struggle to translate descriptive physical knowledge into precise, predictive control.
}

\begingroup
 
\footnotetext[1]{Corresponding Authors.}
\endgroup

\begin{document}

\maketitle

\section{Introduction} \label{sec:intro}

Vision Language Models (VLMs) have demonstrated remarkable results in both static visual content understanding tasks \cite{ye2024mplugowl3longimagesequenceunderstanding, yao2024minicpm}.
Building on this success, a significant research frontier has emerged in applying these models to dynamic, interactive visual environments (including games \cite{DBLP:journals/tmlr/WangX0MXZFA24}, GUI \cite{qin2025ui,gu2025mobile}, and embodied AI \cite{zitkovich2023rt}).
However, prevalent benchmarks and environments exhibit significant limitations in simulating the authenticity and complexity of physical interactions.
Game environments \cite{bellemare2013arcade,fan2022minedojo,tan2025stardojo} typically offer high-level observation/action space and simplified physics, bypassing the need for low-level physical reasoning. 
GUI environments \cite{xie2024osworld,rawles2024androidworlddynamicbenchmarkingenvironment} are not grounded in real-world physics, featuring discrete, non-continuous actions. 
And embodied AI environments \cite{kolve2017ai2,cheng2024legent,robocasa} focus primarily on semantic-level interactions, usually oversimplifying physical dynamics. 
This insufficient modeling of complex physical phenomena restricts agents' ability to learn deep causal relationships between actions and longer-term physical consequences.
To address this gap, we propose DeepPHY, a benchmark emphasizing agents' need to perceive and understand physical consequences of their actions through sustained interaction.

DeepPHY systematically integrates six challenging physics-based simulation environments: PHYRE \cite{bakhtin2019phyre}, I-PHYRE \cite{iphyre}, Kinetix \cite{Kinetix}, Pooltool \cite{kiefl2024pooltool}, and games Angry Birds and Cut the Rope\footnote{Popular physics-based puzzle games by Rovio Entertainment and ZeptoLabs, respectively.}. 
\textbf{None} of which have been previously aggregated for benchmarking agentic VLMs. 
This integrated collection stands in stark contrast to existing LLM physical reasoning benchmarks that primarily evaluate physical reasoning through static question-answering formats or text-based physics problems \cite{wang2025phy,shen2025phyx,xiang2025seephys,xu2025ugphysics}. 
DeepPHY instead immerses agents in interactive sandboxes where success hinges on performing actions and understanding their physical consequences over time. 
By curating these diverse challenges from different environments, we create the first comprehensive benchmark specifically dedicated to evaluating the interactive physical reasoning capabilities of agentic VLMs.

\begin{table*}[t]
    \centering
    \caption{Comparative analysis of the different DeepPHY benchmark suite environments across five key dimensions. Legend: \textbf{\doublecmark} - core challenge or implemented with high fidelity; \textbf{\cmark} - present and relevant for solving tasks; and \textbf{\xmark} - not a primary focus or missing.}
    \label{tab:deepphy_detailed}
    \resizebox{\textwidth}{!}{
    \begin{tabular}{lcccccc}
    \toprule
        \textbf{Category} & \textbf{PHYRE} & \textbf{I-PHYRE} & \textbf{Kinetix} &  \pooltoollogo & \angrybirdlogo & \cutrepologo \\ \midrule
        \multicolumn{7}{l}{\textbf{1. Fundamental Physics}} \\
        \quad Collision \& Stability & \doublecmark & \cmark & \cmark & \doublecmark & \doublecmark & \cmark \\
        \quad Gravity \& Friction & \cmark & \cmark & \cmark & \doublecmark & \cmark & \cmark \\
        \quad Inertia \& Momentum Transfer & \cmark & \cmark & \doublecmark & \doublecmark & \cmark & \cmark \\
        \midrule
        \multicolumn{7}{l}{{\textbf{2. Complex Mechanisms \& Dynamics}}} \\
        \quad Articulated \& Multi-body Dynamics & \xmark & \cmark & \doublecmark & \cmark & \xmark & \cmark \\
        \quad Tension \& Oscillation (Ropes, Springs) & \xmark & \doublecmark & \xmark & \xmark & \xmark & \doublecmark \\
        \quad Leverage \& Rotational Force & \xmark & \cmark & \doublecmark & \doublecmark & \xmark & \xmark \\
        \midrule
        \multicolumn{7}{l}{{\textbf{3. Agent Action \& Control}}} \\
        \quad Decision Horizon & Single-step &  Sequential & Sequential & Sequential & Sequential & Sequential \\
        \quad Planning Strategy & In-advance & In-advance & On-the-fly & On-the-fly & On-the-fly & On-the-fly \\
        \quad Control Complexity & Single Tool & Multi-tool & Multi-tool & Single Tool & Multi Tool & Multi-tool \\
        \quad Timing Criticality & Low & High & Medium & Low & Low & High \\
        \midrule
        \multicolumn{7}{l}{{\textbf{4. Reasoning \& Strategy}}} \\
        \quad Causal Chain Reasoning & \doublecmark & \doublecmark & \cmark & \cmark & \cmark & \doublecmark \\
        \quad Tool Use \& Affordance & \cmark & \cmark & \cmark & \cmark & \doublecmark & \doublecmark \\
        \quad Adaptation to Dynamic Novelty & \xmark & \xmark & \cmark & \xmark & \doublecmark & \doublecmark \\
        \midrule
        \multicolumn{7}{l}{{\textbf{5. Evaluation Setup}}} \\ 
        \quad \# Test Instance & 1000 & 40 & 74 & 100 & 34 & 88 \\
        \quad Evaluation Strategy & Env.  & Env.  & Env.  & Env. & Manual & Manual \\
        \quad Max Attempts or Steps & 10 & 10 & 16 & 15 & \# Birds & 10 \\
    \bottomrule
    \end{tabular}
    }
\end{table*}

Through this work, we aim to reveal the boundaries and core shortcomings of current VLMs. 
Our extensive evaluation across the DeepPHY suite sheds light on their limits in complex physical interaction, long-horizon planning, and dynamic adaptation. 
The key contributions of this paper are:
\begin{itemize}[leftmargin=*]
    \item We introduce \textbf{DeepPHY}, the first comprehensive benchmark suite to systematically evaluate interactive physical reasoning in agentic VLMs.
    \item We develop a \textbf{unified framework and standardized metrics} that transform diverse physics simulators into a rigorous and accessible testbed.
    This platform evaluates VLMs and collects interaction data useful for training more physically realistic AI agents.
    \item We conduct an \textbf{extensive empirical study} of leading open- and closed-source VLMs, providing clear baselines and revealing their limitations in physical interaction, planning, and adaptation.
\end{itemize}

\section{Related Work}

Physical reasoning capability serves as the cornerstone of world model \cite{wu2024ivideogpt,agarwal2025cosmos} construction and embodied intelligence \cite{Being-0} tasks. 
However, most evaluations rely on static problem-solving benchmarks.
These benchmarks, often presented as large-scale QA about object properties \cite{wang2023newtonlargelanguagemodels,chow2025physbench} or as text-based physics exams \cite{wang2025phy,chung2025theoreticalphysics,zhang2025abenchphysics}, test agents' ability to recall scientific knowledge or deduce logical outcomes from fixed context. 
While valuable for assessing declarative knowledge, they fundamentally avoid challenges of real-time visual perception and continuous interaction with dynamic worlds. 
Consequently, such benchmarks are insufficient for holistically evaluating physical intelligence, which necessitates a closed loop of observation, action, and interaction within physics-driven environments.

Another category of research focuses on physical reasoning within simulated environments, but often abstracts away the challenge of perceptual grounding by relying on symbolic inputs.  
Common approaches provide agents with pre-processed symbolic inputs like object property matrices \cite{bakhtin2019phyre,iphyre,Kinetix} or interact with simulators via code generation \cite{LLMPhy}.
While these methods are powerful for isolating specific planning, they limit generalizability by bypassing the understanding of raw sensory data.
DeepPHY, in contrast, is the first benchmark designed specifically to bridge this gap, evaluating agents' interactive physical reasoning directly from visual input in dynamic settings.

Moreover, research on agents in gaming environments \cite{xu2024survey}, while demonstrating impressive capabilities, often operates on a different level of abstraction. 
Even when visual information is involved, most existing game agents operate in narrative-driven environments — such as GTA \cite{yang2024octopus}, Escape Room \cite{wang2025escapecraft}, StarCraft II \cite{shao2024swarmbrain}, Red Dead Redemption II \cite{tan2025cradle}, or Civilization VI \cite{qi2024civreal} — where progression hinges on scripted storylines or discrete mechanics.
This allows agents to learn game-specific rules and heuristics rather than inferring the underlying physical laws governing the world. 
This focus on game mechanics, rather than fundamental physics, sidesteps the core challenge of building agents that can reason from first-principles based on raw visual observation.

To address these issues, we introduce DeepPHY, an interactive physics simulation benchmark designed to comprehensively evaluate the capabilities of agentic VLMs. 
By immersing agents in diverse, dynamic, and vision-based environments that require direct interaction, DeepPHY moves beyond static knowledge recall and symbolic reasoning to directly assess an agent's ability to perceive, act, and reason in worlds governed by physical principles.

\section{DeepPHY}

DeepPHY is a benchmark framework designed to evaluate whether existing VLMs acting as agents possess the ability to understand physical environments and perform sequential action planning tasks.
Table \ref{tab:deepphy_detailed} provides a comparative analysis of the six environments across key dimensions.
This breakdown illustrates how each environment contributes unique challenges to the DeepPHY benchmark, ensuring a comprehensive evaluation of an agent's physical intelligence. 
In the remainder of this section, we introduce the physics environments employed in the benchmark.

\subsection{Problem Formalization}

We formalize the physical reasoning challenge in DeepPHY as a trial-based decision process. 
The underlying environment for any single attempt is a Partially Observable Markov Decision Process (POMDP), $\mathcal{M} = (\mathcal{S}, \mathcal{A}, \mathcal{T}, \mathcal{R}, \Omega, \mathcal{O})$, where $\mathcal{S}$ is the latent physical state, $\mathcal{A}$ is the action space for a single step, and $\mathcal{R}$ is a sparse reward given only for final task success.
An agent is allowed up to $K$ attempts to solve a given task, which always starts from the same initial observation $o_{\text{initial}}$. 
The core challenge is not just to find a solution, but to learn from failed attempts. 
We model this as follows:
Let $k \in \{1, \dots, K\}$ be the attempt index. A single attempt, or trial, results in a trajectory $\tau^{(k)}$.
\begin{itemize}[leftmargin=*]
    \item For \textbf{in-advance planning} environments (e.g., PHYRE), the agent submits a complete action plan $a^{(k)}$ based on $o_{\text{initial}}$ and past history. The resulting trajectory is a tuple capturing the plan and its outcome:
    \begin{equation*}
        \tau^{(k)} = (o^{(k)}, a^{(k)}, o_{\text{final}}^{(k)}, r^{(k)})
    \end{equation*} 
    where $o_{\text{final}}^{(k)}$ is the final visual observation after the plan is executed, and $r^{(k)}$ is the terminal reward (e.g., 1 for success, 0 for failure).

    \item For \textbf{on-the-fly planning} environments (e.g., Kinetix), the agent interacts sequentially with the environment. The trajectory consists of the full sequence of interactions, culminating in a final reward:
    \begin{equation*}
        \tau^{(k)} = (o_0^{(k)}, a_0^{(k)}, o_1^{(k)}, a_1^{(k)}, \dots, o_T^{(k)}, r^{(k)})
    \end{equation*}
    where $o_0^{(k)} = o_{\text{initial}}$, and $r^{(k)}$ is the terminal reward received after the final observation $o_T^{(k)}$.
\end{itemize}

At the beginning of each new attempt $k$, the agent has access to the history of all previous failed trials, $H^{(k)} = \{\tau^{(1)}, \tau^{(2)}, \dots, \tau^{(k-1)}\}$. 
To succeed, the agent must use this history to refine its internal world model, $f_{\text{phy}}$, and improve its planning. 
The policy for generating the plan for the $k$-th attempt, $\pi^{(k)}$, can be expressed as:
\begin{equation*}
    \pi^{(k)} = \underset{\pi \in \Pi}{\arg\max} \, V(f_{\text{phy}}(o_{\text{initial}}, H^{(k)}, \pi))
\end{equation*}
where $\Pi$ is the space of all possible plans (either a single action or a sequence policy). 
The function $f_{\text{phy}}(\cdot)$ now represents the agent's ability to simulate the outcome of a candidate plan $\pi$, conditioned not only on the initial observation but also on the history of past failures $H^{(k)}$. 
The value function $V(\cdot)$ estimates the likelihood of success for that simulated outcome.
This formulation highlights that DeepPHY fundamentally evaluates an agent's capacity for iterative refinement and learning from failure by updating its predictive model, $f_{\text{phy}}$, across trials.

\subsection{Environments}

DeepPHY integrates the following environments:

\begin{table*}[h!]
\centering
\caption{Summary of observation \& action space conversions in the DeepPHY benchmark.}
\label{tab:observation_action_space}
\footnotesize
\begin{tabular}{@{} l p{5.8cm} p{7cm} @{}}
\toprule
\textbf{Environments} & \textbf{Observation Space} & \textbf{Action Space} \\ 
\midrule
\textbf{PHYRE} & Initial scene image and identical image overlaid with a 5x5 grid. & \textbf{Discretized selection} of grid \texttt{Cell} (1-25) and \texttt{Radius} level (1-5). \\ 

\textbf{I-PHYRE} & Scene image with interactive blocks annotated. & \textbf{JSON array} specifying a sequence of block \texttt{index} and \texttt{time} (in seconds) for each elimination. \\ 

\textbf{Kinetix} & High-clarity rendered image, plus an annotated version of controllable motors and thrusters (e.g., `M0', `T0'). & \textbf{JSON array} of integers where each integer corresponds to a discrete action (e.g., 0: \textit{off}, 1: \textit{forward}, 2: \textit{reverse}). \\ 

\pooltoollogo & A 3D to 2D top-down rendered view of the pool table. & \textbf{Discretized selection} from a predefined list of \texttt{Speed} options (``Low", ``Medium", ``High") and \texttt{Strikespot} options (e.g., ``Top Spin", ``Bottom Left Spin"). \\ 

\angrybirdlogo & Screenshot of the current game state, including structures, pigs, and available birds. & \textbf{Python code} specifying shoot \texttt{angle} (0-90) and \texttt{power} (0.0-1.0). \\ 

\cutrepologo & Annotated screenshot with IDs on all interactive elements (pins, cushions, etc.). Bubbles are labeled in real-time. & \textbf{Python code}, predefined action like \texttt{cut\_pin(id)}, \texttt{pop \_bubble(id)}, \texttt{tap\_air\_cushion(id, times)}, etc. \\ 

\bottomrule
\end{tabular}
\end{table*}

\textbf{PHYRE} \cite{bakhtin2019phyre}. PHYRE provides a suite of 2D physical reasoning tasks that require agents to achieve specified goals by placing interactive objects in the scene that trigger correct physical chain reactions within limited attempts.
Detailed in Appendix \ref{appendix:phyre}.

\textbf{I-PHYRE} \cite{iphyre}. I-PHYRE is a dynamically evolving interactive physics reasoning benchmark where agents solve puzzles by removing obstacles at precise timings in correct sequences. 
Detailed in Appendix \ref{appendix:i-phyre}.

\textbf{Kinetix} \cite{Kinetix}. An open-source 2D physics simulation platform designed for agents. 
It generates vast and diverse physical control tasks, spanning scenarios from robotic locomotion and grasping to classical control problems. 
Detailed in Appendix \ref{appendix:kinetix}.

\pooltoollogo\ \cite{kiefl2024pooltool}. A high-fidelity \texttt{billiards} simulation benchmark that accurately models multibody collisions, \texttt{English spin} effects, and friction-induced trajectory alterations. 
Detailed in Appendix \ref{appendix:pooltool}.

\angrybirdlogo. A physics-based puzzle game where agents strategically launch \texttt{birds} to dismantle complex structures (e.g., wood, stone, ice), aiming to eliminate designated targets through precise force and angle calculations, and trigger large-scale chain-reaction collapses for efficient puzzle resolution.
Detailed in Appendix \ref{appendix:angrybirds}.

\cutrepologo. A physics-based puzzle game where agents manipulate a dynamic system by cutting ropes and touching other props (e.g., bubbles and cushions) at precise moments to guide candy to a little green monster named \textit{Om Nom}.
Detailed in Appendix \ref{appendix:rope}.

\subsection{Observation Space}

DeepPHY's primary aim is to evaluate the physical reasoning of agents, rather than mere perceptual localization. 
To this end, we refactor and augment each environments' observation space, providing clear annotated-image renderings of interactive objects’ locations and identities, thus minimizing the burden of object detection. 
This design directs agents' to understanding physical dynamics and planning manipulations, enabling a more targeted assessment of their physical reasoning capabilities.
This augmentation is realized in distinct ways across the benchmark suite:

\textbf{Gridded and Labeled Overlays:} Visual scenes are optionally overlaid with grids or numerical IDs. In PHYRE, a 5x5 grid is superimposed on the scene to discretize object placement locations. In I-PHYRE, Kinetix and Cut the Rope, interactive elements are annotated with numerical labels. 

\textbf{Dimensionality Reduction:} For Pooltool, the native 3D environment is converted into a more VLM-friendly 2D top-down view, simplifying the visual information and making spatial relationships clearer.

\textbf{Direct Visual Input:} In Angry Birds, observation space consists of a raw screenshot of the game—showing structures and pigs without explicit labels—and textual information of available birds.

By providing these modified observations, DeepPHY ensures that the core challenge remains centered on understanding and predicting physical dynamics.

\subsection{Action Space}

A core challenge for current VLMs not specifically pre-trained or fine-tuned for agentic control is their poor performance in generating actions within a continuous space. Specifying precise coordinates, forces, or angles in a free-form text response is often unreliable. 
To address this, a common principle across all DeepPHY environments is the transformation of continuous or complex action spaces into discrete and structured formats. 
This conversion is tailored to each environment to preserve its core physical challenges, while making interaction more feasible for VLMs. 
This strategy is vital as existing games are typically designed for humans, who can provide timely and precise analog feedback. 
For VLMs, especially in a zero-shot setting, these discrete and structured action spaces make the tasks tractable.

Our approach differs across the suite environments, demonstrating various methods of discretization:

\textbf{Discretized Parameter Space:} In PHYRE, the continuous action of placing a ball at any (x, y) coordinates with any radius is converted into selecting one of 25 grid cells and one of 5 radius levels. Similarly, in Pooltool, the agent chooses from a small, predefined set of named options for shot power (3 selections) and spin type (9 selections), abstracting away the need to specify continuous force and offset values.

\textbf{Integer-to-Command Mapping:} In Kinetix, the agent outputs a simple integer vector. Each integer maps to a specific action for a corresponding motor or thruster. This allows for coordinated control of multiple components through simple, structured output.

\textbf{Structured Command Language:} For games like Angry Birds and Cut the Rope, which involve a variety of interaction types, we have designed a predefined Python code. In Angry Birds, the agent must generate a launch command with angle and power parameters - \texttt{shoot(angle, power)}. In Cut the Rope, the agent generates actions as \texttt{cut\_pin({Index})}, \texttt{pop\_bubble({Index})}, etc. This provides the agent with a powerful yet constrained set to interact with the environment.

\textbf{Structured Data Format:} In I-PHYRE, the agent outputs a JSON list where each object specifies the index of a block and the precise time of its removal, enabling complex timed sequences of actions.

Table \ref{tab:observation_action_space} provides a summary of the observation and action space conversions for each environment in the DeepPHY benchmark suite.

\section{Evaluation Protocol}

In this section, we describe our protocols for evaluating various current state-of-the-art VLMs on DeepPHY. 

\subsection{Evaluation Setting}

We aim to keep the evaluation setting simple and consistent across environments. During each timestep of interaction, agents are prompted to output their next action, conditioned on their past interaction history with the environment. To perform successfully in DeepPHY, models must demonstrate robust instruction-following capabilities, including reading visual observation scenes and interpreting environment rules, understanding the action space, inferring physical principles, and producing valid actions to complete tasks effectively.

\noindent \textbf{Planning Strategy.} We categorize the environments into two distinct interaction paradigms based on their decision-making process. 

\begin{itemize}
    \item \textbf{In-advance Planning}, where an agent is required to devise a complete solution plan from the initial state and output it as a single action (or a full sequence of actions). If the plan fails, the agent receives feedback on the failure and must generate a new complete plan in the next attempt. This setup tests the agent's ability for comprehensive causal reasoning and foresight.
    \item \textbf{On-the-fly Planning}, requiring sequential, turn-by-turn interaction with the environment. At each step, the agent outputs a single action, observes immediate physical consequences, and decides on the next action based on the new state. This mode evaluates the agent's capacity for continuous observation, dynamic adaptation, and reactive control.
\end{itemize}

\noindent \textbf{Prompt Format.} We use two prompting strategies. 

\begin{itemize}
    \item \textbf{Vision-Language-Action (VLA)}, where the model receives environment rules, current visual observation, and history of failed attempts, then directly outputs an action.
    \item \textbf{World Model (WM)}, which builds on the VLA prompt, also requiring predicting the environmental changes that will result from the model's chosen action.
\end{itemize}

\noindent \textbf{Metrics.} 
We use three primary metrics.

\begin{itemize}
    \item  \textbf{Success Rate:} The fraction of tasks solved successfully.
    \item \textbf{Pass@K:} The proportion of tasks solved within a maximum of K attempts.
    \item \textbf{Average Attempts:} The mean number of interactive attempts taken to solve a task, averaged over successful trials only.
\end{itemize}

\subsection{Models}

We evaluate 17 popular open- and closed-source models, as detailed in Table~\ref{tab:model_abbreviations}.
Open-source models include Qwen2.5-VL-3B/7B/32B/72B-Instruct \cite{qwen2.5-VL}.
Closed-source models include Claude 3.5/3.7/4.0 Sonnet and Claude 4.0 Opus \cite{anthropic2024claude, anthropic2025claude}; 
Gemini-2.0-Flash, Gemini-2.5-Pro-06-17, and Gemini-2.5-Flash-06-17 \cite{mallick2025gemini, comanici2025gemini2.5};
and GPT-4-Vision-Preview, GPT-4o-mini-0718, GPT-4o-0806, GPT-o3-0416, and GPT-o4-mini-0416 \cite{achiam2023gpt,openai2024gpt4o, openai2025o3ando4mini}.
Furthermore, a \texttt{\textit{MOCK}} model, which performs random actions, is included as a baseline.

\section{Experimental Results}

Here, we evaluate all VLMs on DeepPHY, to establish their \textbf{zero-shot} baseline performance. We report Success Rate, Pass@K, and Average Attempts over 3 runs under the different physical environments, with a temperature set to 0.1.

\begin{figure}[h]
    \centering
    \begin{subfigure}[b]{0.7\linewidth}
        \centering
        \includegraphics[width=\linewidth]{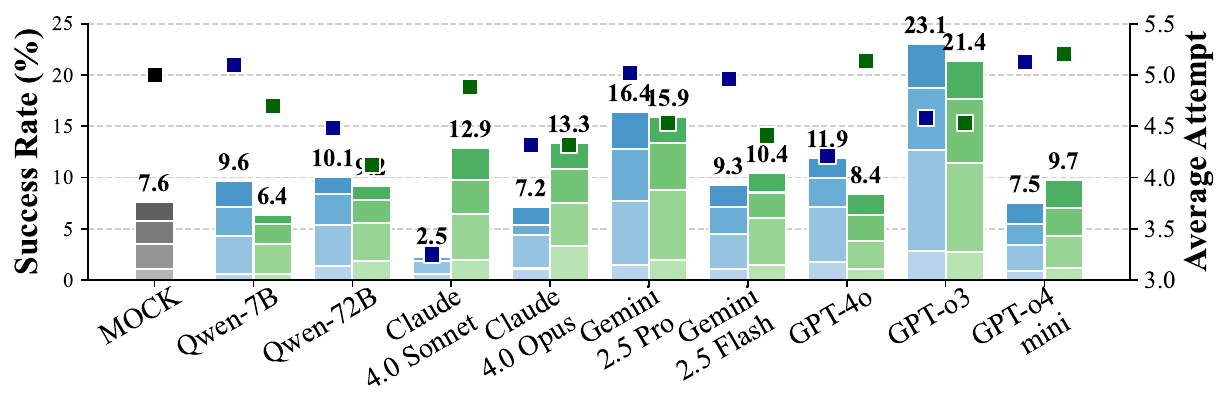}
    \end{subfigure}
    \begin{subfigure}[b]{0.7\linewidth}
        \centering
        \includegraphics[width=1\linewidth]{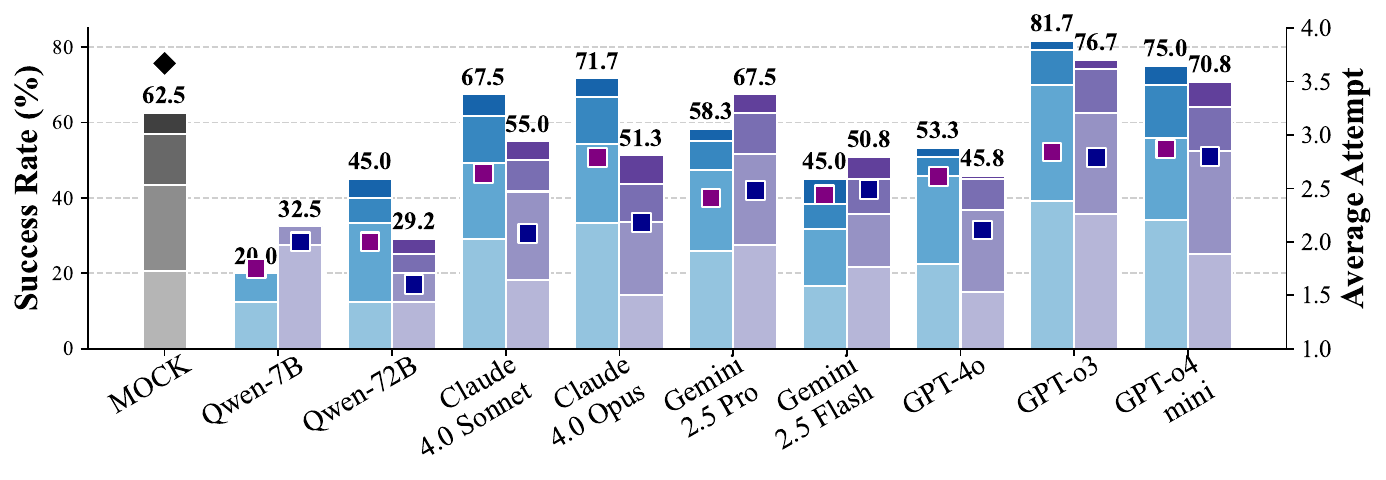}
    \end{subfigure}
    \caption{Performance comparison on the PHYRE (top) and I-PHYRE (bottom) environments.
The stacked bars (VLA - left side, WM - right side) show the cumulative success rate (left y-axis), with segments indicating gains after 1, 4, 7, and 10 attempts (Detailed in Appendices \ref{appendix:phyre} \& \ref{appendix:i-phyre}). Overlaid scatter points correspond to Average Attempts (right y-axis).}
\label{fig:PHYRE_IPHYRE}
\end{figure}
\begin{figure*}[h]
\centering
\includegraphics[width=1\linewidth]{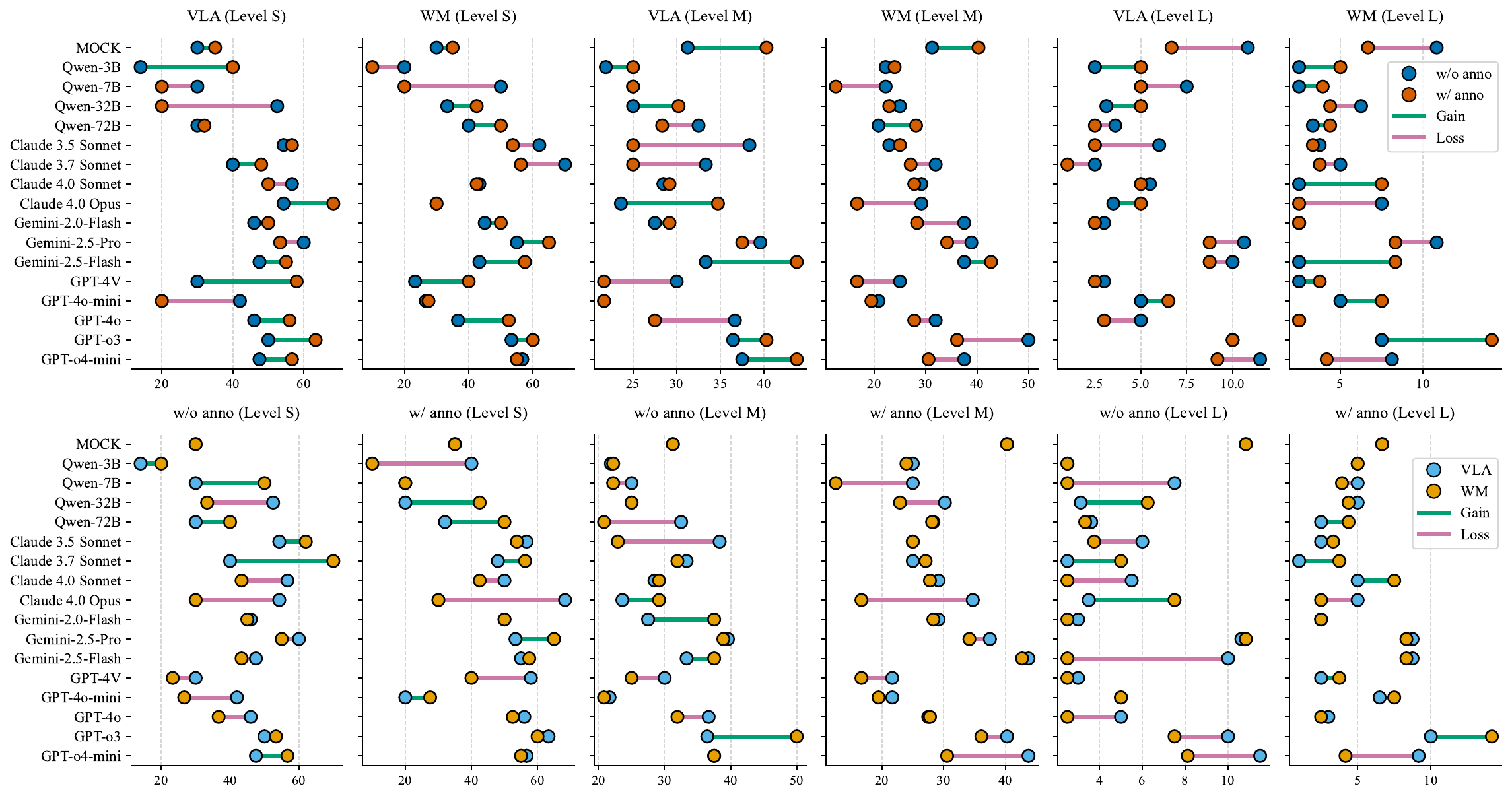}
\caption{Model performance comparison in the Kinetix environment across task difficulty (S, M, L), Prompt Format (VLA vs. WM), and ablation study of annotations (``anno", as illustrated in Figure \ref{fig:Kinetix_render_example} in the Appendices).}
\label{fig:kinetix_dumbbell_plot_all}
\end{figure*}
\begin{table}[h!]
\centering
\caption{Model performance summary across Pooltool, Angry Birds, and Cut the Rope. Stars (\goldstar) in Angry Birds are based on fewer launches and greater building destruction, while in Cut the Rope, they are collected during gameplay. \firstlogo\ and \secondlogo\ denote the highest model scores in each environment (besides \textit{Human}).}
\label{tab:model_performance_master_summary}
\resizebox{0.7\linewidth}{!}{
\begin{tabular}{l | ll | ll | ll}
\toprule
\multirow{2}{*}{\textbf{Model}} & \multicolumn{2}{c|}{\pooltoollogo} & \multicolumn{2}{c|}{\angrybirdlogo} & \multicolumn{2}{c}{\cutrepologo} \\
\cmidrule(lr){2-3} \cmidrule(lr){4-5} \cmidrule(lr){6-7}
& \textbf{Att. 15} & \textbf{Avg. Att.} & \textbf{SuccRate} & \textbf{Total\goldstar} & \textbf{SuccRate} & \textbf{Total\goldstar} \\
\midrule
\texttt{\textit{MOCK}} & 48.00\% & 7.88 & 17.65\% & 14.00 & 11.36\% & 14.00 \\
\midrule
\rowcolor{lightgray}\multicolumn{7}{c}{\textbf{Open-Source Models}} \\ \midrule
\textbf{Qwen-3B} & 50.00\% & 2.00 & 17.65\% & 13.00 & 7.95\% & 11.00 \\
\textbf{Qwen-7B} & 26.50\% & 2.58 & 20.59\% & 13.00 & 9.09\% & 9.00 \\
\textbf{Qwen-32B} & 14.29\% & 7.00 & 26.47\% & 14.00 & 6.82\% & 10.00 \\
\textbf{Qwen-72B} & 18.00\% & 10.75 & 29.41\% & 23.00 & 13.64\% & 22.00 \\
\midrule
\rowcolor{lightgray}\multicolumn{7}{c}{\textbf{Close-Source Models}} \\ \midrule
\rowcolor{mylightgray}\multicolumn{7}{c}{\textbf{Claude Series}} \\
\textbf{Claude 3.5 Sonnet} & 67.00\% & 7.04 & 26.47\% & 20.00 & 21.59\% & 29.00 \\
\textbf{Claude 3.7 Sonnet} & 72.50\% & 3.64 & 41.18\% \firstlogo & 27.00 & 20.45\% & 24.00 \\
\textbf{Claude 4.0 Sonnet} & 44.00\% & 10.30 & 35.29\% \secondlogo & 24.00 & 22.73\% \secondlogo & 29.00 \\
\textbf{Claude 4.0 Opus} & 49.00\% & 9.98 & 32.35\% & 23.00 & 26.14\% \firstlogo & 33.00 \\
\midrule
\rowcolor{mylightgray}\multicolumn{7}{c}{\textbf{Gemini Series}} \\
\textbf{Gemini-2.0-Flash} & 75.00\% \secondlogo & 7.51 & 20.59\% & 16.00 & 18.18\% & 25.00 \\
\textbf{Gemini-2.5-Flash} & 29.00\% & 7.19 & 29.41\% & 22.00 & 12.50\% & 15.00 \\
\textbf{Gemini-2.5-Pro} & 68.00\% & 4.17 & 35.29\% \secondlogo & 25.00 & 22.73\% \secondlogo & 30.00 \\
\midrule
\rowcolor{mylightgray}\multicolumn{7}{c}{\textbf{GPT Series}} \\
\textbf{GPT-4V} & 39.50\% & 10.16 & 29.41\% & 21.00 & 15.91\% & 15.00 \\
\textbf{GPT-4o-mini} & 100.00\% \firstlogo & 7.50 & 23.53\% & 14.00 & 7.95\% & 8.00 \\
\textbf{GPT-4o} & 34.50\% & 9.86 & 32.35\% & 24.00 & 17.05\% & 19.00 \\
\textbf{GPT-o3} & 25.67\% & 8.64 & 35.29\% \secondlogo & 24.00 & 18.18\% & 30.00 \\
\textbf{GPT-o4-mini} & 53.00\% & 8.47 & 32.35\% & 22.00 & 17.05\% & 26.00 \\
\midrule
\rowcolor{mylightgray}\textit{\textbf{Human}} & {100\%} & {12.34} & 64.71\% & 44.00  & 41.36\% & 91.60 \\
\bottomrule
\end{tabular}
}
\end{table}

\begin{figure}[h]
\centering
\begin{subfigure}[b]{0.35\linewidth}
    \centering
    \includegraphics[width=\linewidth]{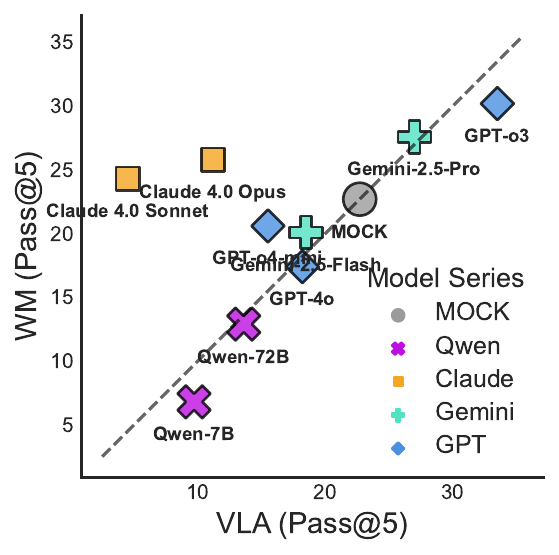}
\end{subfigure}
\begin{subfigure}[b]{0.35\linewidth}
    \centering
    \includegraphics[width=\linewidth]{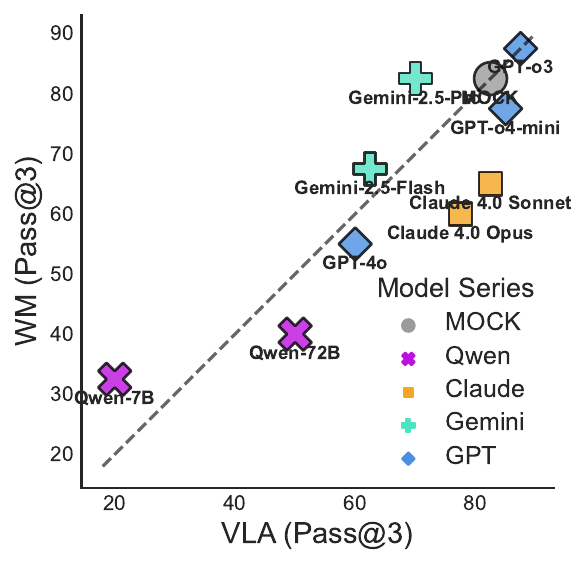}
\end{subfigure}
\caption{Performance comparison between prompt formats on (a) PHYRE (Pass@5) and (b) I-PHYRE (Pass@3) environments. Points represent each model and the diagonal line indicates equal prompt format performance (points above the line demonstrate WM advantage over VLA.}
\label{fig:PHYRE_IPHYRE_VLA_WM}
\end{figure}
\begin{figure}[h]
\centering
\includegraphics[width=0.7\linewidth]{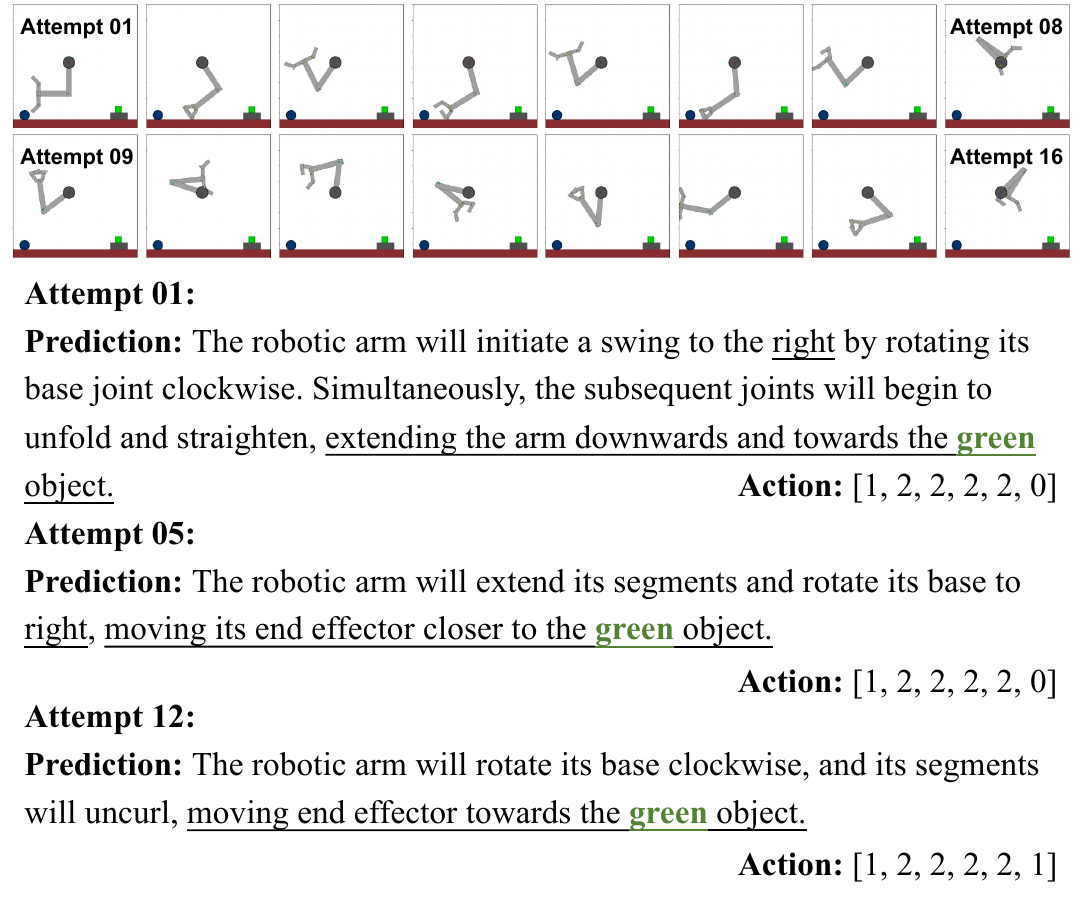}
\caption{Case Study of descriptive physical reasoning in Kinetix - Claude 4.0 Sonnet (WM).}
\label{fig:kinetix_casestudy_claude_sonnet4}
\end{figure}

\subsection{Overall Performance Analysis}

The experimental results reveal that current VLMs face significant challenges in interactive physical reasoning tasks.
As shown in Figures \ref{fig:PHYRE_IPHYRE} and \ref{fig:kinetix_dumbbell_plot_all}, and Table \ref{tab:model_performance_master_summary}, most models (especially open-source ones) still cannot surpass \texttt{\textit{MOCK}} results, even with the simplified action spaces designed for VLMs.
This indicates a lack of deep understanding of underlying physical principles and for zero-shot planning ability.
Among all tested models, the latest flagship closed-source models (e.g., GPT-o3, Gemini-2.5-Pro, and Claude 4.0 Opus) generally demonstrate superior performance to \texttt{\textit{MOCK}}, which suggests that model scale, data quality, and more advanced architectures benefit physical reasoning capabilities. However, all models behave differently across environments, and crucially, a stark performance gap exists when compared to humans. Models' success rates fall considerably shorter than desirable, underscoring the long road ahead in bridging the gap to achieve physical intelligence.

\subsection{Detailed Analysis by Environment}

\textbf{PHYRE:} Even the most advanced models generally achieve Success Rate at 1st attempt below 4\% (Figure \ref{fig:PHYRE_IPHYRE}, and Tables \ref{tab:model_performance_comparison_phyre_vla} \& \ref{tab:model_performance_comparison_phyre_wm} in Appendix \ref{appendix:phyre}).
While successes increase with additional attempts, the improvement is slow, suggesting that models struggle to learn effectively from failed attempts and to revise their strategies accordingly (best at only 23.1\%).

\textbf{I-PHYRE:} Leading models such as GPT-o3 achieve a relatively high Success Rate in this setting, reaching 81.67\% at Attempt 10 (Figure \ref{fig:PHYRE_IPHYRE}, and Tables \ref{tab:iphyre_model_performance_vla_avg_step} and \ref{tab:iphyre_model_performance_wm_avg_step} in Appendix \ref{appendix:i-phyre}), demonstrating that these models are capable of effective temporal planning and causal reasoning. 
However, open-source models still perform poorly in this setting, resulting in lower success rates than the \texttt{\textit{MOCK}} baseline.

\textbf{Kinetix:} Model success rates decrease significantly as task difficulty increases (Figure \ref{fig:kinetix_dumbbell_plot_all}, and Tables \ref{tab:Kinetix_acc_average}, \ref{tab:Kinetix_acc_total} in Appendix \ref{appendix:kinetix}).
The effect of visual annotation is mixed. 
In simple S-level tasks, annotations help models identify controllable components and improve performance. 
However, on more difficult M- and L-level tasks, this benefit vanishes or even harms performance, especially with the WM prompt. 
This suggests on harder tasks, the extra labels become a cognitive distraction.
Furthermore, the WM prompt also fails to help, often lowering success rates. 
This reveals a VLM's inability to form an accurate world model from visual input.

\pooltoollogo: 
As shown in Table \ref{tab:model_performance_master_summary} (for extra details, see Tables \ref{tab:model_performance_comparison_pooltool_vla} \& \ref{tab:model_performance_comparison_pooltool_wm}), top models, e.g., GPT-4o-mini, achieve high success rates, sometimes even outperforming the human player in average steps.
However, a closer look reveals this ``efficiency" stems not from strategy, but from a ``brute-force heuristic".
Models consistently use maximum power to shoot the target ball along the most direct path into a pocket. 
This strategy works on simple layouts, but ignores the core skill: cue ball control. 
Our analysis shows that no tested model uses complex physics for planning. 
For example, they rarely use side spin to alter angles or top/bottom spin to control the cue ball for the next shot. 
The 100\% success rate of GPT-4o-mini is thus misleading, a product of deterministic output and a non-random environment. The model simply repeats the same brute-force solution that happened to work.

\angrybirdlogo\ \& \cutrepologo:
As shown in Table \ref{tab:model_performance_master_summary} 
(for extra details, see Tables \ref{tab:model_performance_comparison_angrybirds_vla} \& \ref{tab:model_performance_comparison_cutrope_vla}), a vast performance gap separates all models from the human player in these games. 
This exposes a core weakness of current VLMs: they struggle with complex, multi-stage physics tasks that require precise timing. 
Our design gives clear visual labels to all interactive elements, like slingshots, ropes, and bubbles. 
This shifts the challenge from low-level perception to high-level reasoning. 
Although models can see what is interactive, they still cannot create reliable action sequences to reach goals. 
This highlights their fundamental limitations in spatiotemporal reasoning for dynamic physical processes, with most failures stemming from wrong timing or sequencing. 
For example, in Cut the Rope, a model might cut a rope too early, stopping the candy from gaining enough momentum. 
In Angry Birds, models find it hard to plan attack sequences that cause chain reactions, failing to understand how one bird's impact changes the structure for the next attack. 
These systematic failures show that current models cannot build a coherent predictive internal world model for multi-step decisions in dynamic environments.

\subsection{Case Study of Prompt Format}

Our comparative analysis of VLA v.s. WM prompt formats (Figures \ref{fig:PHYRE_IPHYRE}, \ref{fig:kinetix_dumbbell_plot_all}, and \ref{fig:PHYRE_IPHYRE_VLA_WM}) exposes the intrinsic limitations of current agentic VLMs. Across all environments, we find that the WM approach currently offers only limited benefits, primarily in simpler tasks (e.g., in PHYRE/I-PHYRE). Forcing a prediction might help the model avoid purely random exploration in initial, ``zero-history" situations, but this advantage is fragile and quickly diminishes as complexity increases.
In fact, WM often becomes a liability in more complex tasks. 

We theorize that when the intrinsic difficulty of a physical planning task already pushes a model to its limits, the additional demand of generating an accurate dynamic prediction (the WM task) imposes excessive overhead. Moreover, the world modeling capabilities of current models remain underdeveloped.
Revealingly, even if models can generate textually correct predictions (as shown in Figure \ref{fig:kinetix_casestudy_claude_sonnet4}), offering accurate descriptions of the desired physical outcome, they still fail to translate this descriptive knowledge into a precise, executable control signal to realize that outcome. 
This highlights models' \textbf{physical understanding} as largely descriptive, rather than possessing true predictive and procedural control capabilities.

\section{Conclusion}

In this paper, we introduce DeepPHY, the first comprehensive benchmark to evaluate VLMs' interactive physical reasoning. Our systematic evaluation reveals that even state-of-the-art models struggle with precise, multi-step planning in dynamic environments. 
We discover a fundamental disconnect between a model's ability to describe physical phenomena and its ability to use that knowledge to predict and control outcomes. 
We release DeepPHY as a rigorous testbed to benchmark such limitations and facilitate the development of more physically grounded AI agents.

\bibliographystyle{unsrt}
\bibliography{ref}

\clearpage
\appendix

\clearpage

\section{Model Abbreviations}

\begin{table}[h]
    \centering
    \caption{List of Models Abbreviations}
    \label{tab:model_abbreviations}
    \begin{tabular}{ll}
        \toprule
        \textbf{Model} & \textbf{Abbreviation} \\
        \midrule
        \rowcolor{lightgray}\multicolumn{2}{c}{\textbf{Open-Source Models}} \\ \midrule
        \rowcolor{mylightgray}\multicolumn{2}{c}{\textbf{Qwen2.5-VL Series}} \\
        {Qwen2.5-VL-3B-Instruct}  & Qwen-3B \\
        Qwen2.5-VL-7B-Instruct & Qwen-7B \\
        Qwen2.5-VL-32B-Instruct & Qwen-32B \\
        Qwen2.5-VL-72B-Instruct & Qwen-72B \\
        \bottomrule
        \rowcolor{lightgray}\multicolumn{2}{c}{\textbf{Closed-Source Models}} \\ \midrule
        \rowcolor{mylightgray}\multicolumn{2}{c}{\textbf{Claude Series}} \\
        Claude 3.5 Sonnet & Claude 3.5 Sonnet \\
        Claude 3.7 Sonnet & Claude 3.7 Sonnet \\
        Claude 4.0 Sonnet & Claude 4.0 Sonnet \\
        Claude 4.0 Opus & Claude 4.0 Opus \\
        \midrule
        \rowcolor{mylightgray}\multicolumn{2}{c}{\textbf{Gemini Series}} \\
        Gemini-2.0-Flash & Gemini-2.0-Flash \\
        Gemini-2.5-Pro-06-17 & Gemini-2.5-Pro \\
        Gemini-2.5-Flash-06-17 & Gemini-2.5-Flash \\
        \midrule
        \rowcolor{mylightgray}\multicolumn{2}{c}{\textbf{GPT Series}} \\
        GPT-4-Vision-Preview & GPT-4V \\
        GPT-4o-mini-0718 & GPT-4o-mini \\
        GPT-4o-0806 & GPT-4o \\
        GPT-o3-0416 & GPT-o3 \\
        GPT-o4-mini-0416 & GPT-o4-mini \\
        \bottomrule
    \end{tabular}
\end{table}

\section{Benchmark: PHYRE} \label{appendix:phyre}

\begin{figure*}[h]
    \centering
    \begin{subfigure}[b]{0.16\textwidth}
        \centering
        \includegraphics[width=\linewidth]{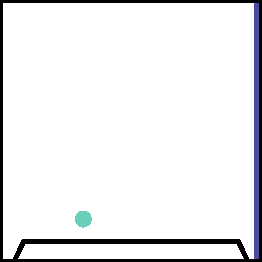}
    \end{subfigure}
    \begin{subfigure}[b]{0.16\textwidth}
        \centering
        \includegraphics[width=\linewidth]{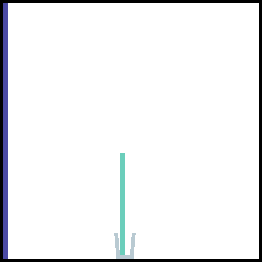}
    \end{subfigure}
    \begin{subfigure}[b]{0.16\textwidth}
        \centering
        \includegraphics[width=\linewidth]{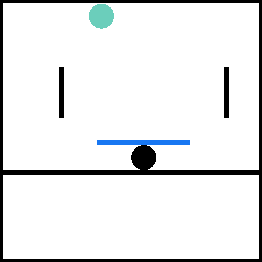}
    \end{subfigure}
    \begin{subfigure}[b]{0.16\textwidth}
        \centering
        \includegraphics[width=\linewidth]{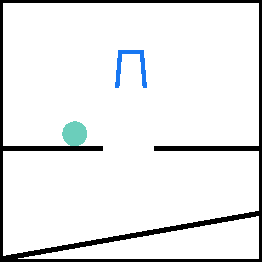}
    \end{subfigure}
    \begin{subfigure}[b]{0.16\textwidth}
        \centering
        \includegraphics[width=\linewidth]{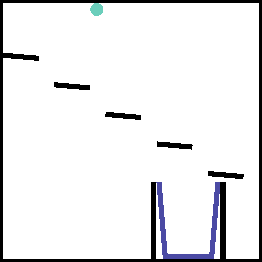}
    \end{subfigure}
    \begin{subfigure}[b]{0.16\textwidth}
        \centering
        \includegraphics[width=\linewidth]{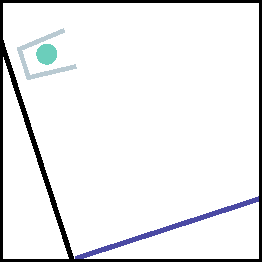}
    \end{subfigure}

    \begin{subfigure}[b]{0.16\textwidth}
        \centering
        \includegraphics[width=\linewidth]{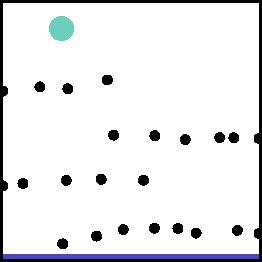}
    \end{subfigure}
    \begin{subfigure}[b]{0.16\textwidth}
        \centering
        \includegraphics[width=\linewidth]{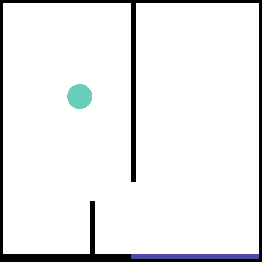}
    \end{subfigure}
    \begin{subfigure}[b]{0.16\textwidth}
        \centering
        \includegraphics[width=\linewidth]{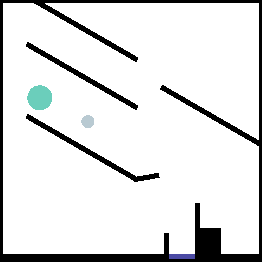}
    \end{subfigure}
    \begin{subfigure}[b]{0.16\textwidth}
        \centering
        \includegraphics[width=\linewidth]{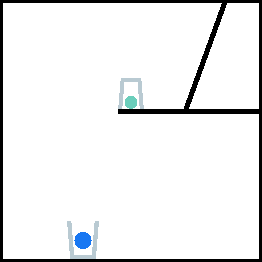}
    \end{subfigure}
    \begin{subfigure}[b]{0.16\textwidth}
        \centering
        \includegraphics[width=\linewidth]{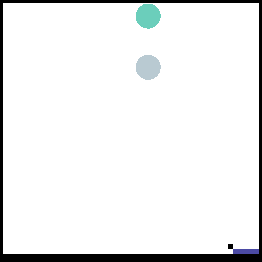}
    \end{subfigure}
    \begin{subfigure}[b]{0.16\textwidth}
        \centering
        \includegraphics[width=\linewidth]{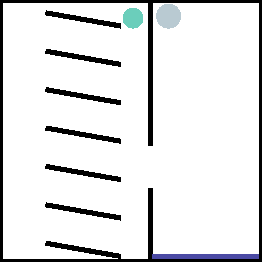}
    \end{subfigure}

    \begin{subfigure}[b]{0.16\textwidth}
        \centering
        \includegraphics[width=\linewidth]{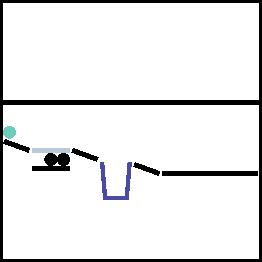}
    \end{subfigure}
    \begin{subfigure}[b]{0.16\textwidth}
        \centering
        \includegraphics[width=\linewidth]{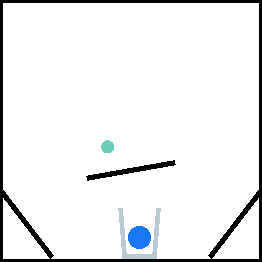}
    \end{subfigure}
    \begin{subfigure}[b]{0.16\textwidth}
        \centering
        \includegraphics[width=\linewidth]{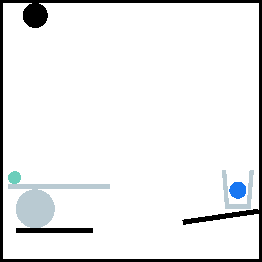}
    \end{subfigure}
    \begin{subfigure}[b]{0.16\textwidth}
        \centering
        \includegraphics[width=\linewidth]{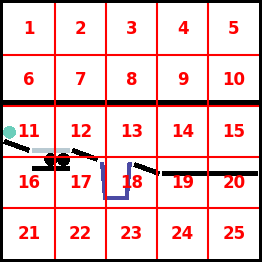}
    \end{subfigure}
    \begin{subfigure}[b]{0.16\textwidth}
        \centering
        \includegraphics[width=\linewidth]{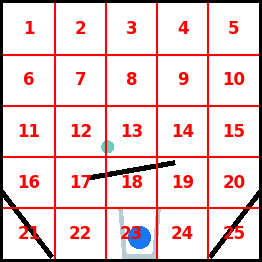}
    \end{subfigure}
    \begin{subfigure}[b]{0.16\textwidth}
        \centering
        \includegraphics[width=\linewidth]{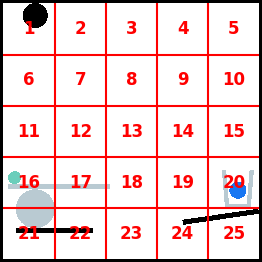}
    \end{subfigure}
    
    \caption{Sample tasks from PHYRE benchmark. The images illustrate the diversity of initial setups for the physical reasoning puzzles.  The last three images on the bottom right show the scene overlaid with the 5x5 grid we use to discretize the placement locations for the agent's actions.}
    \label{fig:PHYRE_sample}
\end{figure*}

The PHYRE\footnote{{https://github.com/facebookresearch/phyre}} benchmark consists of physics puzzles in a simulated 2D world and features two tiers of tasks: PHYRE-B and PHYRE-2B. Tasks in PHYRE-B can be solved by placing a single dynamic ball, whereas tasks in PHYRE-2B require placing two balls. We selected the PHYRE-B tier for our experiments. For tasks in this tier (as shown in Figure \ref{fig:PHYRE_sample}), the objective is to bring the green (dynamic) ball into contact with either the purple (static) ball or the blue (dynamic) ball; black objects are static and grey objects are dynamic. The task distribution of PHYRE is shown in Table \ref{tab:phyre_sample}. The `Within-Template' setting tests generalization to new tasks from seen templates, while the Cross-Template setting tests generalization to tasks from completely unseen templates. 
As our suite does not involve model training or fine-tuning, we combine the test sets from both `Within-Template' and `Cross-Template' settings (1,000 tasks in total) for our final evaluation.

\begin{table}[h]
    \centering
    \caption{Task distribution for the training, validation, and test splits in the PHYRE benchmark.}
    \label{tab:phyre_sample}
    \begin{tabular}{c|ccc}
        \toprule
         \textbf{Generalization Setting} & \textbf{Train} & \textbf{Eval} & \textbf{Test} \\ \midrule
         Within-Template & 1600 & 400 & 500 \\
         Cross-Template & 1600 & 400 & 500 \\ \bottomrule
    \end{tabular}
\end{table}

\begin{figure}[h]
    \centering
    \begin{subfigure}[b]{0.2\textwidth}
        \centering
        \includegraphics[width=\linewidth]{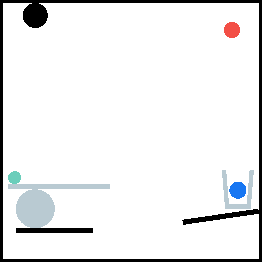}
        \caption{Cell:5, Radius:1}
    \end{subfigure}
    \begin{subfigure}[b]{0.2\textwidth}
        \centering
        \includegraphics[width=\linewidth]{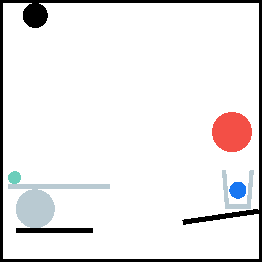}
        \caption{Cell:15, Radius:3}
    \end{subfigure}
    \begin{subfigure}[b]{0.2\textwidth}
        \centering
        \includegraphics[width=\linewidth]{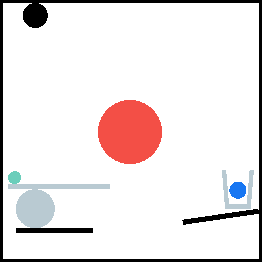}
        \caption{Cell:13, Radius:5}
    \end{subfigure}
    \caption{Examples of the discretized action space in PHYRE. Actions are defined by selecting a grid cell number for ball placement and specifying a discrete radius level.}
    \label{fig:PHYRE_action_example}
\end{figure}

\subsection{Conversion Details}

In PHYRE, instead of outputting continuous coordinates in the $x,y \in [0,1]$ range, we discretize the space into \{GRID\_SIZE[0] * GRID\_SIZE[1]\} equally divided grids (as shown in the last 3 images of Figure \ref{fig:PHYRE_sample}). The agentic VLM must select the center of a target grid cell for object placement and specify a discrete ball size level (1 to \{RADIUS\_LEVELS\}) (as shown in Figure \ref{fig:PHYRE_action_example}). 
In our experiments, we discretize the action space by defining a $5 \times 5$ grid for placement locations and 5 discrete levels for the ball's radius, resulting in a total of 125 possible actions. Action example: ``Cell: 17, Radius: 3".

\subsection{Prompts}

\begin{systempromptbox}{PHYRE: System Prompt}

You are a premier AI agent, an expert in solving physics-based puzzles. Your task is to analyze a given scene and determine the single optimal placement of a new object to achieve a specific goal.

\vspace{1em}

\textbf{Scene Legend:}

\quad \textbf{- Red Ball:} The dynamic ball you will place. 

\quad \textbf{- Green Ball:} The primary dynamic object you must manipulate. 

\quad \textbf{- Blue Ball:} A dynamic target. 

\quad \textbf{- Purple Ball:} A static (immovable) target. 

\quad \textbf{- Grey Ball:} Other dynamic objects. 

\quad \textbf{- Black Object:} Static obstacles.

\vspace{1em}

\textbf{Mission Objective:}\\
Strategically place a single red ball to trigger a chain reaction. The successful outcome is achieved when the green ball makes physical contact with either the blue ball or the purple ball.

\textbf{Action Parameters:}

You must define your action using two parameters:

\quad \textbf{- `Cell':} A grid cell number from 1 to \{GRID\_SIZE[0] * GRID\_SIZE[1]\}. 

\quad \textbf{- `Radius':} A size level from 1 (smallest, approx. 1/128 of image width) to \{RADIUS\_LEVELS\} (largest, approx. 1/8 of image width).

\vspace{1em}

\textbf{Placement Constraints:} \\
Your action is only valid if it meets these strict conditions: 

\quad \textbf{- No Collisions:} The red ball cannot overlap with any existing objects. 

\quad \textbf{- In Bounds:} The red ball must be placed entirely within the puzzle's boundaries.

\vspace{1em}

\textbf{Required Output Format} \\
Your response must be a single line of text, strictly conforming to the format below. Do not include any other words, notes, or explanations.

\vspace{1em}

\quad - `Cell: [NUMBER], Radius: [NUMBER]'

\quad - Example: `Cell: 13, Radius: 3'

\vspace{1em}

\textless initial\_image\textgreater: Image 1 (Initial Scene): The original puzzle scene. \\
\textless gridded\_image\textgreater: Image 2 (Gridded Scene): The scene with a \{GRID\_SIZE[0]\}x\{GRID\_SIZE[1]\} grid. You MUST use this gridded image to select your action.
\end{systempromptbox}

\begin{userpromptbox}{PHYRE: User Prompt}
    You have made previous attempts that failed. Here is a summary: \\

    (If the last action is valid.) \\
    Attempt \{past\_num\}: You chose (Cell: \{past\_cell\}, Radius: \{past\_radius\}), which FAILED to solve the puzzle. \\
    \textless keyframe\textgreater: from Attempt \{past\_num\} \\

    (If the last action is invalid.) \\
    Attempt \{past\_num\}: You chose (Cell: \{past\_cell\}, Radius: \{past\_radius\}), which was an INVALID ACTION (out of bounds or overlapping). \\
    
    Based on this complete history and the initial scene images, analyze your mistakes and propose a new, better action. \\
    You MUST propose a NEW action. DO NOT repeat any of the Cell, Radius combinations from the previous failed attempts. \\
    Look carefully at the gridded scene (`Image 2') for your placement. \\
    Remember, Radius is a level from 1 (very small) to \{RADIUS\_LEVELS\} (large). \\
    Your response MUST be ONLY in the format: `Cell: [NUMBER], Radius: [NUMBER]'.
\end{userpromptbox}

\subsection{Experiment Results}

\begin{table*}[h!]
\centering
\caption{Benchmark PHYRE. Model performance comparison across 10 attempts. Prompt Format: VLA.}
\label{tab:model_performance_comparison_phyre_vla}
\resizebox{\textwidth}{!}{
\begin{tabular}{l | lllll lllll | l}
\toprule

\textbf{Model} & \textbf{Att. 1} & \textbf{Att. 2} & \textbf{Att. 3} & \textbf{Att. 4} & \textbf{Att. 5} & \textbf{Att. 6} & \textbf{Att. 7} & \textbf{Att. 8} & \textbf{Att. 9} & \textbf{Att. 10} & \textbf{Ave. Att.} \\

\midrule

\texttt{\textit{MOCK}} & 1.12\% & 1.98\% & 2.70\% & 3.56\% & 4.44\% & 5.12\% & 5.74\% & 6.50\% & 7.08\% & 7.64\% & 5.00 \\

\midrule
\rowcolor{lightgray}\multicolumn{12}{c}{\textbf{Open-Source Models}} \\ \midrule

\rowcolor{mylightgray}\multicolumn{12}{c}{\textbf{Qwen2.5-VL Series}} \\
\textbf{Qwen-3B} & 1.90\% & 3.00\% & 3.72\% & 4.56\% & 5.50\% & 6.86\% & 7.68\% & 8.02\% & 8.20\% & 8.20\% & 3.98 \\
\textbf{Qwen-7B} & 0.60\% & 2.10\% & 4.10\% & 4.30\% & 5.50\% & 6.40\% & 7.16\% & 8.16\% & 8.82\% & 9.62\% & 5.10 \\
\textbf{Qwen-32B} & 2.34\% & 4.11\% & 5.48\% & 6.68\% & 7.24\% & 7.98\% & 8.52\% & 9.11\% & 9.81\% & 10.16\% & 3.97 \\
\textbf{Qwen-72B} & 1.38\% & 3.36\% & 4.64\% & 5.34\% & 6.52\% & 7.38\% & 8.40\% & 8.98\% & 9.52\% & 10.10\% & 4.48 \\

\rowcolor{lightgray}\multicolumn{12}{c}{\textbf{Closed-Source Models}} \\ \midrule

\rowcolor{mylightgray}\multicolumn{12}{c}{\textbf{Claude Series}} \\
\textbf{Claude 3.5 Sonnet} & 0.40\% & 1.64\% & 2.63\% & 3.76\% & 4.75\% & 5.55\% & 6.54\% & 7.29\% & 7.87\% & 8.97\% & 5.51 \\
\textbf{Claude 3.7 Sonnet} & 1.82\% & 3.36\% & 5.58\% & 7.56\% & 9.70\% & 11.16\% & 12.54\% & 13.78\% & 14.84\% & 15.64\% & 4.78 \\
\textbf{Claude 4.0 Sonnet} & 0.59\% & 1.36\% & 1.65\% & 1.87\% & 1.97\% & 2.11\% & 2.27\% & 2.37\% & 2.40\% & 2.45\% & 3.25 \\
\textbf{Claude 4.0 Opus} & 1.13\% & 2.87\% & 3.48\% & 4.44\% & 5.11\% & 5.24\% & 5.38\% & 6.33\% & 6.81\% & 7.16\% & 4.32 \\
\midrule

\rowcolor{mylightgray}\multicolumn{12}{c}{\textbf{Gemini Series}} \\
\textbf{Gemini-2.0-Flash} & 1.80\% & 3.49\% & 4.22\% & 5.04\% & 6.11\% & 6.75\% & 7.35\% & 7.86\% & 8.42\% & 8.84\% & 4.23 \\
\textbf{Gemini-2.5-Flash} & 1.06\% & 2.27\% & 3.36\% & 4.50\% & 5.45\% & 6.42\% & 7.13\% & 7.94\% & 8.56\% & 9.26\% & 4.96 \\
\textbf{Gemini-2.5-Pro} & 1.50\% & 3.56\% & 5.61\% & 7.70\% & 9.59\% & 11.42\% & 12.77\% & 14.23\% & 15.47\% & 16.37\% \secondlogo & 5.02 \\
\midrule

\rowcolor{mylightgray}\multicolumn{12}{c}{\textbf{GPT Series}} \\
\textbf{GPT-4V} & 1.92\% & 4.36\% & 6.30\% & 7.60\% & 8.54\% & 9.34\% & 10.34\% & 11.06\% & 11.88\% & 12.82\% & 4.40 \\
\textbf{GPT-4o-mini} & 2.62\% & 4.30\% & 4.86\% & 5.52\% & 6.32\% & 7.08\% & 7.64\% & 8.28\% & 9.10\% & 9.86\% & 4.40 \\
\textbf{GPT-4o} & 1.80\% &	4.40\% &	6.18\% &	7.16\% &	8.04\% &	9.14\% &	9.94\% &	10.60\% &	11.28\% &	11.88\% & 4.21 \\
\textbf{GPT-o3} & 2.87\% & 6.85\% & 9.99\% & 12.70\% & 14.89\% & 16.91\% & 18.76\% & 20.30\% & 21.58\% & 23.06\% \firstlogo & 4.58 \\
\textbf{GPT-o4-mini} & 0.92\% & 1.82\% & 2.62\% & 3.46\% & 4.36\% & 4.92\% & 5.50\% & 6.20\% & 6.90\% & 7.52\% & 5.13 \\

\bottomrule
\end{tabular}
}
\end{table*}

\begin{table*}[h!]
\centering
\caption{Benchmark PHYRE. Model performance comparison across 10 attempts. Prompt Format: WM.}
\label{tab:model_performance_comparison_phyre_wm}
\resizebox{\textwidth}{!}{
\begin{tabular}{l | lllll lllll | l}
\toprule

\textbf{Model} & \textbf{Att. 1} & \textbf{Att. 2} & \textbf{Att. 3} & \textbf{Att. 4} & \textbf{Att. 5} & \textbf{Att. 6} & \textbf{Att. 7} & \textbf{Att. 8} & \textbf{Att. 9} & \textbf{Att. 10} & \textbf{Ave. Att.} \\

\midrule

\texttt{\textit{MOCK}} & 1.12\% & 1.98\% & 2.70\% & 3.56\% & 4.44\% & 5.12\% & 5.74\% & 6.50\% & 7.08\% & 7.64\% & 5.00 \\

\midrule

\rowcolor{lightgray}\multicolumn{12}{c}{\textbf{Open-Source Models}} \\ \midrule

\rowcolor{mylightgray}\multicolumn{12}{c}{\textbf{Qwen2.5-VL Series}} \\
\textbf{Qwen-3B} & 1.42\% & 1.99\% & 2.29\% & 2.40\% & 2.45\% & 2.45\% & 2.45\% & 2.45\% & 2.45\% & 2.45\% & 1.72 \\
\textbf{Qwen-7B} & 0.60\% & 1.43\% & 2.57\% & 3.53\% & 3.93\% & 4.93\% & 5.47\% & 5.81\% & 5.93\% & 6.38\% & 4.70 \\
\textbf{Qwen-32B} & 3.32\% & 3.58\% & 3.98\% & 4.20\% & 4.62\% & 5.08\% & 5.58\% & 5.94\% & 6.38\% & 6.82\% & 3.76 \\
\textbf{Qwen-72B} & 1.86\% & 3.15\% & 4.53\% & 5.57\% & 6.46\% & 7.26\% & 7.80\% & 8.30\% & 8.78\% & 9.18\% & 4.12 \\

\rowcolor{lightgray}\multicolumn{12}{c}{\textbf{Closed-Source Models}} \\ \midrule

\rowcolor{mylightgray}\multicolumn{12}{c}{\textbf{Claude Series}} \\
\textbf{Claude 3.5 Sonnet} & 1.18\% & 2.97\% & 4.12\% & 4.88\% & 5.83\% & 6.43\% & 6.84\% & 7.37\% & 7.86\% & 8.25\% & 4.25 \\
\textbf{Claude 3.7 Sonnet} & 1.82\% & 3.67\% & 5.19\% & 6.47\% & 7.72\% & 8.69\% & 10.00\% & 10.82\% & 11.86\% & 12.82\% & 4.85 \\
\textbf{Claude 4.0 Sonnet} & 1.93\% & 3.81\% & 5.08\% & 6.43\% & 7.59\% & 8.76\% & 9.77\% & 10.74\% & 11.83\% & 12.86\% & 4.88 \\
\textbf{Claude 4.0 Opus} & 3.35\% & 5.19\% & 6.19\% & 7.51\% & 8.59\% & 9.85\% & 10.83\% & 11.69\% & 12.51\% & 13.33\% & 4.32 \\
\midrule

\rowcolor{mylightgray}\multicolumn{12}{c}{\textbf{Gemini Series}} \\
\textbf{Gemini-2.0-Flash} & 1.16\% & 2.10\% & 3.42\% & 4.36\% & 5.24\% & 5.90\% & 6.72\% & 7.30\% & 7.92\% & 8.54\% & 4.84 \\
\textbf{Gemini-2.5-Flash} & 1.45\% & 3.45\% & 4.95\% & 6.05\% & 7.00\% & 7.73\% & 8.54\% & 9.22\% & 9.89\% & 10.40\% & 4.41 \\
\textbf{Gemini-2.5-Pro} & 1.95\% & 4.42\% & 6.77\% & 8.77\% & 10.63\% & 11.90\% & 13.33\% & 14.30\% & 14.98\% & 15.92\% \secondlogo & 4.53 \\
\midrule

\rowcolor{mylightgray}\multicolumn{12}{c}{\textbf{GPT Series}} \\
\textbf{GPT-4V} & 1.80\% & 3.27\% & 4.67\% & 5.70\% & 6.80\% & 7.75\% & 8.78\% & 9.60\% & 10.40\% & 11.07\% & 4.67 \\
\textbf{GPT-4o-mini} & 1.59\% & 3.14\% & 4.07\% & 4.58\% & 5.10\% & 5.97\% & 6.74\% & 7.80\% & 8.35\% & 8.90\% & 4.79 \\
\textbf{GPT-4o} & 1.10\% & 2.04\% & 2.85\% & 3.85\% & 4.59\% & 5.37\% & 6.35\% & 7.03\% & 7.65\% & 8.40\% & 5.14 \\
\textbf{GPT-o3} & 2.70\% & 6.70\% & 9.30\% & 11.40\% & 13.90\% & 16.30\% & 17.70\% & 19.00\% & 20.10\% & 21.40\% \firstlogo & 4.53\\
\textbf{GPT-o4-mini} & 1.22\% & 2.32\% & 3.16\% & 4.28\% & 5.56\% & 6.34\% & 7.02\% & 7.90\% & 8.72\% & 9.72\% & 5.20 \\

\bottomrule
\end{tabular}
}
\end{table*}

\begin{table}[h!]
\centering
\caption{Benchmark PHYRE. Model performance Pass@K Comparison across 10 attempts. Prompt Format: VLA.}
\label{tab:model_performance_comparison_phyre_vla_passk}
\resizebox{0.6\linewidth}{!}{
\begin{tabular}{l | ccccc}
\toprule

\textbf{Model} & \textbf{Pass@1} & \textbf{Pass@2} & \textbf{Pass@3} & \textbf{Pass@4} & \textbf{Pass@5} \\

\midrule

\texttt{\textit{{MOCK}}}& 7.80\% & 13.60\% & 17.70\% & 20.00\% & 22.70\% \\

\midrule

\rowcolor{lightgray}\multicolumn{6}{c}{\textbf{Open-Source Models}} \\ \midrule

\rowcolor{mylightgray}\multicolumn{6}{c}{\textbf{Qwen2.5-VL Series}} \\
\textbf{Qwen-3B} & 8.2\% & 8.2\% & 8.2\% & 8.2\% & 8.2\%\\
\textbf{Qwen-7B} & 9.6\% & 9.6\% & 9.7\% & 9.7\% & 9.7\%\\
\textbf{Qwen-32B} & 9.4\% & 13.5\% & 13.9\% & 14.9\% & 15.7\%\\
\textbf{Qwen-72B} & 10.0\% & 11.2\% & 12.3\% & 13.0\% & 13.6\%\\

\rowcolor{lightgray}\multicolumn{6}{c}{\textbf{Closed-Source Models}} \\ \midrule

\rowcolor{mylightgray}\multicolumn{6}{c}{\textbf{Claude Series}} \\
\textbf{Claude 3.5 Sonnet} & 8.6\% & 13.7\% & 16.3\% & 18.2\% & 18.9\%\\
\textbf{Claude 3.7 Sonnet} & 14.9\% & 19.2\% & 20.7\% & 22.4\% & 23.5\%\\
\textbf{Claude 4.0 Sonnet} & 2.1\% & 2.8\% & 3.8\% & 4.2\% & 4.5\%\\
\textbf{Claude 4.0 Opus} & 7.4\% & 8.7\% & 9.8\% & 10.9\% & 11.2\%\\
\midrule

\rowcolor{mylightgray}\multicolumn{6}{c}{\textbf{Gemini Series}} \\
\textbf{Gemini-2.0-Flash} & 8.3\% & 12.7\% & 14.3\% & 15.4\% & 17.7\%\\
\textbf{Gemini-2.5-Flash} & 8.8\% & 12.5\% & 15.1\% & 17.1\% & 18.5\%\\
\textbf{Gemini-2.5-Pro} & 17.2\% & 21.5\% & 24.6\% & 26.1\% & 27.0\% \secondlogo\\

\midrule

\rowcolor{mylightgray}\multicolumn{6}{c}{\textbf{GPT Series}} \\
\textbf{GPT-4V} & 12.0\% & 15.6\% & 16.7\% & 17.8\% & 18.9\%\\
\textbf{GPT-4o-mini} & 10.6\% & 12.1\% & 13.0\% & 13.4\% & 13.8\%\\
\textbf{GPT-4o} & 11.2\% & 15.5\% & 16.9\% & 17.7\% & 18.2\%\\
\textbf{GPT-o3} & 22.5\% & 28.1\% & 30.6\% & 32.3\% & 33.5\% \firstlogo\\
\textbf{GPT-o4-mini} & 7.0\% & 10.9\% & 13.5\% & 14.8\% & 15.5\%\\
\bottomrule
\end{tabular}
}
\end{table}

\begin{table}[h!]
\centering
\caption{Benchmark PHYRE. Model performance Pass@K Comparison across 10 attempts. Prompt Format: WM.}
\label{tab:model_performance_comparison_phyre_wm_passk}
\resizebox{0.6\linewidth}{!}{
\begin{tabular}{l | lllll}
\toprule

\textbf{Model} & \textbf{Pass@1} & \textbf{Pass@2} & \textbf{Pass@3} & \textbf{Pass@4} & \textbf{Pass@5} \\

\midrule

\texttt{\textit{MOCK}} & 7.80\% & 13.60\% & 17.70\% & 20.00\% & 22.70\% \\

\midrule

\rowcolor{lightgray}\multicolumn{6}{c}{\textbf{Open-Source Models}} \\ \midrule

\rowcolor{mylightgray}\multicolumn{6}{c}{\textbf{Qwen2.5-VL Series}} \\
\textbf{Qwen-3B} & 2.6\% & 2.6\% & 2.6\% & 2.6\% & 2.6\%\\
\textbf{Qwen-7B} & 6.4\% & 6.4\% & 6.6\% & 6.8\% & 6.8\%\\
\textbf{Qwen-32B} & 7.6\% & 9.2\% & 10.3\% & 11.7\% & 13.3\%\\
\textbf{Qwen-72B} & 8.7\% & 10.6\% & 11.6\% & 12.0\% & 12.9\%\\

\rowcolor{lightgray}\multicolumn{6}{c}{\textbf{Closed-Source Models}} \\ \midrule

\rowcolor{mylightgray}\multicolumn{6}{c}{\textbf{Claude Series}} \\
\textbf{Claude 3.5 Sonnet} & 8.8\% & 12.2\% & 13.8\% & 15.9\% & 17.0\%\\
\textbf{Claude 3.7 Sonnet} & 12.8\% & 16.8\% & 19.1\% & 20.4\% & 22.2\%\\
\textbf{Claude 4.0 Sonnet} & 11.1\% & 17.6\% & 20.9\% & 23.0\% & 24.3\%\\
\textbf{Claude 4.0 Opus} & 13.8\% & 18.7\% & 21.8\% & 24.2\% & 25.8\%\\
\midrule

\rowcolor{mylightgray}\multicolumn{6}{c}{\textbf{Gemini Series}} \\
\textbf{Gemini-2.0-Flash} & 9.2\% & 11.7\% & 14.9\% & 15.6\% & 17.0\% \\
\textbf{Gemini-2.5-Pro} & 15.9\% & 21.2\% & 23.9\% & 25.8\% & 27.6\%\secondlogo\\
\textbf{Gemini-2.5-Flash} & 10.9\% & 15.3\% & 17.0\% & 18.7\% & 20.1\%\\
\midrule

\rowcolor{mylightgray}\multicolumn{6}{c}{\textbf{GPT Series}} \\
\textbf{GPT-4V} & 11.1\% & 14.5\% & 17.7\% & 19.9\% & 21.2\%\\
\textbf{GPT-4o-mini} & 9.3\% & 11.6\% & 13.0\% & 14.2\% & 14.4\%\\
\textbf{GPT-4o} & 8.6\% & 12.4\% & 14.3\% & 16.0\% & 17.4\%\\
\textbf{GPT-o3} & 20.2\% & 26.6\% & 28.8\% & 29.8\% & 30.2\% \firstlogo \\
\textbf{GPT-o4-mini} & 8.8\% & 13.5\% & 16.1\% & 19.4\% & 20.6\%\\
\bottomrule
\end{tabular}
}
\end{table}

Across all tested models, the overall success rate is remarkably low, underscoring the difficulty of the PHYRE tasks. Even the top-performing model, GPT-o3 (Prompt Format: VLA), only achieves a final success rate of 23.1\% after 10 attempts.  Open-source models (e.g., Qwen) often perform near or below the \texttt{\textit{MOCK}} baseline, further emphasizing this limitation.

Moreover, most models score below 4\% on their first attempt (Pass@1) (see Tables \ref{tab:model_performance_comparison_phyre_vla_passk} \& \ref{tab:model_performance_comparison_phyre_wm_passk}). This indicates a profound weakness in in-advance planning, where models must devise a complete and correct solution from a single static observation without any prior interaction history. Models largely fail to utilize internalized physical principles of the environment to make a successful initial prediction.

Performance improves marginally with more attempts, but learning efficiency is low: Successful trials require 4–5 attempts on average (Figure \ref{fig:PHYRE_IPHYRE}). This implies that the feedback from a failed trajectory (i.e., seeing the outcome of a wrong action) is not sufficient for models to build an accurate and predictive internal world model that can guide subsequent decisions. 

A counter-intuitive finding emerges from prompt comparisons: The WM strategy consistently underperforms the simpler VLA approach across most models (Figure \ref{fig:PHYRE_IPHYRE_VLA_WM}(a)). While WM accelerates early progress, its gains diminish in later attempts. Most data points fall below the diagonal line of equivalence. This finding points to a fundamental disconnect: even if a model can describe a potential outcome, this descriptive knowledge does not translate into improved procedural control. This aligns with our broader conclusion that current VLMs possess only a descriptive, rather than a predictive and controllable, understanding of physics.

\section{Benchmark: I-PHYRE} \label{appendix:i-phyre}

The I-PHYRE\footnote{{https://github.com/lishiqianhugh/IPHYRE}} benchmark comprises 40 interactive physics scenarios, each requiring an agent to direct red balls into the abyss. Its sole interaction mechanism is the removal of designated gray blocks at exact moments in time. The different task environments feature a diverse set of objects to foster complex physical reasoning: static black blocks serve as fixed obstacles, while dynamic blue blocks respond to gravity and collisions. To further elevate the complexity, elements such as springs and rigid sticks are introduced, enriching the physical dynamics and ensuring that successful task completion requires more than trivial planning.

\subsection{Conversion Details}

\begin{figure*}[h]
    \centering
    \begin{subfigure}[b]{0.19\textwidth}
        \centering
        \includegraphics[width=\linewidth]{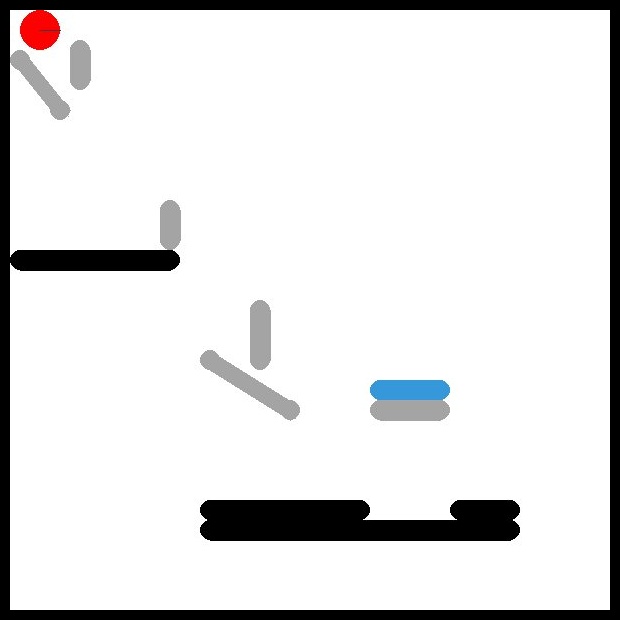}
    \end{subfigure}
    \begin{subfigure}[b]{0.19\textwidth}
        \centering
        \includegraphics[width=\linewidth]{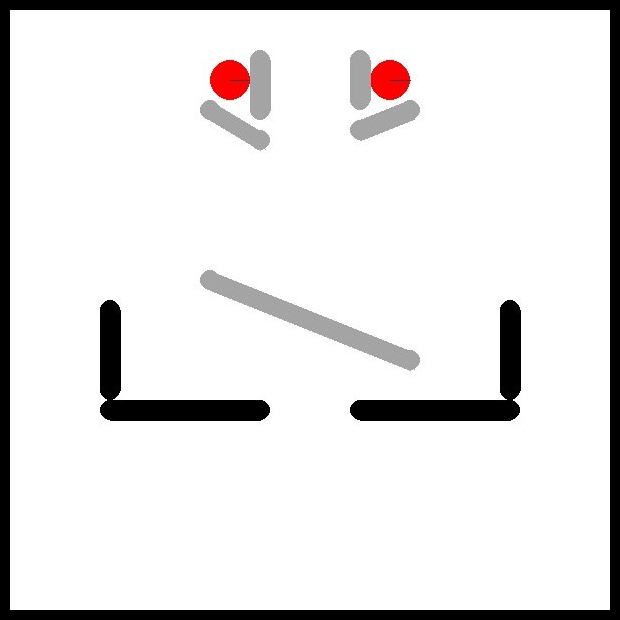}
    \end{subfigure}
    \begin{subfigure}[b]{0.19\textwidth}
        \centering
        \includegraphics[width=\linewidth]{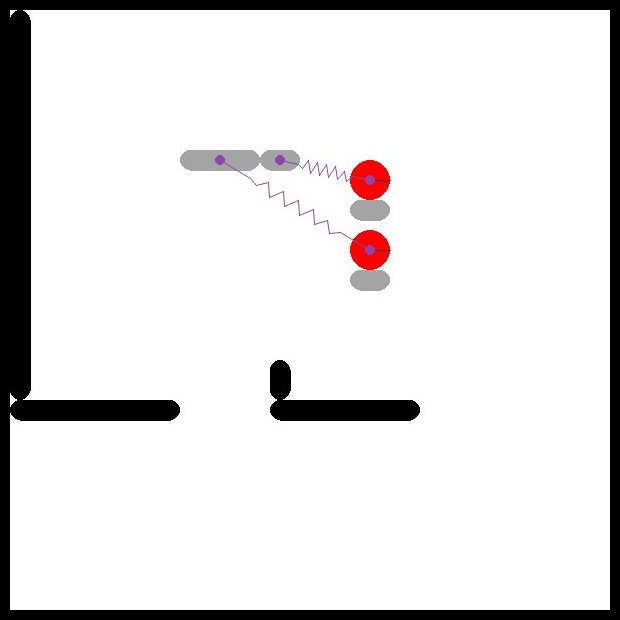}
    \end{subfigure}
    \begin{subfigure}[b]{0.19\textwidth}
        \centering
        \includegraphics[width=\linewidth]{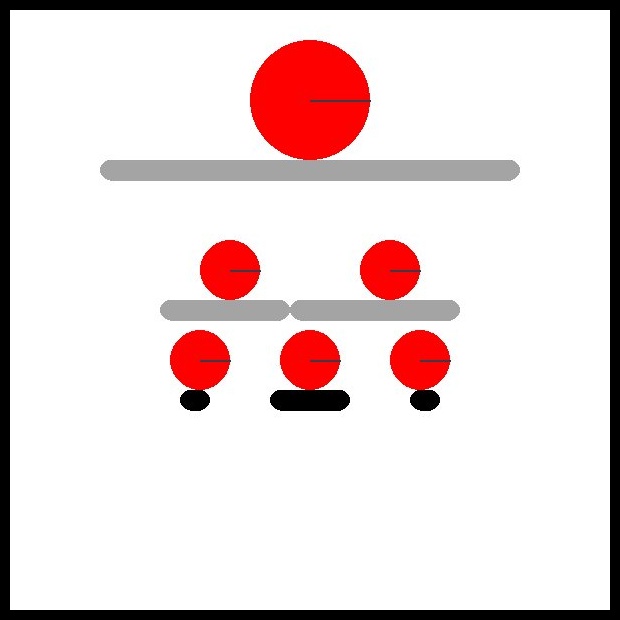}
    \end{subfigure}
    \begin{subfigure}[b]{0.19\textwidth}
        \centering
        \includegraphics[width=\linewidth]{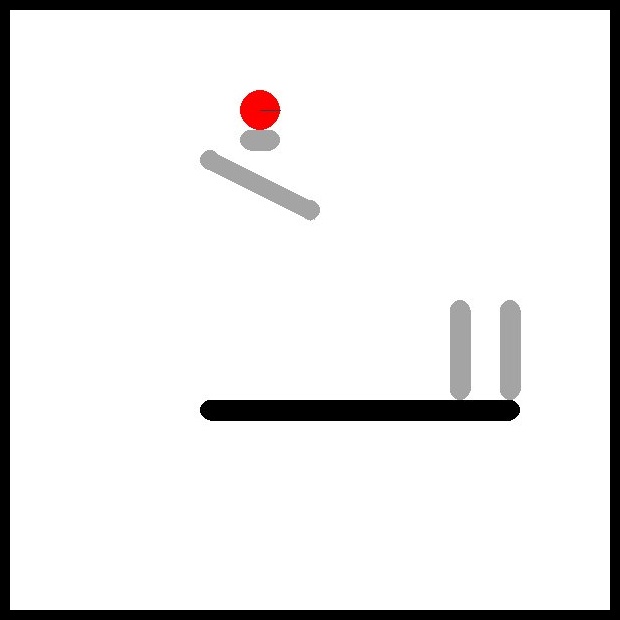}
    \end{subfigure}
    \begin{subfigure}[b]{0.19\textwidth}
        \centering
        \includegraphics[width=\linewidth]{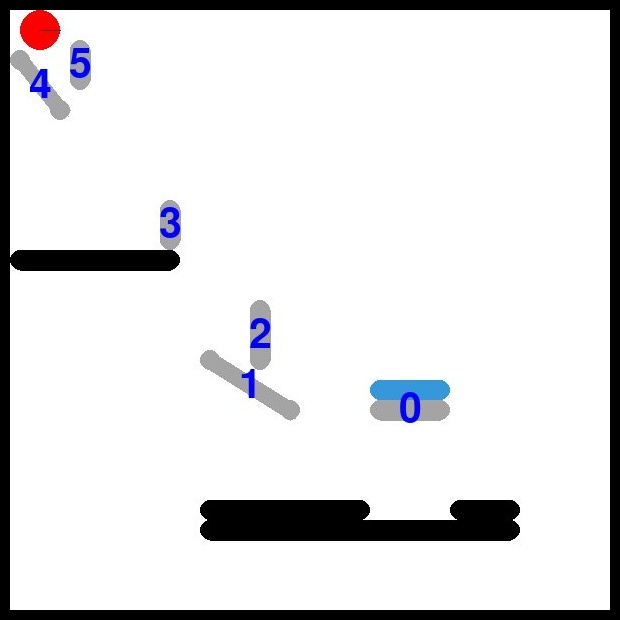}
        \caption{Hinder Fill}
    \end{subfigure}
    \begin{subfigure}[b]{0.19\textwidth}
        \centering
        \includegraphics[width=\linewidth]{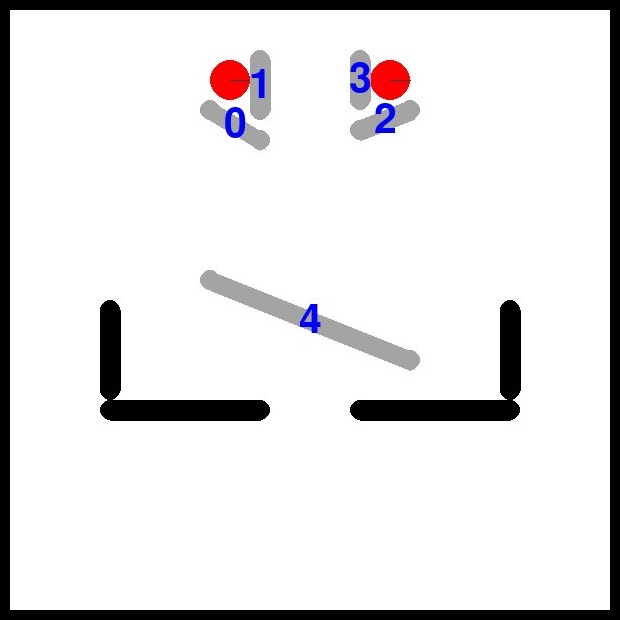}
        \caption{Multi Ball Hole}
    \end{subfigure}
    \begin{subfigure}[b]{0.19\textwidth}
        \centering
        \includegraphics[width=\linewidth]{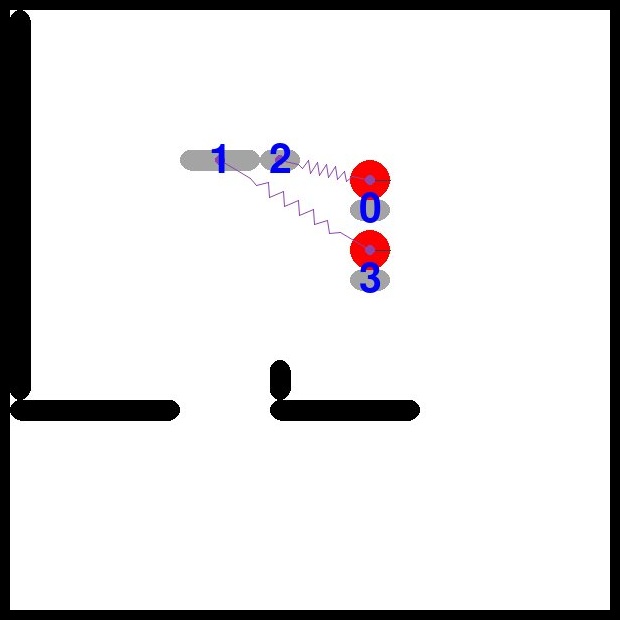}
        \caption{Multi Ball Spring}
    \end{subfigure}
    \begin{subfigure}[b]{0.19\textwidth}
        \centering
        \includegraphics[width=\linewidth]{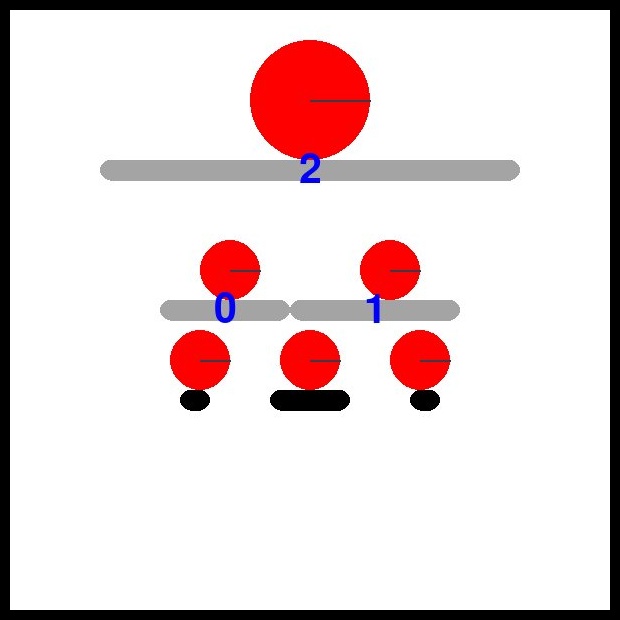}
        \caption{Multi Ball Stack}
    \end{subfigure}
    \begin{subfigure}[b]{0.19\textwidth}
        \centering
        \includegraphics[width=\linewidth]{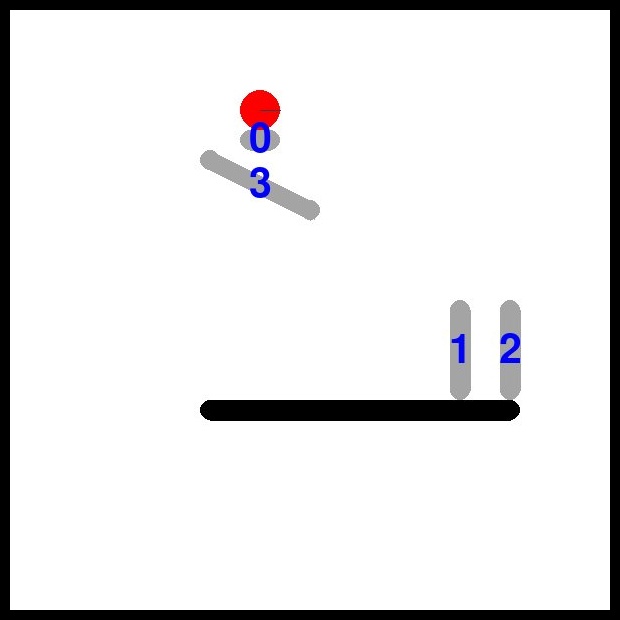}
        \caption{Support Hinder}
    \end{subfigure}
    \caption{IPHYRE Benchmark Observation Space. Top: Original scene renderings. Bottom: Indexed versions enabling precise object reference for VLMs.}
    \label{fig:IPHYRE_observation_space}
\end{figure*}

While I-PHYRE \cite{iphyre} was originally evaluated on GPT-4 without images, by using symbolic matrix representations (as shown in the orange box), our implementation incorporates visual inputs. We augment the raw I-PHYRE images with numerical indices to facilitate object selection by VLMs. Figure \ref{fig:IPHYRE_observation_space} illustrates this modification, with the upper row showing the original rendered scenes and the lower row displaying their indexed counterparts.

Consistent with the original methodology, we employ \textit{Planning in Advance} for evaluation, where each turn permits a maximum of 10 attempts.

\begin{examplebox}{I-PHYRE: Initial Observation Space (Plain text)}
You are given a simulated square-shaped 2D world of size 600*600 consisting of some objects. \\

The object configuration array is as follows (blocks and balls are all objects): \\

\textbf{For blocks:}

\quad - [x1, y1, x2, y2, radius, eli, dynamic, joint, spring]

\quad - [x1, y1] indicates the left end point and [x2, y2] indicates the right end point. The height (twice the `radius') of blocks is 20.\\
\textbf{For balls:}

\quad - [x, y, x, y, radius, eli, dynamic, joint, spring]

\quad - [x, y] indicates the center of the ball. `radius' is the radius of the ball.

\textbf{eli:}

\quad 0/1 indicates whether the corresponding object can be eliminated.

\quad 1 is eliminable and 0 is not eliminable.

\textbf{dynamic:}

\quad 0/1 indicates whether the corresponding object can move by external force. 

\quad 1 is dynamic and 0 is static.\\
\textbf{joint:}

\quad 0/1 indicates whether the corresponding object is connected to a stick. 

\quad 1 is connected and 0 is not connected.

\textbf{spring:}

\quad 0/1 indicates whether the corresponding object is connected to a spring. 

\quad 1 is connected and 0 is not connected.\\

Given the specific object configuration dict below:\\

[[100. 400. 400. 400.  10.   0.   0.   0.   0.] 

\ [200. 350. 210. 350.  10.   1.   0.   0.   0.] 
 
\ [200. 300. 210. 300.  10.   1.   0.   0.   0.] 
 
\ [480. 400. 550. 400.  10.   0.   0.   0.   0.] 
 
\ [390. 350. 400. 350.  10.   1.   0.   0.   0.] 
 
\ [390. 300. 400. 300.  10.   1.   0.   0.   0.] 
 
\ [100. 380. 100. 360.  10.   0.   0.   0.   0.] 
 
\ [550. 380. 550. 360.  10.   0.   0.   0.   0.] 
 
\ [200. 280. 400. 280.  10.   0.   1.   0.   0.] 
 
\ [260. 250. 260. 250.  20.   0.   1.   0.   0.] 
 
\ [340. 250. 340. 250.  20.   0.   1.   0.   0.] 
 
\ [\ \ \ 0.   \ \ \ 0.   \ \ 0.   \ \ \ 0.   \ \ 0.   0.   0.   0.   0.]] 
\end{examplebox}

\subsection{Prompts}

\begin{systempromptbox}{I-PHYRE: System Prompt}
You are an expert in 2D physics puzzle solving specializing in trajectory manipulation through strategic block elimination. Your objective is to guide the red ball(s) into the abyss by precisely timing the removal of designated gray blocks (the only eliminable objects).

\textbf{Game Environment:}\\
\quad 1. Core Mechanics:

\quad \quad - Goal: Get all red balls into the abyss

\quad \quad - Only gray blocks can be eliminated

\quad \quad - Static blocks (gray/black) form fixed structures

\quad \quad - Dynamic blue blocks respond to gravity/physics (e.g. Spring mechanisms, Rigid stick constraints)\\

\textbf{Action Requirements:}

\quad 1. Generate a sequence of elimination actions as a JSON array

\quad 2. Each action must specify:

\quad \quad    - `time': Execution Timestamp (0.0-15.0 seconds)

\quad \quad    - `index': Target block's configuration array position (0-\{max\_index\})

\quad 3. Verify block numbers against the frame before selection

\quad 4. \{num\_blocks\} gray blocks are currently eligible for elimination

\textbf{Output Rules:}

\quad Absolute adherence to JSON syntax

\quad Time precision to one decimal place recommended

\textbf{Output Format:}

\`\ \`\ \`\  json

[

\quad \{``time": 0.5, ``index": 2 \},

\quad \{``time": 2.1, ``index": 0 \}
    
]

\`\ \`\ \`\  
\end{systempromptbox}

\begin{userpromptbox}{I-PHYRE: User Prompt}
Analyze the initial scene configuration and frame to develop an optimal block elimination sequence. Consider: \\

1. Ball's initial trajectory \\
2. Block removal timing consequences \\ 
3. Chain reactions from each elimination \\
4. Physical constraints of the environment \\

\textless initial\_image\textgreater 

(After 1st attempt:) \\

\textbf{Previous Attempt Analysis:} \\
\textless keyframe\textgreater: from Attempt \{past\_num\}\\

\textbf{Failure Diagnosis:}\\
1. Timing Issues: \\
   - Early removals causing [specific effects] \\
   - Late removals resulting in [undesired outcomes] \\
2. Sequencing Errors: \\
   - Incorrect block priority \\
   - Missed chain reaction opportunities \\
3. Physical Miscalculations: \\
   - Trajectory inaccuracies \\
   - Collision mispredictions

\textbf{Revised Strategy Requirements:} \\
1. Avoid all previously failed approaches \\
2. Incorporate kinematic insights from failed attempts \\
3. Optimize for minimal actions \\
4. Account for newly observed physical behaviors \\

Propose a refined solution incorporating these lessons, the improved action sequence is:
\end{userpromptbox}

\subsection{Experiment Results}

\begin{table*}[h!]
\centering
\caption{Benchmark I-PHYRE. Model performance comparison across 10 attempts. Prompt Format: VLA.}
\label{tab:iphyre_model_performance_vla_avg_step}
\resizebox{\textwidth}{!}{
\begin{tabular}{l | llllllllll | l} 
\toprule

\textbf{Model} & \textbf{Att. 1} & \textbf{Att. 2} & \textbf{Att. 3} & \textbf{Att. 4} & \textbf{Att. 5} & \textbf{Att. 6} & \textbf{Att. 7} & \textbf{Att. 8} & \textbf{Att. 9} & \textbf{Att. 10} & \textbf{Avg. Att.} \\

\midrule
\texttt{\textit{MOCK}} & 20.50\% & 30.00\% & 38.50\% & 43.50\% & 50.00\% & 55.00\% & 57.00\% & 58.50\% & 59.50\% & 62.50\% & 3.67 \\
\midrule

\rowcolor{lightgray}\multicolumn{12}{c}{\textbf{Open-Source Models}} \\ \midrule

\rowcolor{mylightgray}\multicolumn{12}{c}{\textbf{Qwen2.5-VL Series}} \\
\textbf{Qwen-3B} & 35.83\% & 38.33\% & 38.33\% & 38.33\% & 38.33\% & 38.33\% & 38.33\% & 38.33\% & 38.33\% & 38.33\% & 2.87 \\
\textbf{Qwen-7B} & 12.50\% & 20.00\% & 20.00\% & 20.00\% & 20.00\% & 20.00\% & 20.00\% & 20.00\% & 20.00\% & 20.00\% & 1.75 \secondlogo \\
\textbf{Qwen-32B} & 04.17\% & 05.83\% & 07.50\% & 10.83\% & 13.33\% & 14.17\% & 15.00\% & 15.83\% & 16.67\% & 17.50\% & 1.48 \firstlogo \\
\textbf{Qwen-72B} & 12.50\% & 30.00\% & 31.67\% & 33.33\% & 37.50\% & 40.00\% & 40.00\% & 45.00\% & 45.00\% & 45.00\% & 2.00 \\
        
\rowcolor{lightgray}\multicolumn{12}{c}{\textbf{Closed-Source Models}} \\ \midrule

\rowcolor{mylightgray}\multicolumn{12}{c}{\textbf{Claude Series}} \\
\textbf{Claude 3.5 Sonnet} & 26.67\% & 39.17\% & 45.83\% & 46.67\% & 46.67\% & 46.67\% & 47.50\% & 50.83\% & 52.50\% & 52.50\% & 2.73 \\
\textbf{Claude 3.7 Sonnet} & 22.50\% & 38.33\% & 48.33\% & 52.50\% & 52.50\% & 52.50\% & 58.33\% & 63.33\% & 65.00\% & 66.67\% & 2.85 \\
\textbf{Claude 4.0 Sonnet} & 29.17\% & 40.83\% & 45.00\% & 49.17\% & 51.67\% & 55.83\% & 61.67\% & 63.33\% & 65.83\% & 67.50\% & 2.64 \\
\textbf{Claude 4.0 Opus} & 33.33\% & 43.33\% & 47.50\% & 54.17\% & 60.00\% & 64.17\% & 66.67\% & 69.17\% & 70.00\% & 71.67\% & 2.79 \\
\midrule
        
\rowcolor{mylightgray}\multicolumn{12}{c}{\textbf{Gemini Series}} \\
\textbf{Gemini-2.0-Flash} & 22.50\% & 28.33\% & 29.17\% & 29.17\% & 33.33\% & 35.00\% & 36.67\% & 40.83\% & 43.33\% & 43.33\% & 2.46 \\
\textbf{Gemini-2.5-Flash} & 16.67\% & 25.00\% & 30.83\% & 31.67\% & 35.00\% & 37.50\% & 38.33\% & 40.83\% & 45.00\% & 45.00\% & 2.44 \\
\textbf{Gemini-2.5-Pro} & 25.83\% & 36.67\% & 40.83\% & 47.50\% & 50.00\% & 50.83\% & 55.00\% & 55.83\% & 55.83\% & 58.33\% & 2.41 \\
\midrule

\rowcolor{mylightgray}\multicolumn{12}{c}{\textbf{GPT Series}} \\
\textbf{GPT-4V} & 23.33\% & 36.67\% & 40.00\% & 41.67\% & 44.17\% & 46.67\% & 46.67\% & 46.67\% & 46.67\% & 46.67\% & 2.71 \\
\textbf{GPT-4o-mini} & 16.67\% & 29.17\% & 34.17\% & 36.67\% & 38.33\% & 40.00\% & 43.33\% & 45.00\% & 45.00\% & 45.00\% & 2.33 \\
\textbf{GPT-4o} & 22.50\% & 38.33\% & 43.33\% & 45.83\% & 48.33\% & 49.17\% & 50.83\% & 53.33\% & 53.33\% & 53.33\% & 2.61 \\
\textbf{GPT-o3} & 39.17\% & 54.17\% & 64.17\% & 70.00\% & 75.00\% & 75.83\% & 79.17\% & 80.83\% & 80.83\% & 81.67\% \firstlogo & 2.84  \\
\textbf{GPT-o4-mini} & 34.17\% & 42.50\% & 51.67\% & 55.83\% & 60.00\% & 65.83\% & 70.00\% & 70.00\% & 72.50\% & 75.00\% \secondlogo & 2.87  \\
\bottomrule
\end{tabular}
}
\end{table*}

\begin{table*}[h!]
\centering
\caption{Benchmark I-PHYRE. Model performance comparison across 10 attempts. Prompt Format: WM.}
\label{tab:iphyre_model_performance_wm_avg_step}
\resizebox{\textwidth}{!}{
\begin{tabular}{l | llllllllll | l} 

\toprule

\textbf{Model} & \textbf{Att. 1} & \textbf{Att. 2} & \textbf{Att. 3} & \textbf{Att. 4} & \textbf{Att. 5} & \textbf{Att. 6} & \textbf{Att. 7} & \textbf{Att. 8} & \textbf{Att. 9} & \textbf{Att. 10} & \textbf{Avg. Att.} \\

\midrule
\texttt{\textit{MOCK}} & 20.50\% & 30.00\% & 38.50\% & 43.50\% & 50.00\% & 55.00\% & 57.00\% & 58.50\% & 59.50\% & 62.50\% & 3.67 \\
\midrule

\rowcolor{lightgray}\multicolumn{12}{c}{\textbf{Open-Source Models}} \\ \midrule

\rowcolor{mylightgray}\multicolumn{12}{c}{\textbf{Qwen2.5-VL Series}} \\
\textbf{Qwen-3B} & 15.00\% & 15.00\% & 15.00\% & 15.00\% & 15.00\% & 15.00\% & 15.00\% & 15.00\% & 15.00\% & 15.00\% & 2.00 \\
\textbf{Qwen-7B} & 27.50\% & 27.50\% & 27.50\% & 32.50\% & 32.50\% & 32.50\% & 32.50\% & 32.50\% & 32.50\% & 32.50\% & 2.00 \\
\textbf{Qwen-32B} & 00.83\% & 00.83\% & 00.83\% & 00.83\% & 02.50\% & 04.17\% & 04.17\% & 05.00\% & 05.83\% & 05.83\% & 1.71 \secondlogo \\
\textbf{Qwen-72B} & 12.50\% & 19.17\% & 19.17\% & 20.00\% & 20.83\% & 24.17\% & 25.00\% & 26.67\% & 27.50\% & 29.17\% & 1.60 \firstlogo \\
        
\rowcolor{lightgray}\multicolumn{12}{c}{\textbf{Closed-Source Models}} \\ \midrule

\rowcolor{mylightgray}\multicolumn{12}{c}{\textbf{Claude Series}} \\
\textbf{Claude 3.5 Sonnet} & 15.83\% & 26.67\% & 31.67\% & 39.17\% & 39.17\% & 39.17\% & 40.83\% & 44.17\% & 45.00\% & 46.67\% & 2.20 \\
\textbf{Claude 3.7 Sonnet} & 15.29\% & 32.94\% & 38.82\% & 40.00\% & 40.00\% & 40.00\% & 44.71\% & 49.41\% & 51.76\% & 54.12\% & 2.48 \\
\textbf{Claude 4.0 Sonnet} & 18.33\% & 26.67\% & 33.33\% & 41.67\% & 45.00\% & 46.67\% & 50.00\% & 50.83\% & 53.33\% & 55.00\% & 2.08 \\
\textbf{Claude 4.0 Opus} & 14.29\% & 24.37\% & 31.93\% & 33.61\% & 36.97\% & 38.66\% & 43.70\% & 46.22\% & 48.74\% & 51.26\% & 2.18 \\
\midrule
        
\rowcolor{mylightgray}\multicolumn{12}{c}{\textbf{Gemini Series}} \\
\textbf{Gemini-2.0-Flash} & 15.00\% & 28.33\% & 31.67\% & 32.50\% & 36.67\% & 36.67\% & 37.50\% & 37.50\% & 38.33\% & 39.17\% & 1.98 \\
\textbf{Gemini-2.5-Flash} & 21.67\% & 29.17\% & 33.33\% & 35.83\% & 40.83\% & 44.17\% & 45.00\% & 49.17\% & 50.00\% & 50.83\% & 2.49 \\
\textbf{Gemini-2.5-Pro} & 27.50\% & 39.17\% & 45.00\% & 51.67\% & 56.67\% & 60.83\% & 62.50\% & 63.33\% & 65.83\% & 67.50\% & 2.48 \\
\midrule

\rowcolor{mylightgray}\multicolumn{12}{c}{\textbf{GPT Series}} \\
\textbf{GPT-4V} & 20.00\% & 31.67\% & 36.67\% & 41.67\% & 41.67\% & 43.33\% & 45.83\% & 45.83\% & 45.83\% & 45.83\% & 2.22 \\
\textbf{GPT-4o-mini} & 14.17\% & 23.33\% & 27.50\% & 30.00\% & 32.50\% & 32.50\% & 34.17\% & 34.17\% & 34.17\% & 34.17\% & 2.05 \\
\textbf{GPT-4o} & 15.00\% & 29.17\% & 35.83\% & 36.67\% & 40.83\% & 43.33\% & 45.00\% & 45.83\% & 45.83\% & 45.83\% & 2.11 \\
\textbf{GPT-o3} & 35.83\% & 50.83\% & 56.67\% & 62.50\% & 70.00\% & 72.50\% & 74.17\% & 75.83\% & 75.83\% & 76.67\% \firstlogo & 2.79\\
\textbf{GPT-o4-mini} & 25.00\% & 37.50\% & 45.00\% & 52.50\% & 58.33\% & 61.67\% & 64.17\% & 66.67\% & 70.00\% & 70.83\% \secondlogo & 2.80 \\
\bottomrule
\end{tabular}
}
\end{table*}

\begin{table}[ht]
\centering
\caption{Benchmark I-PHYRE. Model performance comparison across 10 attempts Pass@K (\%).}
\label{tab:iphyre_pass_at_k}
\resizebox{0.75\linewidth}{!}{%
\begin{tabular}{l|lll|lll}
\toprule
& \multicolumn{3}{c}{\textbf{VLA}} & \multicolumn{3}{c}{\textbf{WM}} \\
\cmidrule(lr){2-4} \cmidrule(lr){5-7}
\textbf{Model} & \textbf{Pass@1} & \textbf{Pass@2} & \textbf{Pass@3} & \textbf{Pass@1} & \textbf{Pass@2} & \textbf{Pass@3} \\
\midrule
\texttt{\textit{MOCK}} & 57.50 & 72.50 & 82.50 & 57.50 & 72.50 & 82.50 \\
\midrule
\rowcolor{lightgray}\multicolumn{7}{c}{\textbf{Open-Source Models}} \\ \midrule
\rowcolor{mylightgray}\multicolumn{7}{c}{\textbf{Qwen2.5-VL Series}} \\
\textbf{Qwen-3B} & 37.50 & 37.50 & 40.00 & 15.00 & 15.00 & 15.00 \\
\textbf{Qwen-7B} & 20.00 & 20.00 & 20.00 & 32.50 & 32.50 & 32.50 \\
\textbf{Qwen-32B} & 20.00 & 25.00 & 32.50 & 7.50 & 10.00 & 15.00 \\
\textbf{Qwen-72B} & 42.50 & 47.50 & 50.00 & 32.50 & 37.50 & 40.00 \\
\midrule
\rowcolor{lightgray}\multicolumn{7}{c}{\textbf{Closed-Source Models}} \\ \midrule
\rowcolor{mylightgray}\multicolumn{7}{c}{\textbf{Claude Series}} \\
\textbf{Claude 3.5 Sonnet} & 45.00 & 65.00 & 65.00 & 42.50 & 57.50 & 62.50 \\
\textbf{Claude 3.7 Sonnet} & 65.00 & 77.50 & 80.00 & 57.50 & 65.00 & 65.00 \\
\textbf{Claude 4.0 Sonnet} & 67.50 & 80.00 & 82.50 & 55.00 & 60.00 & 65.00 \\
\textbf{Claude 4.0 Opus} & 75.00 & 77.50 & 77.50 & 52.50 & 55.00 & 60.00 \\
\midrule
\rowcolor{mylightgray}\multicolumn{7}{c}{\textbf{Gemini Series}} \\
\textbf{Gemini-2.0-Flash} & 45.00 & 45.00 & 50.00 & 35.00 & 45.00 & 47.50 \\
\textbf{Gemini-2.5-Flash} & 42.50 & 57.50 & 62.50 & 52.50 & 62.50 & 67.50 \\
\textbf{Gemini-2.5-Pro} & 57.50 & 70.00 & 70.00 & 60.00 & 77.50 & 82.50 \\
\midrule
\rowcolor{mylightgray}\multicolumn{7}{c}{\textbf{GPT Series}} \\
\textbf{GPT-4V} & 45.00 & 55.00 & 60.00 & 40.00 & 50.00 & 55.00 \\
\textbf{GPT-4o-mini} & 42.50 & 50.00 & 55.00 & 37.50 & 40.00 & 42.50 \\
\textbf{GPT-4o} & 55.00 & 57.50 & 60.00 & 45.00 & 52.50 & 55.00 \\
\textbf{GPT-o3} & 80.00 & 87.50 & 87.50 \firstlogo & 77.50 & 85.00 & 87.50 \firstlogo \\
\textbf{GPT-o4-mini} & 75.00 & 85.00 & 85.00 \secondlogo & 65.00 & 75.00 & 77.50 \secondlogo \\
\bottomrule
\end{tabular}
}
\end{table}

The results from the I-PHYRE present an intriguing contrast to the struggles observed in PHYRE, revealing specific strengths and weaknesses of the current generation of VLMs.
Unlike the broad failures in other environments, top-tier models demonstrate remarkable proficiency in I-PHYRE. 

As shown in Figure \ref{fig:PHYRE_IPHYRE} and detailed in Table \ref{tab:iphyre_model_performance_vla_avg_step}, leading models like GPT-o3 can achieve a final success rate of 81.67\% within 10 attempts. Similarly, GPT-o4-mini and Claude 4.0 Opus show relatively strong performance, reaching 75.00\% and 71.67\%, respectively. 
This indicates that these advanced models possess significant capacity for temporal planning and causal chain reasoning. The high success rates suggest that when a physics problem is framed as a structured, sequential puzzle, the underlying strengths of VLMs in logic and sequence generation can potentially be effectively leveraged. Furthermore, the average number of attempts for successful models is relatively low (e.g., 2.84 att. for GPT-o3), implying they can learn more efficiently from failed trials and correct their temporal strategies.

Nonetheless, I-PHYRE also exposes a provocative performance gap. While the best closed-source models excel, the open-source models we tested still struggle significantly. For instance, models in the Qwen series consistently perform far below the MOCK (random action) baseline, with Qwen-7B barely reaching a 20\% Success Rate. This stark failure suggests that their ability to parse the visual scene and map it to a structured, time-sensitive action plan is severely underdeveloped. I-PHYRE, therefore, serves as an effective differentiator, separating models with robust sequential reasoning from those that lack this crucial capability.

Our analysis of prompt formats also yields clear conclusiosn for this environment. As illustrated in the comparative plot in Figure \ref{fig:PHYRE_IPHYRE_VLA_WM}(b), the WM prompt format consistently fails to outperform the simpler VLA format. For nearly all high-performing models, data points lie below the diagonal line of equivalence, indicating the VLA format to be currently superior. For example, GPT-o3's success rate drops from 81.67\% (VLA) to 76.67\% (WM). 

This reinforces a core finding of our benchmark paper: forcing a model to generate a descriptive prediction of physical outcomes does not necessarily translate to better procedural control (at least in the current generation of models). In the context of I-PHYRE, the cognitive load of predicting the complex cascade of falling blocks may interfere with the more direct task of generating the correct timed JSON sequence, leading to degraded performance. This further highlights the disconnect between descriptive understanding and effective action generation and the need for such benchmarking in new model generations.

\section{Benchmark: Kinetix} \label{appendix:kinetix}

In the Kinetix\footnote{{https://github.com/FLAIROx/Kinetix}} benchmark, the unified objective across all environments is to \textbf{make the green block contact the blue block while avoiding the red block}. Each environment contains at most four controllable \textbf{motors} (which rotate forward/backward to drive connected components such as robotic arms or wheels) and two controllable \textbf{thrusters}. 

\begin{figure*}[h]
    \centering
    \begin{subfigure}[b]{0.16\textwidth}
        \centering
        \includegraphics[trim={1.36cm 1cm 0cm 0cm}, clip, width=\linewidth]{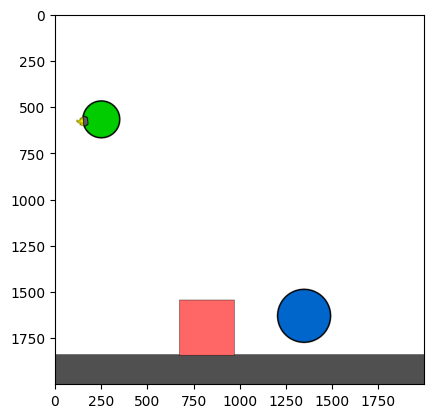}
    \end{subfigure}
    \begin{subfigure}[b]{0.16\textwidth}
        \centering
        \includegraphics[trim={1.36cm 1cm 0cm 0cm}, clip, width=\linewidth]{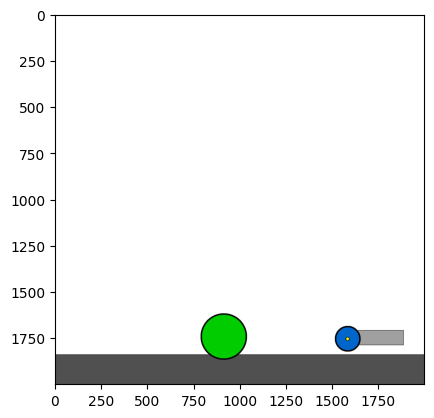}
    \end{subfigure}
    \begin{subfigure}[b]{0.16\textwidth}
        \centering
        \includegraphics[trim={1.36cm 1cm 0cm 0cm}, clip, width=\linewidth]{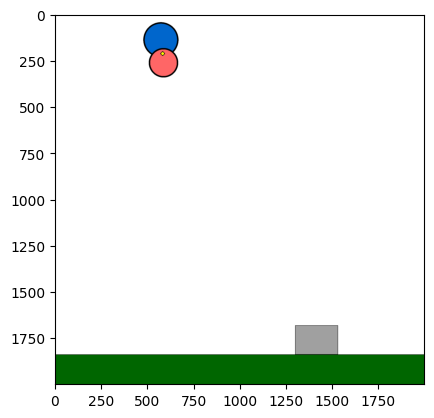}
    \end{subfigure}
    \begin{subfigure}[b]{0.16\textwidth}
        \centering
        \includegraphics[trim={1.36cm 1cm 0cm 0cm}, clip, width=\linewidth]{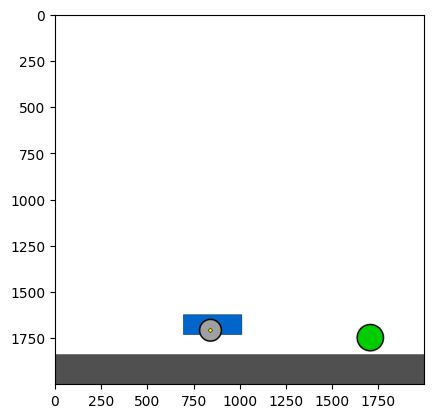}
    \end{subfigure}
    \begin{subfigure}[b]{0.16\textwidth}
        \centering
        \includegraphics[trim={1.36cm 1cm 0cm 0cm}, clip, width=\linewidth]{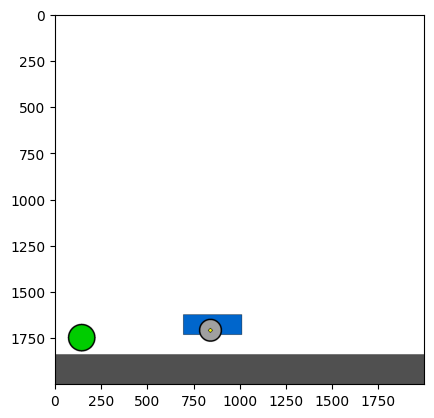}
    \end{subfigure}
    \begin{subfigure}[b]{0.16\textwidth}
        \centering
        \includegraphics[trim={1.36cm 1cm 0cm 0cm}, clip, width=\linewidth]{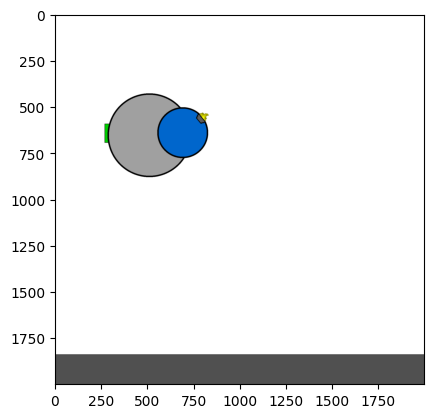}
    \end{subfigure}

    \begin{subfigure}[b]{0.16\textwidth}
        \centering
        \includegraphics[trim={1.36cm 1cm 0cm 0cm}, clip, width=\linewidth]{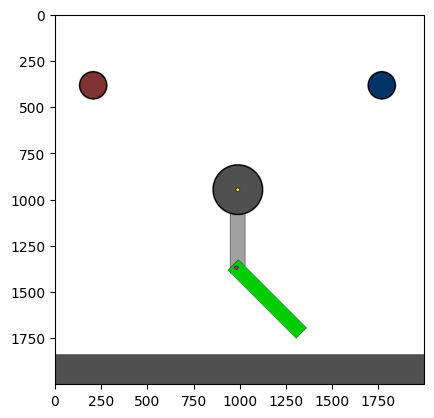}
    \end{subfigure}
    \begin{subfigure}[b]{0.16\textwidth}
        \centering
        \includegraphics[trim={1.36cm 1cm 0cm 0cm}, clip, width=\linewidth]{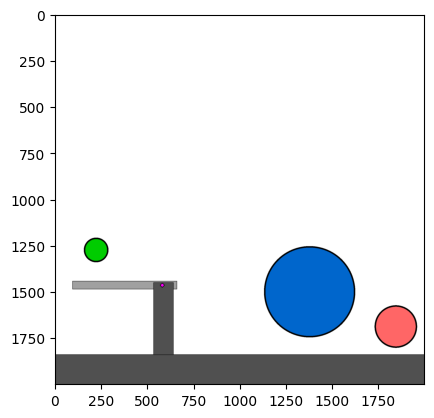}
    \end{subfigure}
    \begin{subfigure}[b]{0.16\textwidth}
        \centering
        \includegraphics[trim={1.36cm 1cm 0cm 0cm}, clip, width=\linewidth]{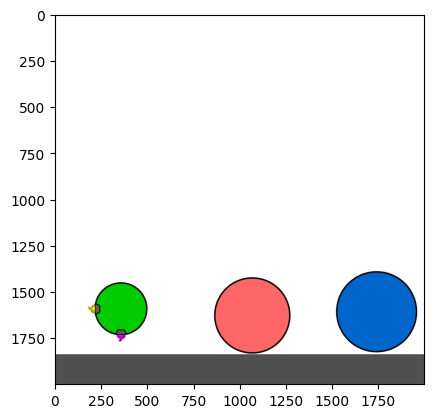}
    \end{subfigure}
    \begin{subfigure}[b]{0.16\textwidth}
        \centering
        \includegraphics[trim={1.36cm 1cm 0cm 0cm}, clip, width=\linewidth]{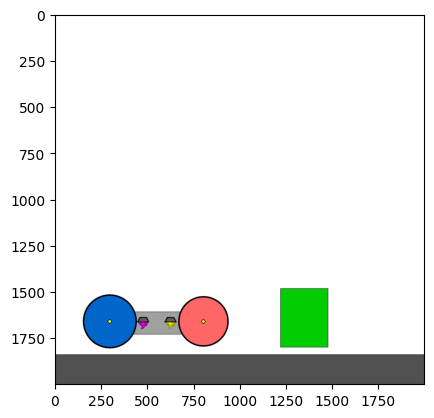}
    \end{subfigure}
    \begin{subfigure}[b]{0.16\textwidth}
        \centering
        \includegraphics[trim={1.36cm 1cm 0cm 0cm}, clip, width=\linewidth]{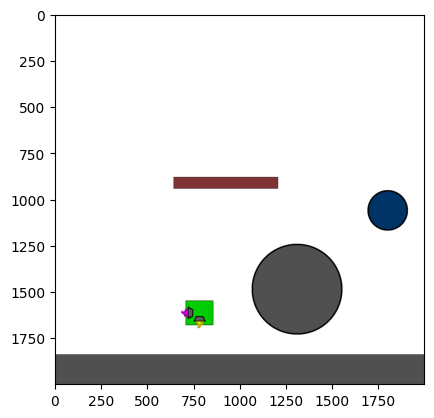}
    \end{subfigure}
    \begin{subfigure}[b]{0.16\textwidth}
        \centering
        \includegraphics[trim={1.36cm 1cm 0cm 0cm}, clip, width=\linewidth]{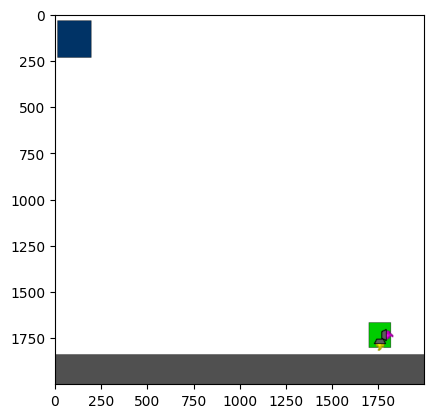}
    \end{subfigure}

    \begin{subfigure}[b]{0.16\textwidth}
        \centering
        \includegraphics[trim={1.36cm 1cm 0cm 0cm}, clip, width=\linewidth]{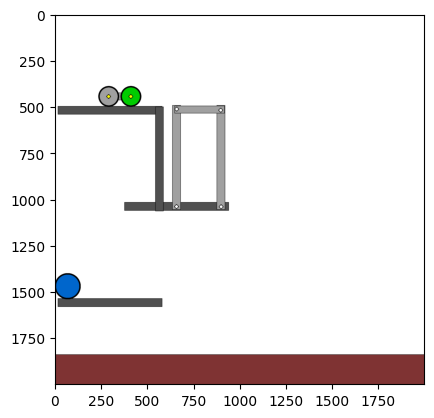}
    \end{subfigure}
    \begin{subfigure}[b]{0.16\textwidth}
        \centering
        \includegraphics[trim={1.36cm 1cm 0cm 0cm}, clip, width=\linewidth]{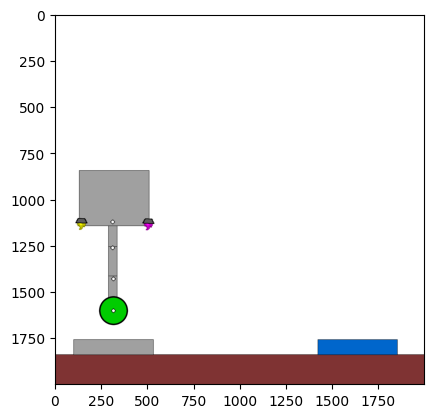}
    \end{subfigure}
    \begin{subfigure}[b]{0.16\textwidth}
        \centering
        \includegraphics[trim={1.36cm 1cm 0cm 0cm}, clip, width=\linewidth]{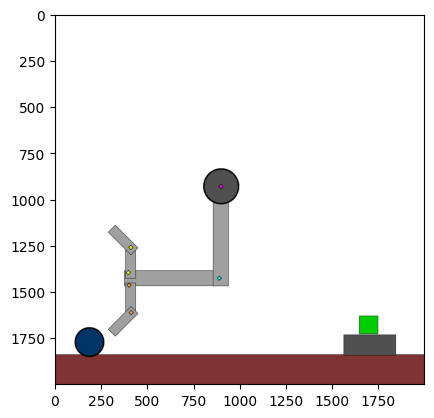}
    \end{subfigure}
    \begin{subfigure}[b]{0.16\textwidth}
        \centering
        \includegraphics[trim={1.36cm 1cm 0cm 0cm}, clip, width=\linewidth]{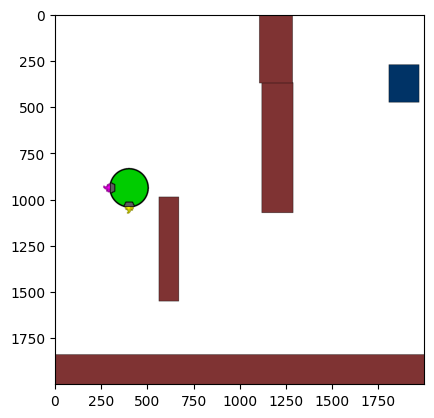}
    \end{subfigure}
    \begin{subfigure}[b]{0.16\textwidth}
        \centering
        \includegraphics[trim={1.36cm 1cm 0cm 0cm}, clip, width=\linewidth]{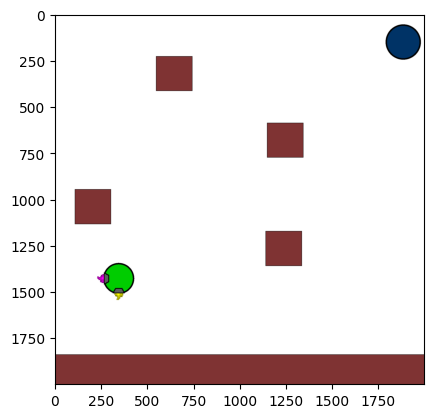}
    \end{subfigure}
    \begin{subfigure}[b]{0.16\textwidth}
        \centering
        \includegraphics[trim={1.36cm 1cm 0cm 0cm}, clip, width=\linewidth]{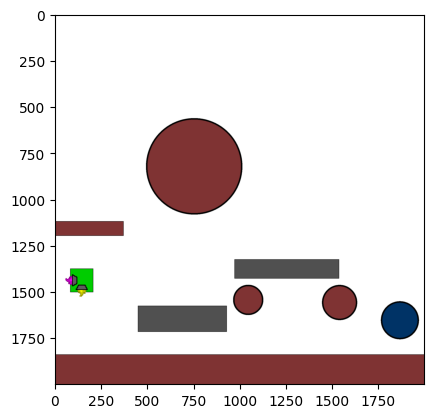}
    \end{subfigure}

    \begin{subfigure}[b]{0.16\textwidth}
        \centering
        \includegraphics[trim={1.36cm 1cm 0cm 0cm}, clip, width=\linewidth]{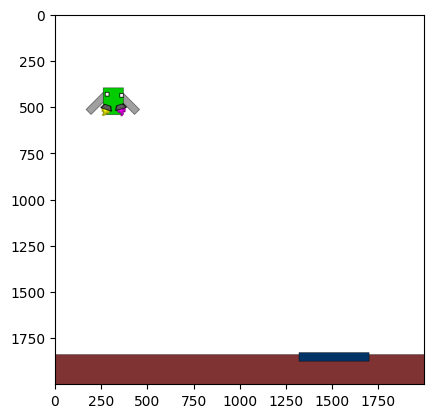}
    \end{subfigure}
    \begin{subfigure}[b]{0.16\textwidth}
        \centering
        \includegraphics[trim={1.36cm 1cm 0cm 0cm}, clip, width=\linewidth]{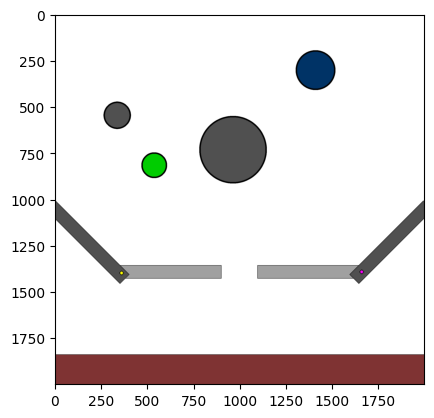}
    \end{subfigure}
    \begin{subfigure}[b]{0.16\textwidth}
        \centering
        \includegraphics[trim={1.36cm 1cm 0cm 0cm}, clip, width=\linewidth]{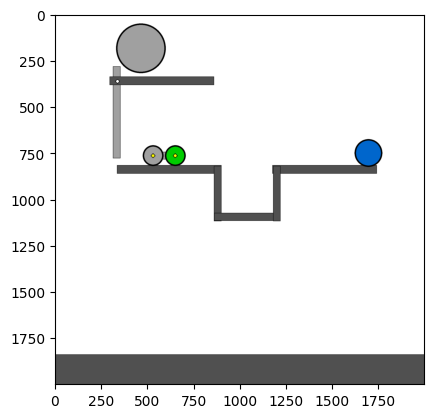}
    \end{subfigure}
    \begin{subfigure}[b]{0.16\textwidth}
        \centering
        \includegraphics[trim={1.36cm 1cm 0cm 0cm}, clip, width=\linewidth]{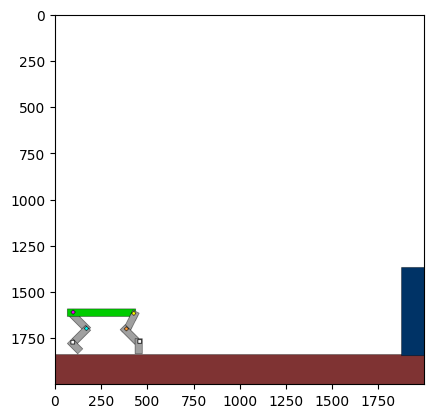}
    \end{subfigure}
    \begin{subfigure}[b]{0.16\textwidth}
        \centering
        \includegraphics[trim={1.36cm 1cm 0cm 0cm}, clip, width=\linewidth]{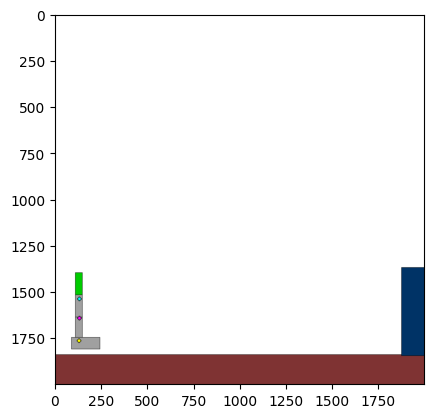}
    \end{subfigure}
    \begin{subfigure}[b]{0.16\textwidth}
        \centering
        \includegraphics[trim={1.36cm 1cm 0cm 0cm}, clip, width=\linewidth]{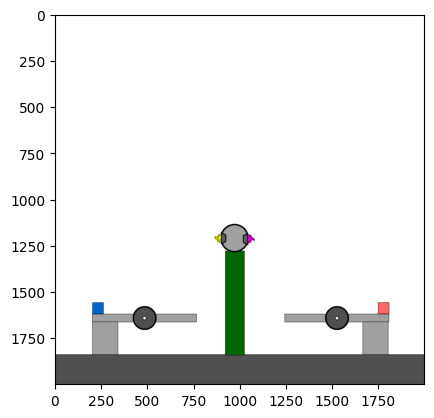}
    \end{subfigure}
    
    \caption{Kinetix Benchmark Observation Space. In difficulty, from top to bottom: S-level tasks (\textit{small, 1st row}), M-level tasks (\textit{medium, 2nd row}), and L-level tasks (\textit{large, 3rd row} and \textit{4th}).}
\label{fig:Kinetix_observation_space}
    
\end{figure*}

By arbitrarily combining \textbf{polygons, joints, motors, and thrusters}, the benchmark generates highly diverse environments (e.g., robot locomotion, object grasping, pinball games). Task difficulty is categorized into three levels (as shown in Figure \ref{fig:Kinetix_observation_space}):

\begin{itemize}
    \item \textbf{S (Small)}: Basic physics puzzles testing agents` mastery of singular controls (e.g., pushing/swinging);  
    \item \textbf{M (Medium)}: Complex mechanisms requiring coordinated control of multiple motors/thrusters;  
    \item \textbf{L (Large)}: Systemically challenging tasks replicating classical hard problems (e.g., robot walking) to evaluate advanced planning and fine-grained control.  
\end{itemize}

\noindent \textbf{Task Distribution:  }

\begin{itemize}
    \item \textbf{Training}: Programmatically sampled levels (theoretically infinite, with tens of millions per difficulty tier);  
    \item \textbf{Testing}: Fixed handcrafted levels forming an interpretable test set (S: 10 levels, M: 24 levels, L: 40 levels), totaling $10+24+40=74$ levels.  
\end{itemize}

\subsection{Conversion Details}

\begin{figure}[h]
    \centering
    \begin{subfigure}[b]{0.15\textwidth}
        \centering
        \includegraphics[trim={1.15cm 1cm 0cm 0cm}, clip, width=\linewidth]{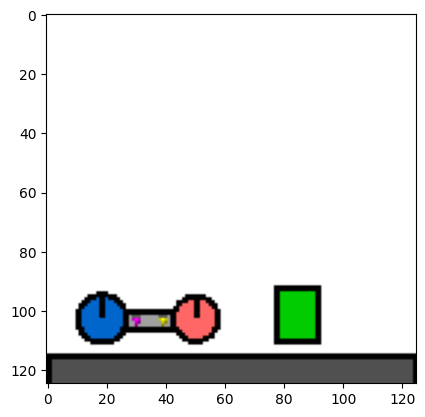}
    \end{subfigure}
    \begin{subfigure}[b]{0.15\textwidth}
        \centering
        \includegraphics[trim={1.36cm 1cm 0cm 0cm}, clip, width=\linewidth]{fig/03_kinetix/Kinetix_sample/00_m/h7_car_flip.png}
    \end{subfigure}
    \begin{subfigure}[b]{0.15\textwidth}
        \centering
        \includegraphics[trim={1.36cm 1cm 0cm 0cm}, clip, width=\linewidth]{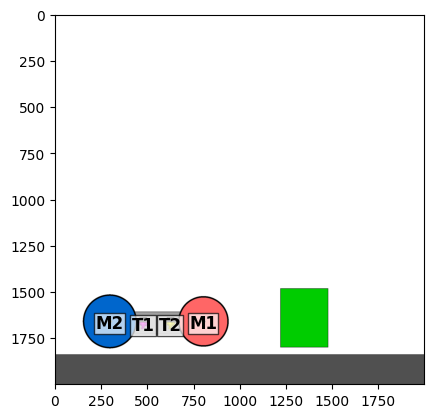}
    \end{subfigure}

    \begin{subfigure}[b]{0.15\textwidth}
        \centering
        \includegraphics[trim={1.15cm 1cm 0cm 0cm}, clip, width=\linewidth]{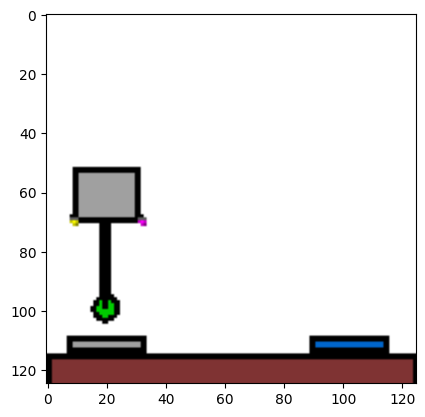}
    \end{subfigure}
    \begin{subfigure}[b]{0.15\textwidth}
        \centering
        \includegraphics[trim={1.36cm 1cm 0cm 0cm}, clip, width=\linewidth]{fig/03_kinetix/Kinetix_sample/00_l/chain_lander.png}
    \end{subfigure}
    \begin{subfigure}[b]{0.15\textwidth}
        \centering
        \includegraphics[trim={1.36cm 1cm 0cm 0cm}, clip, width=\linewidth]{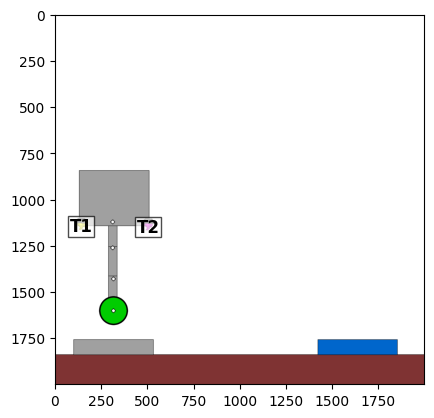}
    \end{subfigure}

    \begin{subfigure}[b]{0.15\textwidth}
        \centering
        \includegraphics[trim={1.15cm 1cm 0cm 0cm}, clip, width=\linewidth]{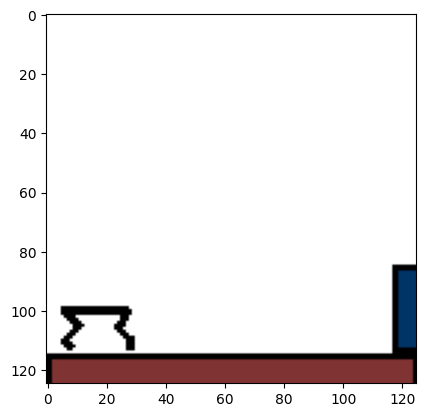}
    \end{subfigure}
    \begin{subfigure}[b]{0.15\textwidth}
        \centering
        \includegraphics[trim={1.36cm 1cm 0cm 0cm}, clip, width=\linewidth]{fig/03_kinetix/Kinetix_sample/00_l/mjc_half_cheetah.png}
    \end{subfigure}
    \begin{subfigure}[b]{0.15\textwidth}
        \centering
        \includegraphics[trim={1.36cm 1cm 0cm 0cm}, clip, width=\linewidth]{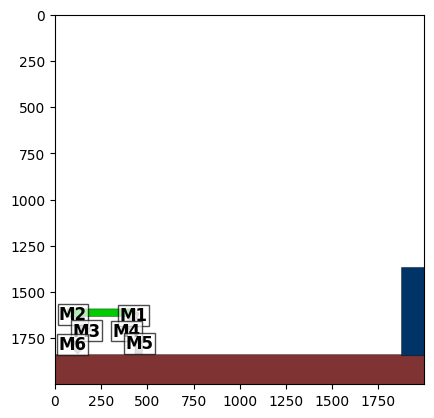}
    \end{subfigure}

    \caption{Augmentations to the Kinetix rendering pipeline to improve VLM \textit{compatibility}. \textbf{Left Column}: The original rendering, where motors and thrusters are rendered with colors often nearly indistinguishable, posing a significant challenge for visual recognition. \textbf{Middle Column}: Our refactored rendering enables unambiguous identification. \textbf{Right Column}: Adding explicit identifier labels (e.g., `M0', `T1') to all interactive motors and thrusters for VLMs to select.}
    \label{fig:Kinetix_render_example}
\end{figure}

As Kinetix was not originally designed with VLM usage in mind, its native image rendering exhibits suboptimal resolution and clarity. Thus, we refactor its observation space, also upgrading its rendering pipeline to enhance clarity (as shown in Figure \ref{fig:Kinetix_render_example}), enabling VLMs to parse visual information more accurately. 

Agents interact with the environment through a defined action space (i.e., [a\_motor, b\_motor, c\_motor, d\_motor, a\_thruster, b\_thruster]), where each component is a continuous value in the range [-1, 1]. To accommodate discrete commands from the VLMs, we establish a specific mapping from discrete integer inputs to these continuous values. As detailed in Table \ref{tab:action_mapping}, for motors, inputs 0, 1, 2 correspond to no action, forward torque, and reverse torque, respectively. For thrusters, inputs 0 and 1 control the off and on (maximum thrust) states.

\begin{table}[h]
    \centering
    \caption{Mapping from Kinetix discrete control inputs to continuous action values for motors and thrusters.}
    \label{tab:action_mapping}
    \begin{tabular}{c|c|c}  \toprule
    \textbf{Component} & \textbf{Input} & \textbf{Physical Meaning} \\  
    \midrule
    \textbf{Motor} & 0 $\to$ 0.0 & No action \\  
    & 1 $\to$ 1.0 & Forward torque/velocity \\  
    & 2 $\to$ -1.0 & Reverse torque/velocity \\  
    \midrule 
    \textbf{Thruster} & 0 $\to$ 0.0 & Thruster off \\  
    & 1 $\to$ 1.0 & Maximum thrust \\  \bottomrule
    \end{tabular}  
\end{table}

Environments in Kinetix are \textbf{temporally dense}, with a maximum of 256 steps. To mitigate VLM confusion from minimal inter-step variations, we introduced a step extension factor $step\_times$ (set to 16 in experiments), scaling each step's physical simulation duration by $step\_times$ times. 
The models are provided with a history of $history\_memory\_steps$ steps, which is set to 5 in our experiments.

\subsection{Prompts}

\begin{systempromptbox}{Kinetix: System Prompt}
You are an AI agent controlling entities in a 2D physics environment.\\
Your task is to generate an integer vector of length \{sum\_actions\} to control the labeled entities.\\
You will be given two images of the current scene: one clean image and one with labels on the controllable entities.\\

\textbf{Action Vector Structure:}\\
- The vector has \{sum\_actions\} integer elements. \\
- The first \{num\_active\_joints\} elements correspond to the motors, labeled \{motor\_labels\} in the annotated image.

\quad - `0': No action

\quad - `1': Positive rotation

\quad - `2': Negative rotation\\

- The next \{num\_active\_thrusters\} elements correspond to the thrusters, labeled \{thruster\_labels\} in the annotated image.

\quad - `0': No action

\quad - `1': Positive thrust

Your output MUST be ONLY a JSON-formatted list of integers, like `[action\_M1, action\_M2, ..., action\_T1, ...]'. Do NOT include any explanations.\\
Objective: Make green objects touch blue objects, AND green objects MUST NOT touch red objects.\\

\textless raw\_image \textgreater \\
\textless annotated\_image \textgreater
\end{systempromptbox}

\begin{userpromptbox}{Kinetix: User Prompt}
    Below is the history: the \{history\_memory\_steps\} previous visual scenes, their corresponding actions \{previous\_action\_list\} and the resulting object distances \{distance\_list\}. \\

    \textless\ previous\_image\_list \textgreater \\
    \textless\ current\_raw\_image \textgreater \\
    \textless\ current\_annotated\_image \textgreater

\textbf{Goal: } Make green objects touch blue objects; Green objects must NOT touch red objects. \\
Output the \{num\_active\_joints\} + \{num\_active\_thrusters-dimensional\} action vector NOW.  \\
\textbf{Format: } `[int, int, ..., int]` with \{num\_active\_joints\} + \{num\_active\_thrusters\} integers. \\
Your entire response must be ONLY the vector.
\end{userpromptbox}

\subsection{Experiment Results}

\begin{table}[h!]
\centering
\caption{Benchmark Kinetix. Model performance comparison. The 'w/o anno' condition uses refactored images, while the 'w/ anno' condition provides additional numerical labels on controllable entities (as shown in Figure \ref{fig:Kinetix_render_example}).}
\label{tab:Kinetix_acc_average}
\resizebox{0.75\linewidth}{!}{%
\begin{tabular}{l | llll}
\toprule

\multirow{2}{*}{\textbf{Model}} & 
\multicolumn{2}{c}{VLA} & 
\multicolumn{2}{c}{WM} \\

\cmidrule(lr){2-3} \cmidrule(lr){4-5}

& w/o anno & w/ anno & w/o anno & w/ anno \\

\midrule
\texttt{\textit{MOCK}}
& 20.04\% & 21.40\% & 20.04\% & 21.40\% \\
\midrule

\rowcolor{lightgray}\multicolumn{5}{c}{\textbf{Open-Source Models}} \\ \midrule

\rowcolor{mylightgray}\multicolumn{5}{c}{\textbf{Qwen2.5-VL Series}} \\

\textbf{Qwen-3B} 
& 10.34\% & 16.22\% & 11.26\% & 11.82\% \\

\textbf{Qwen-7B} 
& 16.22\% & 13.51\% & 15.31\% & 08.88\% \\

\textbf{Qwen-32B} 
& 16.89\% & 15.20\% & 15.99\% & 15.54\% \\

\textbf{Qwen-72B} 
& 16.55\% & 14.86\% & 13.96\% & 18.25\% \\

\rowcolor{lightgray}\multicolumn{5}{c}{\textbf{Closed-Source Models}} \\ \midrule

\rowcolor{mylightgray}\multicolumn{5}{c}{\textbf{Claude Series}} \\

\textbf{Claude 3.5 Sonnet}
& 23.01\% & 17.12\% & 17.84\% & 17.17\% \\

\textbf{Claude 3.7 Sonnet}
& 17.57\% & 15.14\% & 22.52\% & 18.41\% \\

\textbf{Claude 4.0 Sonnet} 
& 19.86\% & 18.92\% & 16.67\% & 18.81\% \\

\textbf{Claude 4.0 Opus} 
& 16.89\% & 23.20\% & 17.57\% & 10.81\% \\

\midrule

\rowcolor{mylightgray}\multicolumn{5}{c}{\textbf{Gemini Series}} \\

\textbf{Gemini-2.0-Flash} 
& 16.76\% & 17.57\% & 19.59\% & 17.30\% \\

\textbf{Gemini-2.5-Pro} 
& 26.69\% \firstlogo & 24.10\% & 25.90\% \secondlogo & 24.37\% \\

\textbf{Gemini-2.5-Flash} 
& 22.63\% & 26.35\% & 19.37\% & 26.12\% \secondlogo \\

\midrule

\rowcolor{mylightgray}\multicolumn{5}{c}{\textbf{GPT Series}} \\

\textbf{GPT-4V} 
& 15.41\% & 16.22\% & 12.61\% & 12.84\% \\

\textbf{GPT-4o-mini} 
& 15.41\% & 13.24\% & 13.06\% & 14.08\% \\

\textbf{GPT-4o} 
& 20.81\% & 18.11\% & 16.67\% & 17.46\% \\

\textbf{GPT-o3} 
& 23.99\% & 26.89\% \firstlogo & 27.48\% \firstlogo & 27.48\% \firstlogo \\

\textbf{GPT-o4-mini} 
& 24.80\% \secondlogo & 26.80\% \secondlogo & 24.21\% & 19.60\% \\

\bottomrule
\end{tabular}
}
\end{table}

\begin{table*}[h!]
\centering
\caption{Benchmark Kinetix. Model performance comparison. Results represent averages of triplicate experiments within a maximum of 16 attempts.}
\label{tab:Kinetix_acc_total}
\resizebox{\textwidth}{!}{
\begin{tabular}{l | llll llll llll}
\toprule

\multirow{3}{*}{\textbf{Model}} & \multicolumn{4}{c}{\textbf{Level S }(10)} & \multicolumn{4}{c}{\textbf{Level M }(24)} & \multicolumn{4}{c}{\textbf{Level L }(40)} \\

\cmidrule(lr){2-5} \cmidrule(lr){6-9} \cmidrule(lr){10-13}
& \multicolumn{2}{c}{\textbf{VLA}} & \multicolumn{2}{c}{\textbf{WM}} & \multicolumn{2}{c}{\textbf{VLA}} & \multicolumn{2}{c}{\textbf{WM}} & \multicolumn{2}{c}{\textbf{VLA}} & \multicolumn{2}{c}{\textbf{WM}} \\

\cmidrule(lr){2-3} \cmidrule(lr){4-5} \cmidrule(lr){6-7} \cmidrule(lr){8-9} \cmidrule(lr){10-11} \cmidrule(lr){12-13}

& \textbf{w/o anno} & \textbf{w/ anno} & \textbf{w/o anno} & \textbf{w/ anno} & \textbf{w/o anno} & \textbf{w/ anno} & \textbf{w/o anno} & \textbf{w/ anno} & \textbf{w/o anno} & \textbf{w/ anno} & \textbf{w/o anno} & \textbf{w/ anno} \\

\midrule
\texttt{\textit{MOCK}}
& 30.00\% & 35.00\%  & 30.00\% & 35.00\% 
& 31.25\% & 40.28\% \secondlogo & 31.25\% & 40.28\% \secondlogo
& 10.83\% \secondlogo & 06.67\% & 10.83\% \firstlogo & 06.67\% \\
\midrule

\rowcolor{lightgray}\multicolumn{13}{c}{\textbf{Open-Source Models}} \\ \midrule

\rowcolor{mylightgray}\multicolumn{13}{c}{\textbf{Qwen2.5-VL Series}} \\

\textbf{Qwen-3B} 
& 14.00\% & 40.00\% & 20.00\% & 10.00\% 
& 21.88\% & 25.00\% & 22.22\% & 23.96\% 
& 02.50\% & 05.00\% & 02.50\% & 05.00\% \\

\textbf{Qwen-7B} 
& 30.00\% & 20.00\% & 50.00\% & 20.00\% 
& 25.00\% & 25.00\% & 22.22\% & 12.50\% 
& 07.50\% & 05.00\% & 02.50\% & 03.93\% \\

\textbf{Qwen-32B} 
& 52.50\% & 20.00\% & 33.33\% & 42.50\% 
& 25.00\% & 30.21\% & 25.00\% & 22.92\% 
& 03.12\% & 05.00\% & 06.25\% & 04.38\% \\

\textbf{Qwen-72B} 
& 30.00\% & 32.00\% & 40.00\% & 50.00\% 
& 32.50\% & 28.33\% & 20.83\% & 28.13\% 
& 03.61\% & 02.50\% & 03.33\% & 04.38\% \\

\rowcolor{lightgray}\multicolumn{13}{c}{\textbf{Closed-Source Models}} \\ \midrule

\rowcolor{mylightgray}\multicolumn{13}{c}{\textbf{Claude Series}} \\

\textbf{Claude 3.5 Sonnet}
& 54.29\% & 56.67\% & 62.00\% \secondlogo & 53.75\% 
& 38.33\%  \secondlogo & 25.00\% & 22.92\% & 25.00\% 
& 06.00\% & 02.50\% & 03.75\% & 03.33\% \\

\textbf{Claude 3.7 Sonnet}
& 40.00\% & 48.00\% & 70.00\% \firstlogo & 56.25\% 
& 33.33\% & 25.00\% & 31.94\% & 27.08\% 
& 02.50\% & 01.00\% & 05.00\% & 03.75\% \\

\textbf{Claude 4.0 Sonnet} 
& 56.67\% \secondlogo & 50.00\% & 43.33\% & 42.50\% 
& 28.47\% & 29.17\% & 29.17\% & 27.78\% 
& 05.50\% & 05.00\% & 02.50\% & 07.50\% \\

\textbf{Claude 4.0 Opus} 
& 54.29\% & 68.33\% \firstlogo & 30.00\% & 30.00\% 
& 23.61\% & 34.72\% & 29.17\% & 16.67\% 
& 03.50\% & 05.00\% & 07.50\% & 02.50\% \\

\midrule

\rowcolor{mylightgray}\multicolumn{13}{c}{\textbf{Gemini Series}} \\

\textbf{Gemini-2.0-Flash} 
& 46.00\% & 50.00\% & 45.00\% & 50.00\% 
& 27.50\% & 29.17\% & 37.50\% & 28.33\% 
& 03.00\% & 02.50\% & 02.50\% & 02.50\% \\

\textbf{Gemini-2.5-Pro} 
& 60.00\% \firstlogo & 53.33\% & 55.00\% & 65.00\% \firstlogo
& 39.58\% \firstlogo & 37.50\% & 38.89\% \secondlogo & 34.17\% 
& 10.62\% & 08.75\% & 10.83\% \firstlogo & 08.33\% \secondlogo \\

\textbf{Gemini-2.5-Flash} 
& 47.50\% & 55.00\% & 43.33\% & 57.50\% \secondlogo
& 33.33\% & 43.75\% \firstlogo & 37.50\% & 42.71\% \firstlogo 
& 10.00\% & 08.75\% & 02.50\% & 08.33\% \secondlogo \\

\midrule

\rowcolor{mylightgray}\multicolumn{13}{c}{\textbf{GPT Series}} \\

\textbf{GPT-4V} 
& 30.00\% & 58.00\% & 23.33\% & 40.00\% 
& 30.00\% & 21.67\% & 25.00\% & 16.67\% 
& 03.00\% & 02.50\% & 02.50\% & 03.75\% \\

\textbf{GPT-4o-mini} 
& 42.00\% & 20.00\% & 26.67\% & 27.50\% 
& 21.67\% & 21.67\% & 20.83\% & 19.44\% 
& 05.00\% & 06.50\% & 05.00\% & 07.50\% \\

\textbf{GPT-4o} 
& 46.00\% & 56.00\% & 36.67\% & 52.50\% 
& 36.67\% & 27.50\% & 31.94\% & 27.78\% 
& 05.00\% & 03.00\% & 02.50\% & 02.50\% \\

\textbf{GPT-o3} 
& 50.00\% & 63.33\% \secondlogo & 53.33\% & 60.00\% 
& 36.46\% & 40.28\%  \secondlogo  & 50.00\% \firstlogo  & 36.11\% 
& 10.00\% & 10.00\% \firstlogo & 07.50\%  & 14.17\% \firstlogo \\

\textbf{GPT-o4-mini} 
& 47.50\% & 56.67\% & 56.67\% & 55.00\% 
& 37.50\% & 43.75\% \firstlogo & 37.50\% & 30.56\% 
& 11.50\% \firstlogo & 09.17\% \secondlogo & 08.12\% \secondlogo & 04.17\% \\

\bottomrule
\end{tabular}
}
\end{table*}

The Kinetix environment reveals a clear and steep decline in VLM performance as physical complexity increases, highlighting significant limitations in multi-step, coordinated control. As illustrated in Figure \ref{fig:kinetix_dumbbell_plot_all} and detailed in Tables \ref{tab:Kinetix_acc_average} \& \ref{tab:Kinetix_acc_total}, all models exhibit a significant drop in their success rates when transitioning from simple (S-level) to more complex (M- and L-level) tasks. For instance, top-performing models like GPT-o3 success rates plummet from over 60\% on S-level tasks to under 15\% on L-level tasks. This trend holds true across all model families, indicating that while current VLMs can handle basic single-component manipulation, they falter when faced with tasks requiring the synchronized control of multiple motors and thrusters over a longer time horizon.

Our ablation study on the utility of visual annotations yields a nuanced and counter-intuitive result. In simple S-level tasks, providing explicit labels for controllable parts generally improves performance, as it helps models ground their actions to the correct components. However, this effect often inverts on the more difficult M- and L-level tasks. For many models, particularly when using the WM prompt, performance with annotations becomes worse than without them. 
This suggests that for tasks already within the model's capabilities, adding annotations and using WM prompt format can further enhance their physical reasoning abilities. However, as the number of interactive components increases, these labels and required interpretation may instead become a burden, causing the model to focus on parsing the labels themselves rather than reasoning about the holistic physical structure and dynamics.

Furthermore, the WM prompt format consistently fails to improve, and frequently degrades, performance compared to the direct VLA approach. This is especially evident in the M- and L-level tasks; in line with findings from the previous benchmark results from our suite.

\subsection{Case Study}

\begin{figure}[h]
\centering
\includegraphics[width=0.7\linewidth]{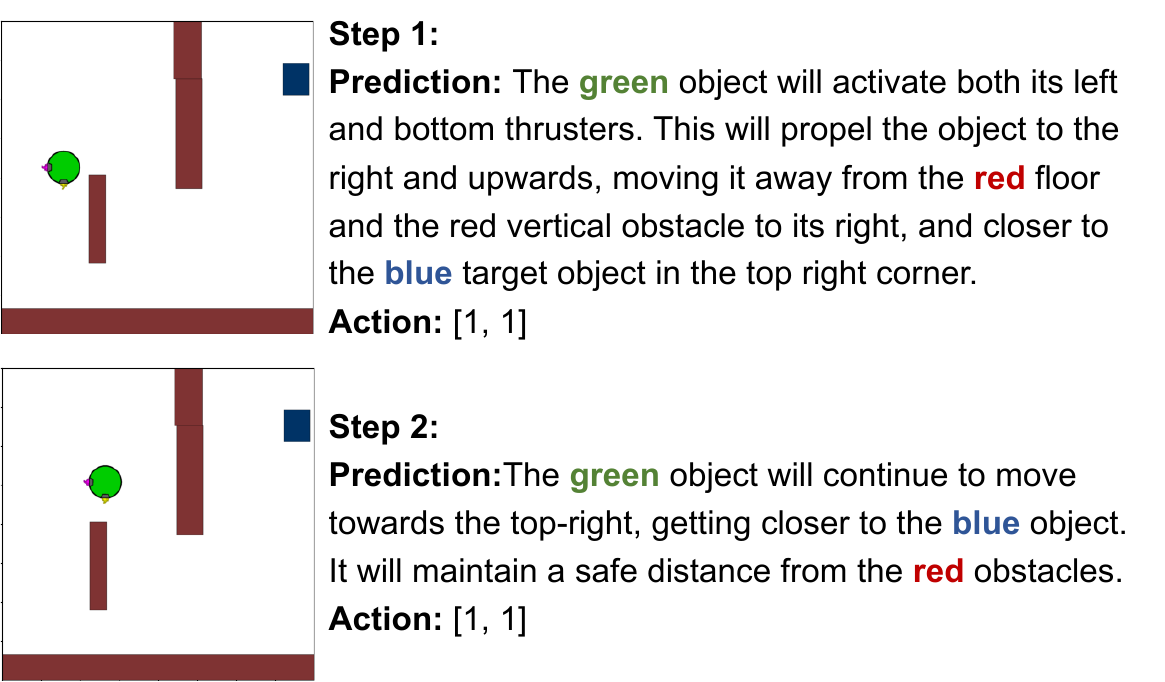}
\caption{Case Study of myopic planning in Kinetix - Gemini-2.5-Flash (WM).}
\label{fig:kinetix_casestudy_gemini}
\end{figure}

As shown in Figure \ref{fig:kinetix_casestudy_gemini}, models can correctly predict the outcome of the first attempt, but fail by repeating the same action without accounting for the agent's new state and momentum. This reveals a fundamental disconnect between descriptive prediction and effective state-aware control.
In the first attempt, the model successfully predicts and executes a move toward the top-right, closing the distance to the target. However, in the second attempt, it repeats the exact same action, failing to account for the object's new position and the momentum it gained from the first move. This myopic repetition causes the agent to collide directly with an obstacle, resulting in task failure.

\section{Benchmark: \pooltoollogo} \label{appendix:pooltool}

\textbf{\textit{Pooltool}}\footnote{{https://github.com/ekiefl/pooltool}} is a billiards sandbox game focused on realistic simulation of physics. 

\subsection{Conversion Details}

As our evaluation prioritizes the performance of a single agent rather than adversarial play, we have adapted its original nine-ball rules: the agent's objective is to pot the 9-ball by hitting the lowest-numbered ball currently on the table. Each model undergoes 100 experiments, with a maximum of 15 attempts per experiment, to assess their average success rate. 
This setup is designed to more precisely measure the current agentic VLMs' understanding of physical rules and their interaction capabilities with the environment, without introducing an adversarial agent/human for now, which could be a research direction for future physical reasoning advancements in agents.

\begin{figure}[h]
    \centering
    \begin{subfigure}[b]{0.34\textwidth}
        \centering
        \includegraphics[width=\linewidth]{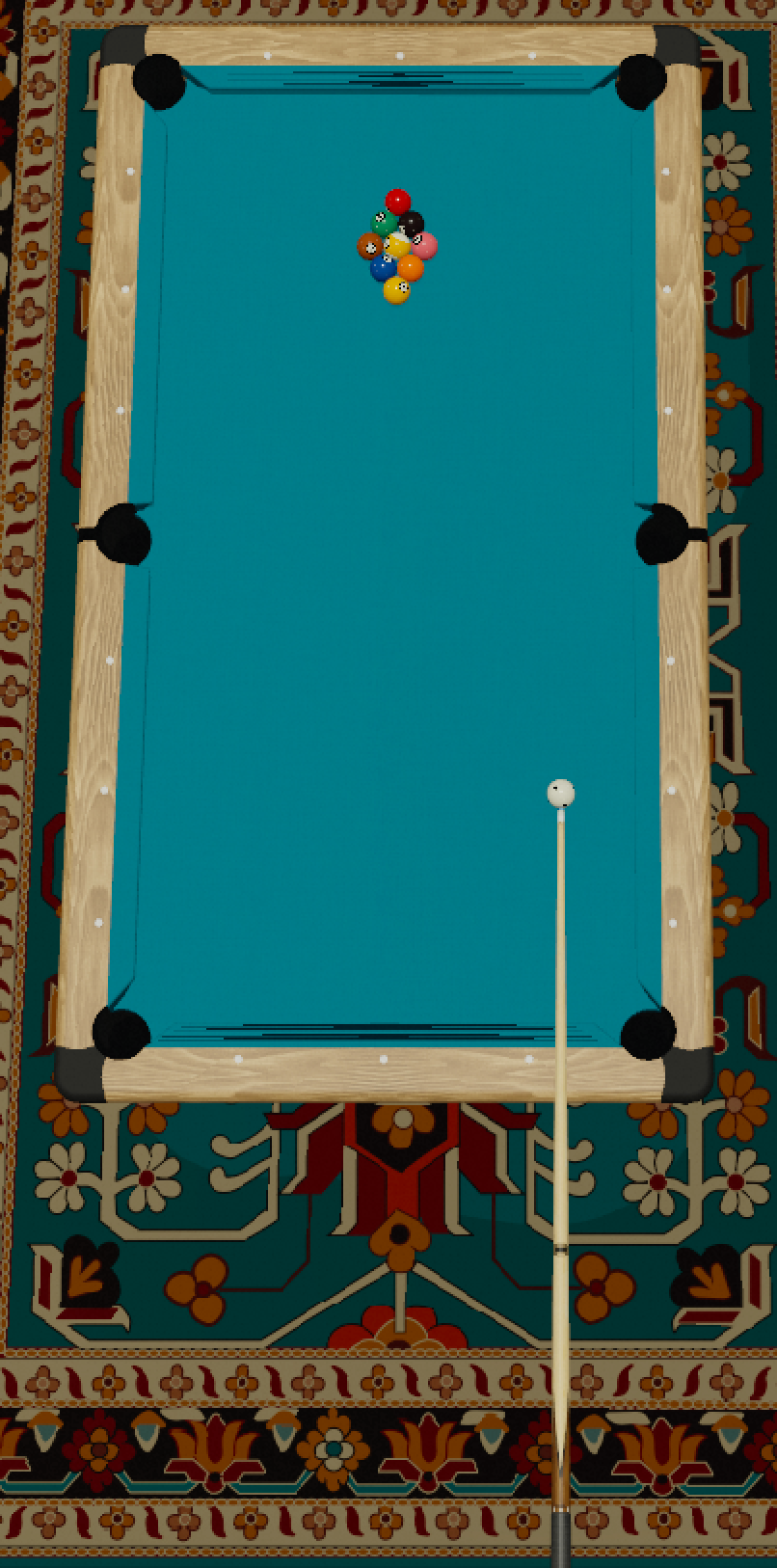}
        \caption{Top-down View.}
    \end{subfigure}
    \begin{subfigure}[b]{0.6\textwidth}
        \centering
        \includegraphics[width=\linewidth]{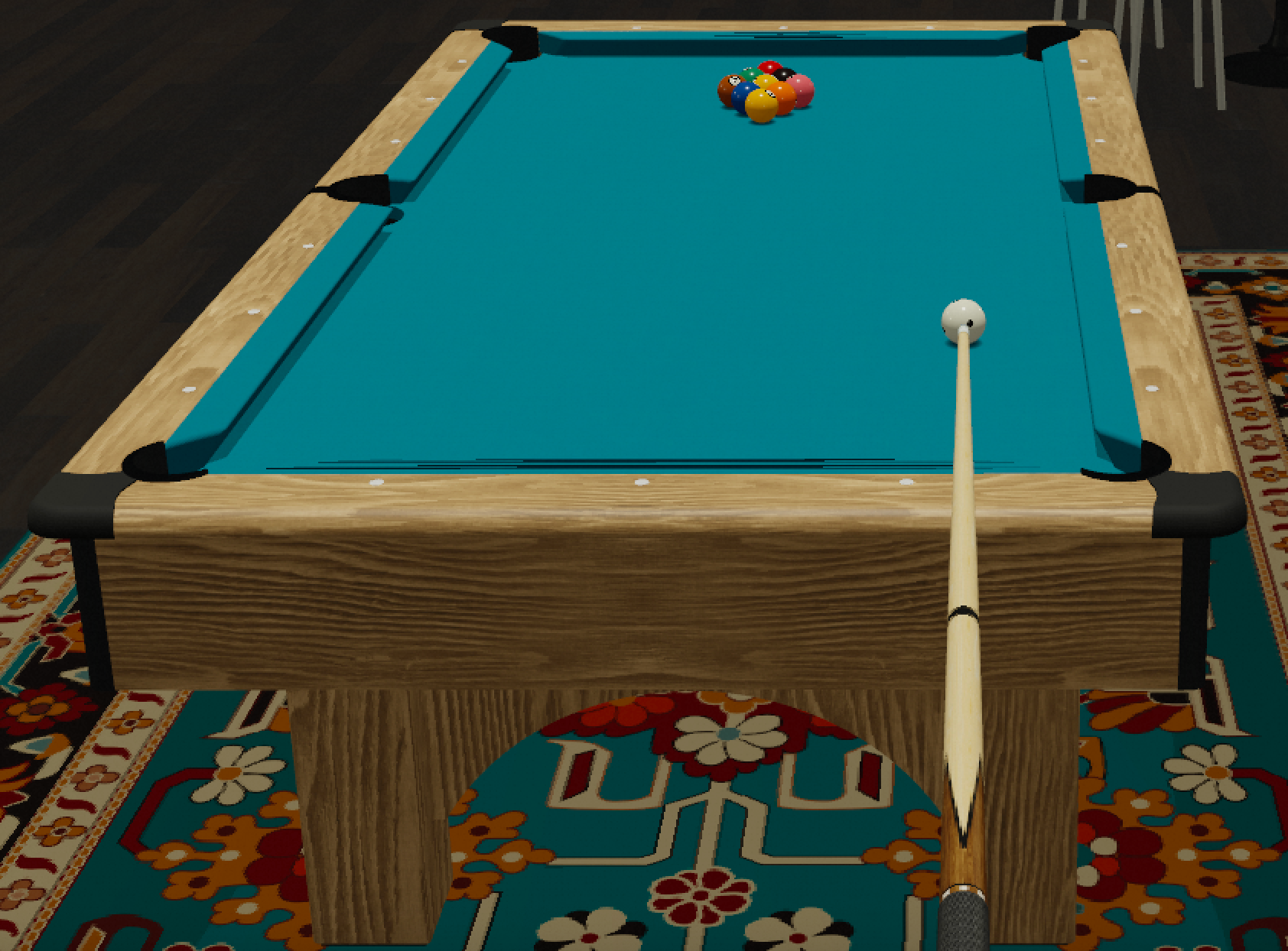}
        \caption{Front View.}
    \end{subfigure}

    \caption{Example 3D Views from Pooltool GUI}
    \label{fig:pooltool_3D}
\end{figure}

\begin{figure}[h]
    \centering
    \begin{subfigure}[b]{0.3\textwidth}
        \centering
        \includegraphics[width=\linewidth]{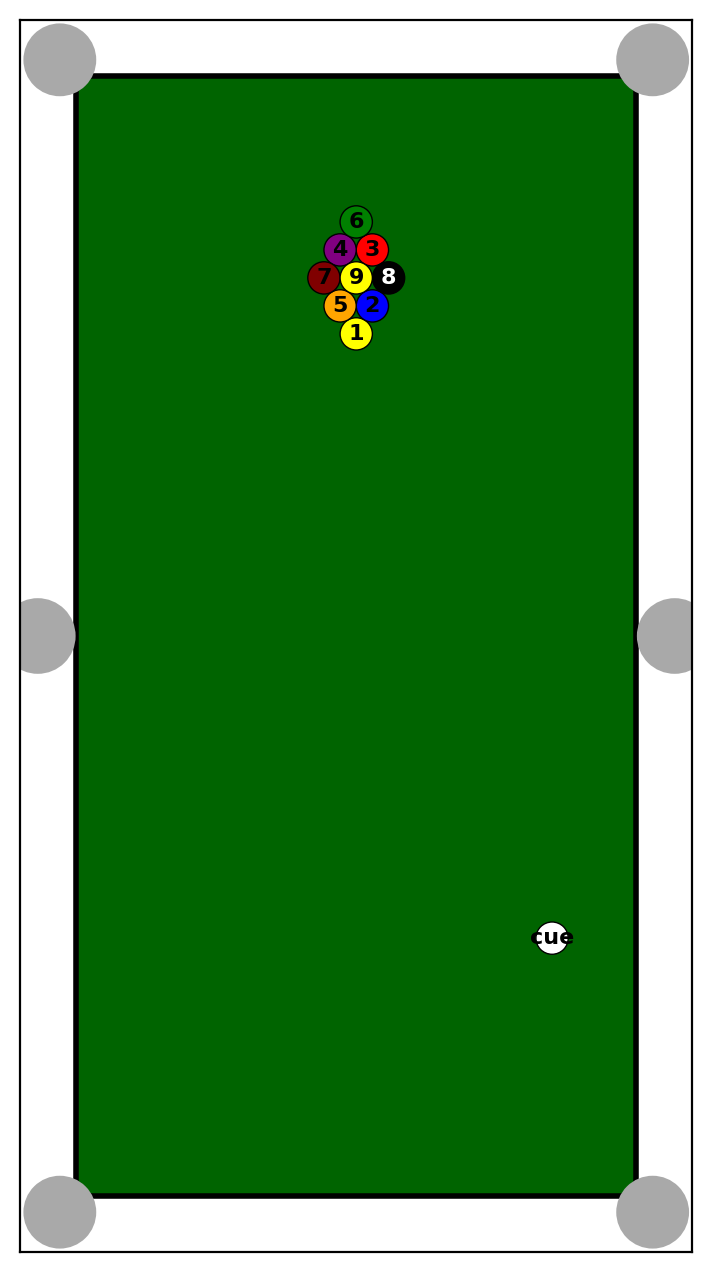}
        \caption{Initial Game State}
    \end{subfigure}
    \begin{subfigure}[b]{0.3\textwidth}
        \centering
        \includegraphics[width=\linewidth]{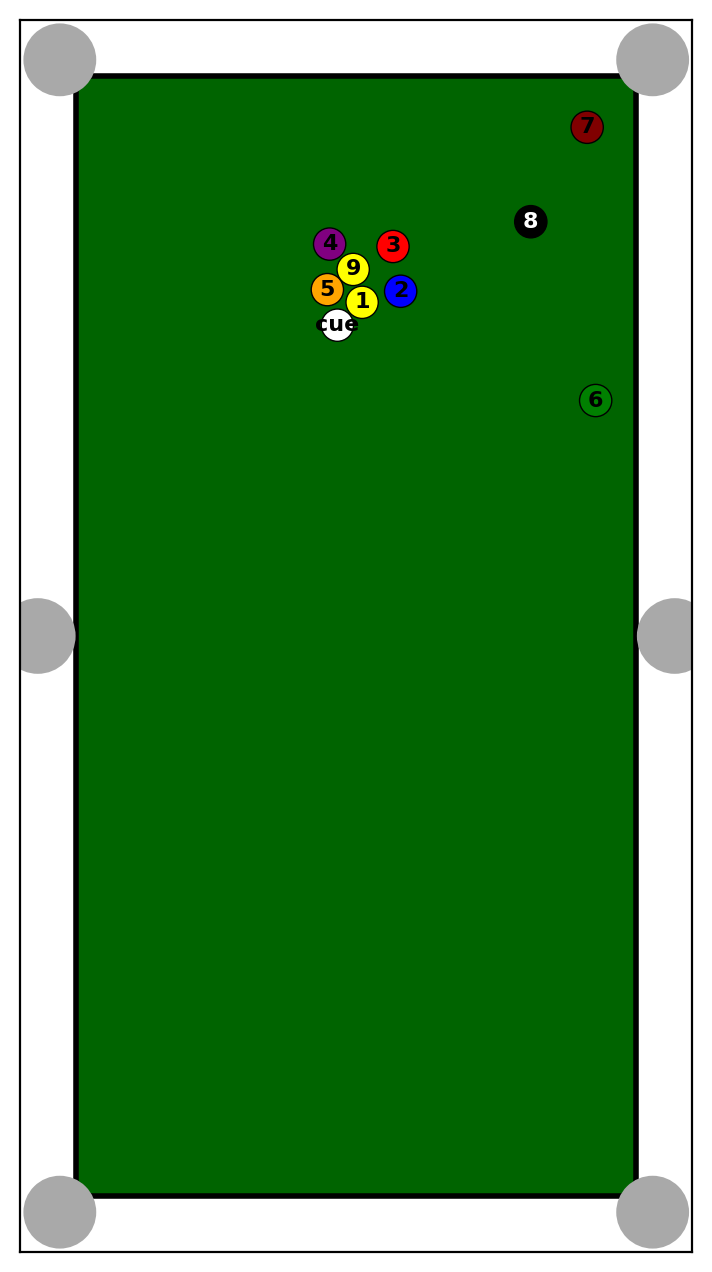}
        \caption{Scene after First Shot}
    \end{subfigure}
    \begin{subfigure}[b]{0.3\textwidth}
        \centering
        \includegraphics[width=\linewidth]{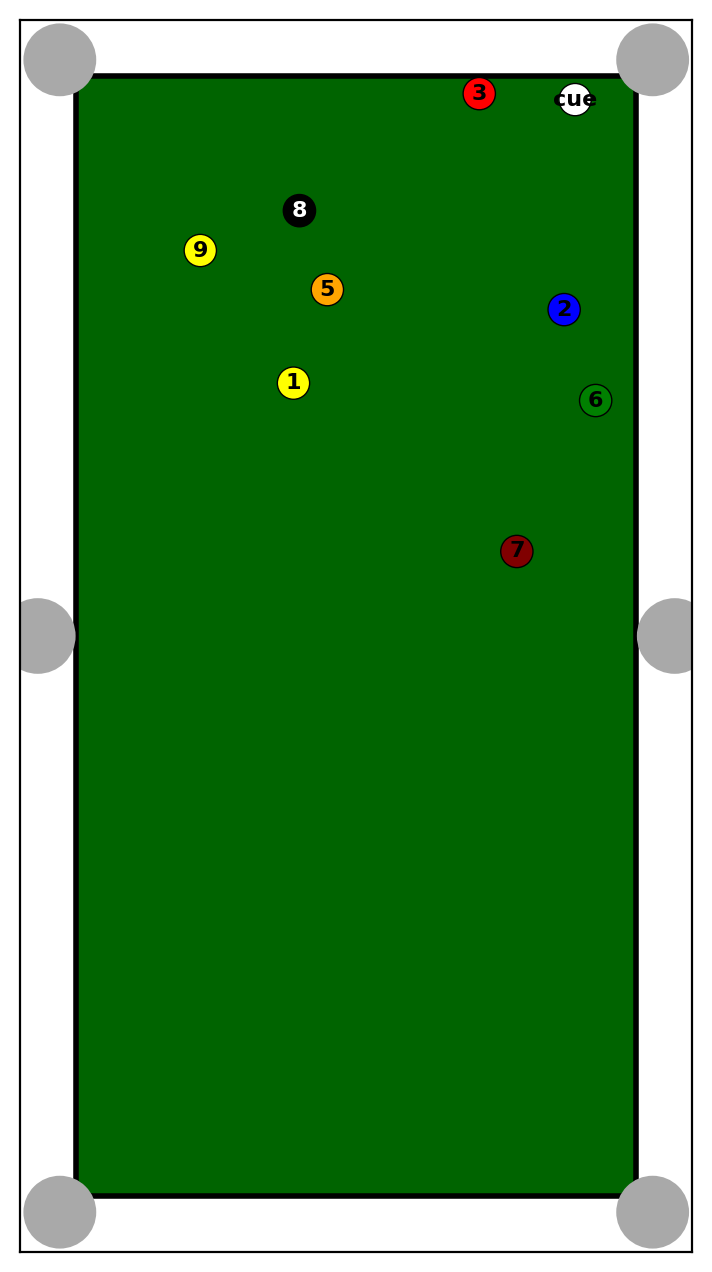}
        \caption{Scene after Second Shot}
    \end{subfigure}

    \caption{Converted 2D views of Pooltool}
    \label{fig:pooltool_2D}
\end{figure}

\textbf{Observation Space:} We converted Pooltool's native 3D rendering (as shown in Figure \ref{fig:pooltool_3D}) into a 2D top-down view rendering (as shown in Figure \ref{fig:pooltool_2D}) to better suit current VLMs' visual input processing.

\textbf{Action Space:} Pooltool offers a Python API for control, where cue ball striking actions are parameterized through the `Cue' object and its `set\_state' method. These parameters collectively determine the shot's power, direction, and spin (\textit{English}). The Pooltool Python API includes settings for the following parameters:

\begin{enumerate}
    \item \textbf{\texttt{v0}} \textbf{(Initial Speed)}:
    \begin{itemize}
        \item \textbf{Definition:} This represents the cue ball's initial velocity (meters/second) after being struck by the cue stick. It directly corresponds to the \textbf{power or intensity of the shot}.
        \item \textbf{VLM Selection:} The VLM selects a predefined discrete power level.
    \end{itemize}
    \begin{examplebox}{}
        \textbf{SPEED\_MAPPING} = \{
        
        \quad    ``Low": 2,     \# Low speed
        
        \quad    ``Medium": 5,  \# Medium speed
        
        \quad    ``High": 8,    \# High speed
        
        \}
    \end{examplebox}
    \item \textbf{\texttt{a}} \textbf{(Lateral Offset)} and \textbf{\texttt{b}} \textbf{(Vertical Offset)}:
    \begin{itemize}
        \item \textbf{Definition:} \texttt{a} defines the horizontal offset distance (meters) from the cue ball's center point, controlling \textbf{side spin (English/Side Spin)}; \texttt{b} defines the vertical offset distance (meters), controlling \textbf{top/bottom spin (Top/Bottom Spin)}.
        \item \textbf{VLM Selection:} The VLM determines the values of \texttt{a} and \texttt{b} by selecting a preset strike spot type.
    \end{itemize}

    \begin{examplebox}{}
        \textbf{STRIKESPOT\_MAPPING} = \{
        
        \quad    ``Top Left Spin": {``a": 0.35, ``b": 0.35},
            
        \quad    ``Top Spin": {``a": 0, ``b": 0.5},
            
        \quad    ``Top Right Spin": {``a": -0.35, ``b": 0.35},
            
        \quad   ``Middle Left Spin": {``a": 0.35, ``b": 0},
            
        \quad    ``Middle Spin": {``a": 0, ``b": 0},
            
        \quad    ``Middle Right Spin": {``a": -0.35, ``b": 0},
            
        \quad    ``Bottom Left Spin": {``a": 0.35, ``b": -0.35},
            
        \quad    ``Bottom Spin": {``a": 0, ``b": -0.5},
            
        \quad    ``Bottom Right Spin": {``a": -0.35, ``b": -0.35}
        
        \}
    \end{examplebox}

    \item \textbf{\texttt{phi}} \textbf{(Horizontal Angle)}:
    \begin{itemize}
        \item  \textbf{Definition:} This defines the horizontal angle (in degrees) of the cue stick strike relative to the cue ball, corresponding to the \textbf{aiming direction}.
        \item \textbf{VLM Simplified Handling:} Considering the challenge for VLMs in selecting precise continuous angles, we simplify the handling: the aiming direction is defaulted to directly target the lowest-numbered ball currently on the table (i.e., by setting \texttt{cue\_ball\_id}). This way, the VLM does not need to directly output an angle value.
    \end{itemize}

    \item  \textbf{\texttt{theta}} \textbf{(Vertical Angle)}:
    \begin{itemize}
        \item \textbf{Definition:} This defines the vertical angle (in degrees) of the cue stick strike relative to the horizontal plane of the table, corresponding to the \textbf{cue height}.
        \item \textbf{VLM Simplified Handling:} We ignore this parameter, as mastering such advanced techniques is overly complex for VLMs.
    \end{itemize}
\end{enumerate} 

\subsection{Prompts}

\begin{systempromptbox}{\pooltoollogo: System Prompt}
You are an expert AI agent specializing in billiards game strategy. Your goal is to guide a player in a 9-ball pool game.\\

\textbf{Game Objective:}

Your ultimate objective is to pocket the number 9-ball. In each turn, you will pocket the number 9-ball by hitting the lowest numbered ball. Your task is to provide the optimal cue ball strike parameters to achieve this.

Crucially, avoid potting the cue ball (white ball). Potting the cue ball is a foul and will negatively impact the game progress.\\

\textbf{Current Table State:}

Image (Current Table State): This is the pool table now. All potted balls from previous shots have been removed. If the cue ball was potted, it has been repositioned. Analyze this image to determine your next move.

You must choose the cue ball's speed and strike spot from the following precise actions.

Available Speed Options: \{``Low", ``Medium", ``High"\}

Available Strikespot Options: \{``Top Left Spin", ``Top Spin", ``Top Right Spin", ``Middle Left Spin", ``Middle Spin", ``Middle Right Spin", ``Bottom Left Spin", ``Bottom Spin", ``Bottom Right Spin"\}

\textbf{Action Parameters:}

\textbf{Strikespot Effects:}
\begin{itemize}
    \item \textbf{Top Spin (High Cue):} Hitting the cue ball above its center. Makes the cue ball continue forward after contact with an object ball, also known as ``follow" or ``roll".
    \item \textbf{Bottom Spin (Low Cue):} Hitting the cue ball below its center. Makes the cue ball spin backward, causing it to rebound or ``draw" after hitting an object ball.
    \item \textbf{Left Spin (Left English):} Hitting the cue ball on its left side. Causes the cue ball to curve or deflect to the left after contact.
    \item \textbf{Right Spin (Right English):} Hitting the cue ball on its right side. Causes the cue ball to curve or deflect to the right after contact.
\end{itemize}

\textbf{Required Output Format:}

Your response must be a single line of text, strictly conforming to the format below. Do not include any other words, notes, or explanations.

\`\ Speed: [Speed Name], Strikespot: [Strikespot Name].\`\ 
\end{systempromptbox}

\begin{userpromptbox}{\pooltoollogo: User Prompt}
    Analyze the scene and determine the best way to pocket the 9-ball into any pocket while avoiding potting the cue ball. Consider the cue ball's position for good follow-up. Provide the optimal speed and strikespot. Then provide the corresponding action that matches the format.
\end{userpromptbox}

\subsection{Experiment Results}

\begin{table*}[h!]
\centering
\caption{Benchmark Pooltool. Model performance comparison across 15 attempts. Prompt Format: VLA.}
\label{tab:model_performance_comparison_pooltool_vla}
\resizebox{0.8\linewidth}{!}{
\begin{tabular}{l | cccccl | l}
\toprule
\textbf{Model} & \textbf{Att. 1} & \textbf{Att. 3} & \textbf{Att. 6} & \textbf{Att. 9} & \textbf{Att. 12} & \textbf{Att. 15} & \textbf{Avg. Att.} \\
\midrule
\texttt{\textit{MOCK}} & 2.33\% & 10.00\% & 18.33\% & 31.33\% & 40.00\% & 48.00\% & 7.88 \\
\midrule
\rowcolor{lightgray}\multicolumn{8}{c}{\textbf{Open-Source Models}} \\ \midrule
\rowcolor{mylightgray}\multicolumn{8}{c}{\textbf{Qwen2.5-VL Series}} \\
\textbf{Qwen-3B} & 0.00\% & 50.00\% & 50.00\% & 50.00\% & 50.00\% & 50.00\% \xmark & 2.00 \xmark \\
\textbf{Qwen-7B} & 23.50\% & 23.50\% & 23.50\% & 23.50\% & 23.50\% & 26.50\% & 2.58 \firstlogo \\
\textbf{Qwen-32B} & 0.00\% & 0.00\% & 0.00\% & 14.29\% & 14.29\% & 14.29\% & 7.00 \\
\textbf{Qwen-72B} & 0.00\% & 0.00\% & 0.00\% & 8.50\% & 11.00\% & 18.00\% & 10.75 \\
\midrule
\rowcolor{lightgray}\multicolumn{8}{c}{\textbf{Closed-Source Models}} \\ \midrule
\rowcolor{mylightgray}\multicolumn{8}{c}{\textbf{Claude Series}} \\
\textbf{Claude 3.5 Sonnet} & 0.00\% & 11.50\% & 32.00\% & 52.00\% & 59.50\% & 67.00\% & 7.04 \\
\textbf{Claude 3.7 Sonnet} & 0.00\% & 28.50\% & 69.00\% & 69.50\% & 70.00\% & 72.50\% \secondlogo & 3.64 \secondlogo\\
\textbf{Claude 4.0 Sonnet} & 0.00\% & 3.00\% & 6.50\% & 14.50\% & 33.00\% & 44.00\% & 10.30 \\
\textbf{Claude 4.0 Opus} & 0.00\% & 1.50\% & 11.00\% & 19.00\% & 31.50\% & 49.00\% & 9.98 \\
\midrule
\rowcolor{mylightgray}\multicolumn{8}{c}{\textbf{Gemini Series}} \\
\textbf{Gemini-2.0-Flash} & 0.00\% & 0.00\% & 42.00\% & 52.50\% & 54.00\% & 75.00\% \firstlogo & 7.51 \\
\textbf{Gemini-2.5-Pro} & 36.50\% & 38.50\% & 51.00\% & 57.50\% & 62.00\% & 68.00\% & 4.17 \\
\textbf{Gemini-2.5-Flash} & 0.00\% & 9.00\% & 15.50\% & 19.50\% & 23.50\% & 29.00\% & 7.19 \\
\midrule
\rowcolor{mylightgray}\multicolumn{8}{c}{\textbf{GPT Series}} \\
\textbf{GPT-4V} & 0.00\% & 1.50\% & 7.00\% & 15.50\% & 28.50\% & 39.50\% & 10.16 \\
\textbf{GPT-4o-mini} & 0.00\% & 0.00\% & 0.00\% & 100.00\% & 100.00\% & 100.00\% \xmark & 7.50 \xmark \\
\textbf{GPT-4o} & 0.00\% & 0.00\% & 0.50\% & 14.50\% & 28.50\% & 34.50\% & 9.86 \\
\textbf{GPT-o3} & 0.00\% & 4.67\% & 10.00\% & 13.00\% & 19.67\% & 25.67\% & 8.64 \\
\textbf{GPT-o4-mini} & 0.00\% & 10.00\% & 21.00\% & 30.00\% & 40.00\% & 53.00\% & 8.47 \\
\bottomrule
\end{tabular}
}
\end{table*}

\begin{table*}[h!]
\centering
\caption{Benchmark Pooltool. Model performance comparison across 15 attempts. Prompt Format: WM.}
\label{tab:model_performance_comparison_pooltool_wm}
\resizebox{0.8\linewidth}{!}{
\begin{tabular}{l | cccccl | l}
\toprule
\textbf{Model} & \textbf{Att. 1} & \textbf{Att. 3} & \textbf{Att. 6} & \textbf{Att. 9} & \textbf{Att. 12} & \textbf{Att. 15} & \textbf{Avg. Att.} \\
\midrule
\texttt{\textit{MOCK}} & 2.33\% & 10.00\% & 18.33\% & 31.33\% & 40.00\% & 48.00\% & 7.88 \\
\midrule
\rowcolor{lightgray}\multicolumn{8}{c}{\textbf{Open-Source Models}} \\ \midrule
\rowcolor{mylightgray}\multicolumn{8}{c}{\textbf{Qwen2.5-VL Series}} \\
\textbf{Qwen-3B} & 0.00\% & 0.00\% & 0.00\% & 0.00\% & 0.00\% & 100.00\% \xmark & 14.00 \xmark \\
\textbf{Qwen-7B} & 0.00\% & 0.00\% & 0.00\% & 0.00\% & 16.50\% & 40.00\% & 13.76 \\
\textbf{Qwen-32B} & 0.00\% & 0.00\% & 12.96\% & 17.13\% & 23.61\% & 30.56\% & 8.43 \\
\textbf{Qwen-72B} & 0.00\% & 4.00\% & 12.50\% & 24.50\% & 39.00\% & 50.50\% & 9.53 \\
\midrule
\rowcolor{lightgray}\multicolumn{8}{c}{\textbf{Closed-Source Models}} \\ \midrule
\rowcolor{mylightgray}\multicolumn{8}{c}{\textbf{Claude Series}} \\
\textbf{Claude 3.5 Sonnet} & 0.00\% & 14.00\% & 43.00\% & 51.00\% & 64.50\% & 73.00\% \firstlogo & 7.34 \secondlogo \\
\textbf{Claude 3.7 Sonnet} & 0.00\% & 13.00\% & 19.50\% & 34.00\% & 42.50\% & 48.00\% & 7.39 \\
\textbf{Claude 4.0 Sonnet} & 0.00\% & 4.50\% & 16.00\% & 36.00\% & 52.00\% & 63.00\% \secondlogo & 8.93 \\
\textbf{Claude 4.0 Opus} & 0.00\% & 0.00\% & 2.50\% & 21.50\% & 36.50\% & 51.50\% & 10.30 \\
\midrule
\rowcolor{mylightgray}\multicolumn{8}{c}{\textbf{Gemini Series}} \\
\textbf{Gemini-2.0-Flash} & 0.00\% & 19.50\% & 37.50\% & 49.00\% & 64.00\% & 73.00\% \firstlogo & 6.86 \firstlogo\\
\textbf{Gemini-2.5-Flash} & 0.00\% & 7.92\% & 21.78\% & 28.71\% & 34.65\% & 41.58\% & 7.29 \\
\textbf{Gemini-2.5-Pro} & 0.00\% & 6.50\% & 14.50\% & 24.50\% & 33.00\% & 43.00\% & 8.58 \\
\midrule
\rowcolor{mylightgray}\multicolumn{8}{c}{\textbf{GPT Series}} \\
\textbf{GPT-4V} & 0.00\% & 6.50\% & 13.50\% & 21.50\% & 30.50\% & 42.50\% & 9.09 \\
\textbf{GPT-4o-mini} & 0.00\% & 50.00\% & 50.00\% & 50.00\% & 50.00\% & 50.00\% \xmark & 2.00 \xmark \\
\textbf{GPT-4o} & 0.00\% & 0.50\% & 1.00\% & 7.50\% & 24.00\% & 40.00\% & 11.66 \\
\textbf{GPT-o3} & 0.00\% & 4.67\% & 12.67\% & 19.00\% & 30.33\% & 42.00\% & 9.25 \\
\textbf{GPT-o4-mini} & 0.00\% & 11.00\% & 14.00\% & 23.00\% & 31.00\% & 43.00\% & 8.65 \\
\bottomrule
\end{tabular}
}
\end{table*}

The results from the Pooltool environment present a deceptive illusion of competence, where high-level success metrics mask a fundamental lack of deep physical reasoning. While several top-tier models seem to achieve high success rates, a closer analysis of the results reveals that this performance stems not from strategic insight, but from a reliance on simplistic, brute-force heuristics.

As shown in Table \ref{tab:model_performance_master_summary}, and detailed in Tables \ref{tab:model_performance_comparison_pooltool_vla} \& \ref{tab:model_performance_comparison_pooltool_wm}, the 100\% success rates of GPT-4o-mini (VLA) and Qwen-3B (WM) are particularly striking, seemingly surpassing the human player in efficiency with fewer average attempts. However, this is a misleading artifact. 
Our investigation reveals that with the temperature set to 0.1, GPT-4o-mini's output became deterministic, consistently producing the same action (Speed: ``Medium", Strikespot: ``Top Spin") for every attempt. This rigid behavior resulted in the model succeeding on the eighth attempt in every single trial. In stark contrast, other successful models (those not marked with an \xmark) achieved victory by dynamically adjusting their shot strategies across attempts. This clearly demonstrates that the models with fixed outputs lack genuine physical reasoning ability. This exposes the core failure of all tested models in this environment: a complete inability to grasp the nuanced physics of billiards.

Furthermore, the comparison between VLA and WM prompt format again underscores the disconnect between descriptive knowledge and procedural control. For most models, the WM prompt failed to yield any significant improvement and often degraded performance (e.g., Gemini-2.5-Pro dropped from 68.00\% with VLA to 43.00\% with WM).

\section{Benchmark: \angrybirdlogo} \label{appendix:angrybirds}

The core objective of the physiscs-based puzzle game \textit{\textbf{Angry Birds}}\footnote{https://apps.apple.com/us/app/rovio-classics-angry-birds/id1596736236} game is to eliminate all the green pigs. This requires an agent to launch a limited number of birds from a slingshot, with potentially different abilities, to hit the pigs or destroy the structures in which they hide. The task not only tests the grasp of projectile motion, but also demands a deep understanding of structural mechanics and the strategic application of each bird's unique abilities. Agents must be able to accurately analyze the visual scene, identify structural weaknesses, predict the bird's trajectory, and plan the appropriate launch sequence to destroy all the pigs. In this benchmark, we have selected 34 levels from the first chapter of the game.

\subsection{Conversion Details}

\textbf{Observation Space:} A screenshot of the game serves as the direct visual input for the Agent, representing the current state of the level.

\begin{figure}[h]
    \centering
    \begin{minipage}[c]{0.54\linewidth}
        \includegraphics[width=\linewidth]{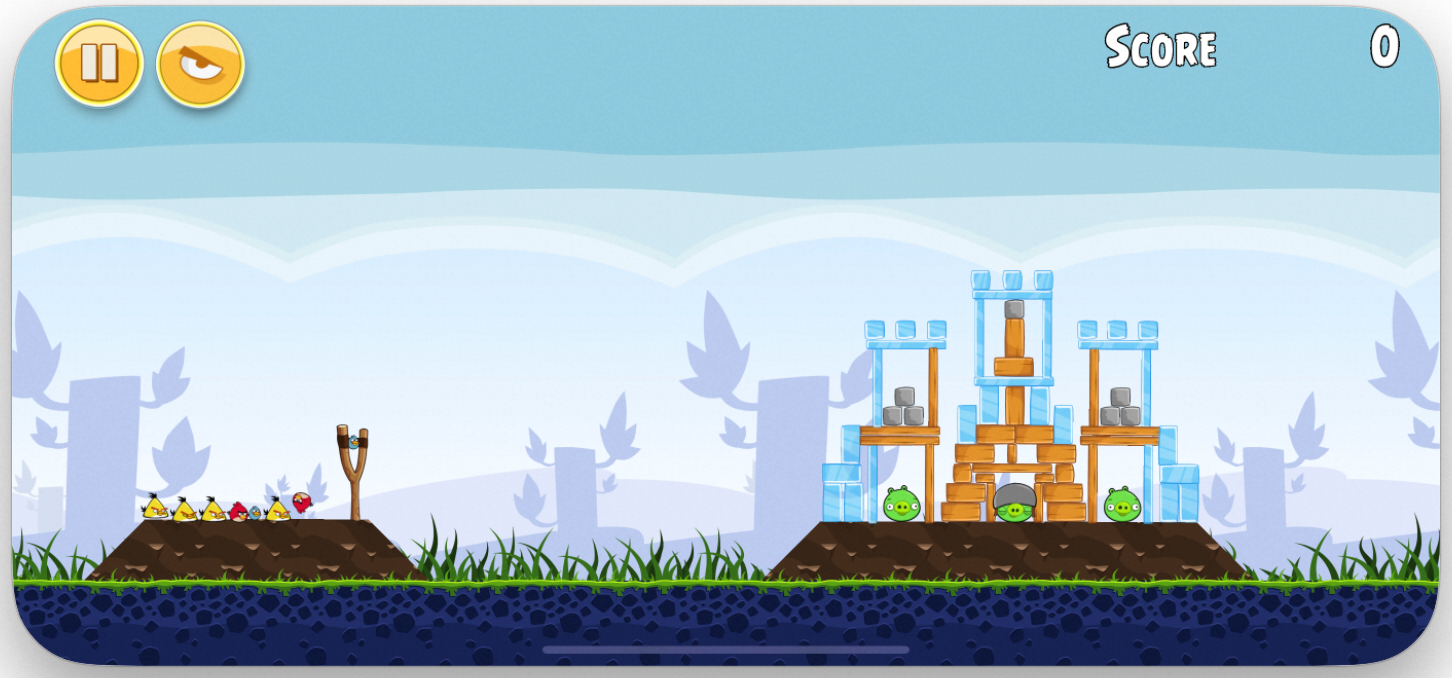}
    \end{minipage}
    \hfill 
    \begin{minipage}[c]{0.45\linewidth}
        \begin{examplebox}{}
        ``level 21": \{

        \quad ``bird\_number": 8,

        \quad ``bird\_type": [``blue", ``red", ``yellow", ``blue", ``red", ``yellow", ``yellow", ``yellow"]

        \}
        \end{examplebox}
    \end{minipage}
    
    \caption{Visual scene of Angry Birds - Level 21 (left) and its corresponding data representation (right).}
    \label{fig:angrybirds_level21_combined}
\end{figure}

\begin{figure}[h]
    \centering
    \begin{minipage}[c]{0.54\linewidth}
        \includegraphics[width=\linewidth]{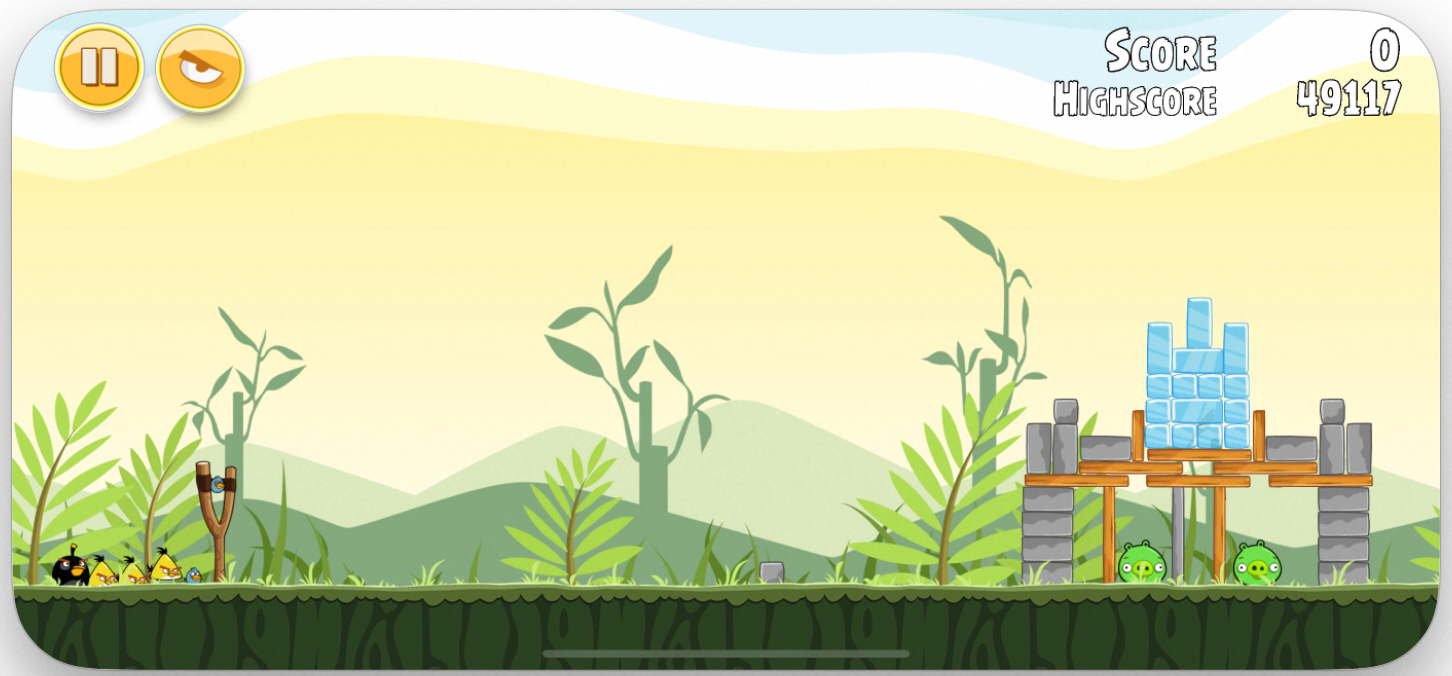}
    \end{minipage}%
    \hfill 
    \begin{minipage}[c]{0.45\linewidth}
        \begin{examplebox}{}
        ``level 33": \{

        \quad ``bird\_number": 6,

        \quad ``bird\_type": [``blue", ``blue", ``yellow", ``yellow", ``yellow", ``black"]
        \}
        \end{examplebox}
    \end{minipage}
    \caption{Visual scene of Angry Birds - Level 33 (left) and its corresponding data representation (right).}
    \label{fig:angrybirds_level33_combined}
\end{figure}

\noindent \textbf{Action Space:} We have annotated the number and types of available birds for each level. The agent's only action is to launch the current bird from the slingshot. For example, the bird configurations for levels 21, 33 are shown in Figures \ref{fig:angrybirds_level21_combined} \& \ref{fig:angrybirds_level33_combined}, respectively.

Each launch required only the specification of an angle of launch (from 0 to 90 degrees\footnote{In the selected 34 game levels, there are no cases requiring the launch of a bird under the horizon or backwards.}) and a power level (from 0.0 to 1.0).

\subsection{Prompts}

\begin{systempromptbox}{\angrybirdlogo: System Prompt}
You are a master strategist for the game \textbf{Angry Birds}. Your goal is to destroy all the pigs using a limited number of birds, launched from a slingshot. You have a deep understanding of projectile motion, structural weaknesses, and the unique abilities of each bird.\\

\textbf{Core Objective:} Eliminate All Pigs

\quad - The level is won when all pigs on the screen have been destroyed.

\quad - The level is lost if you run out of birds before all pigs are eliminated.\\

\textbf{Core Mechanics \& Operations}

\textbf{Shooting:} Your only action is to launch a bird from the slingshot. You must define the \textit{angle} of the shot and the \textit{power} of the launch.

\quad \textbf{angle:} An integer from 0 to 90. 0 is horizontal to the right. 90 is vertically upward. 

\quad \textbf{power:} A float from 0.0 to 1.0, where 1.0 is maximum power (pulling the slingshot back as far as possible).

\quad \textbf{Bird Abilities:} 
Some birds have special abilities that are activated by tapping the screen \textbf{after} they are launched. Your plan should assume the ability will be used optimally. Focus ONLY on the initial launch parameters. \\

\textbf{The Arsenal: Bird Types}

You must know the strengths of each bird to plan your attack.

\quad - \textbf{Red Bird:} The standard bird. Has no special ability. It is best used for direct impact and toppling structures.

\quad - \textbf{Yellow Bird:} Tapping the screen after launch causes it to accelerate in a straight line. It is highly effective against wooden structures.

\quad - \textbf{Blue Birds:} Tapping the screen after launch splits it into three smaller birds. It is extremely effective against glass or ice structures.

\quad - \textbf{Black Bird:} Acts as a bomb. It will explode shortly after impact. It is powerful against stone and can cause massive chain reactions.\\

\textbf{Strategic Principles}

\quad - \textbf{Trajectory is Everything:} Carefully analyze the history of shots, the pigs' location and the surrounding structures to calculate the optimal launch angle and power.

\quad - \textbf{Structural Weakness:} Target the weak points of structures. Removing a key support block can cause a cascade of destruction.

\quad - \textbf{Bird Order:} You must use the birds in the order they are given. Plan your entire strategy around the sequence of available birds.

Your task is to analyze the game state and determine the best single \texttt{shoot} action for the \textbf{current} bird at the slingshot.

\end{systempromptbox}

\begin{userpromptbox}{\angrybirdlogo: User Prompt}
Please analyze the current game screen and plan a single \textit{shoot} action to destroy the pigs.\\

\textbf{Current Level State Analysis:}

\textless\ state\_analysis\_prompt \textgreater \\

\textbf{History of Shots:}

\textless\ history\_prompt\_text \textgreater \\
\textless\ history\_visual\_image \textgreater \\

\textbf{Task Instructions:}

Your goal is to devise the best shot for the \textbf{current bird}. Analyze the structures and pig locations, then provide the launch parameters.

\textless\ action\_instructions\_prompt  \textgreater \\

\textbf{Output Format:}
Adjust the angle and power based on your analysis. The output should be in the format:

Strictly provide only the action code in `[' and `]' brackets.
Do NOT add any other text, reasoning, or comments.
Do NOT output any JSON or other formats.

\end{userpromptbox}

\subsection{Experiment Results}

\begin{table*}[h!]
\centering
\caption{Benchmark Angry Birds. Model performance comparison. Prompt Format: VLA.}
\label{tab:model_performance_comparison_angrybirds_vla}
\resizebox{\linewidth}{!}{
\begin{tabular}{l | lllllll}
\toprule
\textbf{Model} & \textbf{SuccRate} & \textbf{Total\goldstar} & \textbf{Mean\goldstar\ (Overall)} & \textbf{Mean\goldstar\ (Completed)} & \textbf{\# Num. 1\goldstar} & \textbf{\# Num. 2\goldstar} & \textbf{\# Num. 3\goldstar} \\
\midrule

\texttt{\textit{MOCK}}& 17.65\% & 14.00 & 0.41 & 2.33 & 0.00 & 4.00 & 2.00\\
\midrule
\rowcolor{lightgray}\multicolumn{8}{c}{\textbf{Open-Source Models}} \\ \midrule

\textbf{Qwen-3B} & 17.65\% & 13.00 & 0.38 & 2.17 & 1.00 & 3.00 & 2.00\\
\textbf{Qwen-7B} & 20.59\% & 13.00 & 0.38 & 1.86 & 3.00 & 2.00 & 2.00\\
\textbf{Qwen-32B} & 26.47\% & 14.00 & 0.41 & 1.56 & 6.00 & 1.00 & 2.00\\
\textbf{Qwen-72B} & 29.41\% & 23.00 & 0.68 & 2.30 & 1.00 & 5.00 & 4.00\\
\midrule
\rowcolor{lightgray}\multicolumn{8}{c}{\textbf{Closed-Source Models}} \\ \midrule

\rowcolor{mylightgray}\multicolumn{8}{c}{\textbf{Claude Series}} \\

\textbf{Claude 3.5 Sonnet} & 26.47\% & 20.00 & 0.59 & 2.22 & 2.00 & 3.00 & 4.00\\
\textbf{Claude 3.7 Sonnet} & 41.18\% \firstlogo & 27.00 & 0.79 & 1.93 & 4.00 & 7.00 & 3.00\\
\textbf{Claude 4.0 Sonnet} & 35.29\% \secondlogo & 24.00 & 0.71 & 2.00 & 5.00 & 2.00 & 5.00\\
\textbf{Claude 4.0 Opus} & 32.35\% & 23.00 & 0.68 & 2.09 & 3.00 & 4.00 & 4.00\\

\midrule
\rowcolor{mylightgray}\multicolumn{8}{c}{\textbf{Gemini Series}} \\

\textbf{Gemini-2.0-Flash} & 20.59\% & 16.00 & 0.47 & 2.29 & 1.00 & 3.00 & 3.00\\
\textbf{Gemini-2.5-Flash} & 29.41\% & 22.00 & 0.65 & 2.20 & 2.00 & 4.00 & 4.00\\
\textbf{Gemini-2.5-Pro} & 35.29\% \secondlogo & 25.00 & 0.74 & 2.08 & 5.00 & 1.00 & 6.00\\
\midrule
\rowcolor{mylightgray}\multicolumn{8}{c}{\textbf{GPT Series}} \\

\textbf{GPT-4V} & 29.41\% & 21.00 & 0.62 & 2.10 & 2.00 & 5.00 & 3.00\\
\textbf{GPT-4o-mini} & 23.53\% & 14.00 & 0.41 & 1.75 & 4.00 & 2.00 & 2.00\\
\textbf{GPT-4o} & 32.35\% & 24.00 & 0.71 & 2.18 & 3.00 & 3.00 & 5.00\\
\textbf{GPT-o3} & 35.29\% \secondlogo & 24.00 & 0.71 & 2.00 & 4.00 & 4.00 & 4.00\\
\textbf{GPT-o4-mini} & 32.35\% & 22.00 & 0.65 & 2.00 & 3.00 & 5.00 & 3.00\\ 
\midrule
\rowcolor{mylightgray}\textbf{Human} & 64.71\% & 44.00 & 1.29 & 2.00 & 6.00 & 10.00 & 6.00\\
\bottomrule
\end{tabular}
}
\end{table*}

The results from the Angry Birds benchmark reveal a substantial performance gap between current VLMs and human-level strategic physical reasoning. As detailed in Tables \ref{tab:model_performance_master_summary} and \ref{tab:model_performance_comparison_angrybirds_vla}, even the best-performing model (Claude 3.7 Sonnet) achieves a success rate of only 41.18\%, which is significantly lower than the non-expert human benchmark of 64.71\%. This disparity underscores a fundamental deficit in the models' ability to master environments that require a synthesis of projectile motion prediction and long-term strategic planning.

A deeper analysis of the models' failures indicates that their primary weakness lies not in simple trajectory calculation, but in predicting the complex, cascading consequences of an action. While a model might occasionally succeed with a direct hit on an exposed pig, they consistently struggle with the core strategic element of the game: causing large-scale structural collapse through chain reactions. Human players excel by identifying critical structural weaknesses, e.g., a single block whose removal will cause a tower to topple, and planning shots that maximize such destruction.

However, the analysis of star collection offers a more nuanced insight. While the models' overall success rate is low, the Mean \goldstar\ (Completed) metric shows that when they do succeed, the quality of their solution is often respectable. Several top models, such as Qwen-72B (2.30) and Claude 3.5 Sonnet (2.22), achieve a mean star count on par with or even slightly higher than the human player (2.00) for completed tasks. This suggests that their failures are not due to an inability to generate precise outputs, but rather a lack of a reliable physical intuition. When a model's internal simulation happens to align with the puzzle's specific physics, it can produce a near-optimal solution. 

In conclusion, the Angry Birds environment effectively exposes the limitations of current VLMs in multi-stage, dynamic physical reasoning. The task requires more than just perceiving the scene; it demands a predictive understanding of how one action radically alters the environment for all subsequent actions. The models' struggles to plan for chain reactions and adapt their strategy across a sequence of shots highlight a critical gap between descriptive knowledge (knowing a bird's ability) and procedural, predictive control (knowing precisely where and how to launch it for maximum effect).

\section{Benchmark: \cutrepologo} \label{appendix:rope}

The core objective of the \textbf{\textit{Cut the Rope}}\footnote{https://apps.apple.com/cn/app/cut-the-rope/id1024507512} physics-based puzzle game environment is to feed a piece of candy to a small green monster named \textit{Om Nom}. This requires agents to cut ropes at precise moments and interact with a variety of tools, such as air cushions and bubbles, so that candies reach the monster. This task not only tests the grasp of timing and sequencing, but also demands a deep understanding of physical laws like gravity, momentum, inertia, and trajectory. Agents must be able to accurately interpret visual scenes, identify the functions and potential risks of different props, and make precise predictions about dynamic processes such as swinging and falling, ultimately planning the optimal path to successfully deliver the candy pieces, while (preferably) collecting all three stars per level.

\subsection{Conversion Details}

\begin{figure*}[h]
    \centering
    \begin{subfigure}[b]{0.19\textwidth}
        \centering
        \includegraphics[trim={0cm 0cm 0cm 1cm}, clip, width=\linewidth]{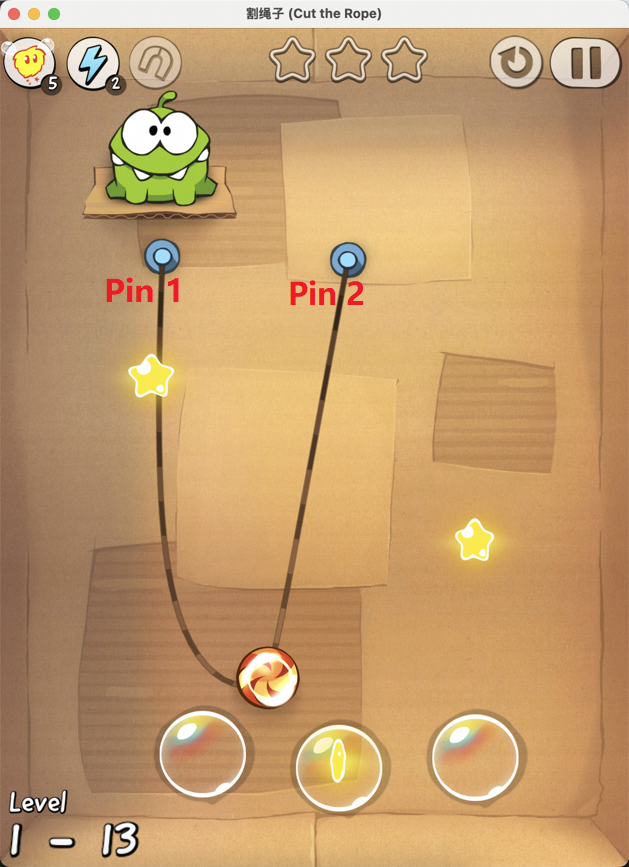}
    \end{subfigure}
    \begin{subfigure}[b]{0.19\textwidth}
        \centering
        \includegraphics[trim={0cm 0cm 0cm 1cm}, clip, width=\linewidth]{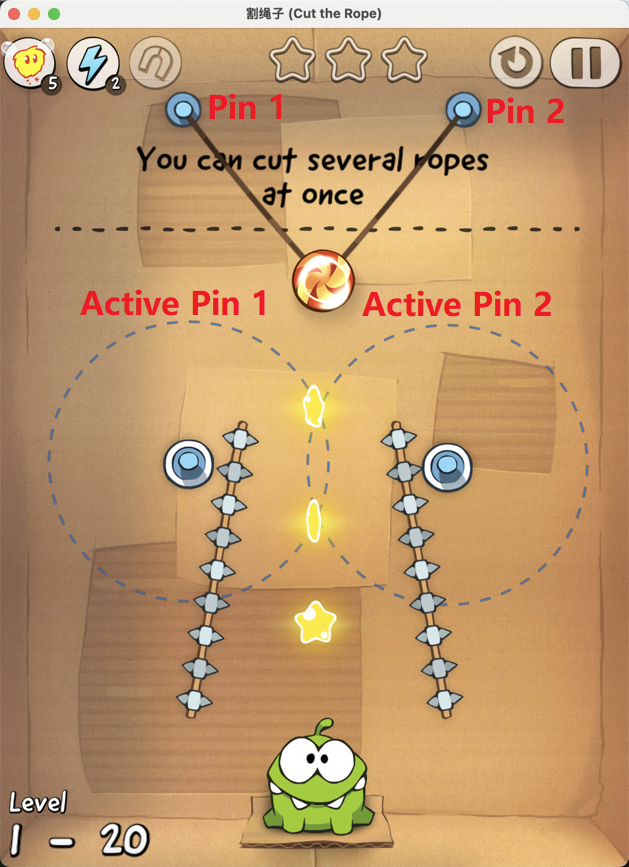}
    \end{subfigure}
    \begin{subfigure}[b]{0.19\textwidth}
        \centering
        \includegraphics[trim={0cm 0cm 0cm 1cm}, clip, width=\linewidth]{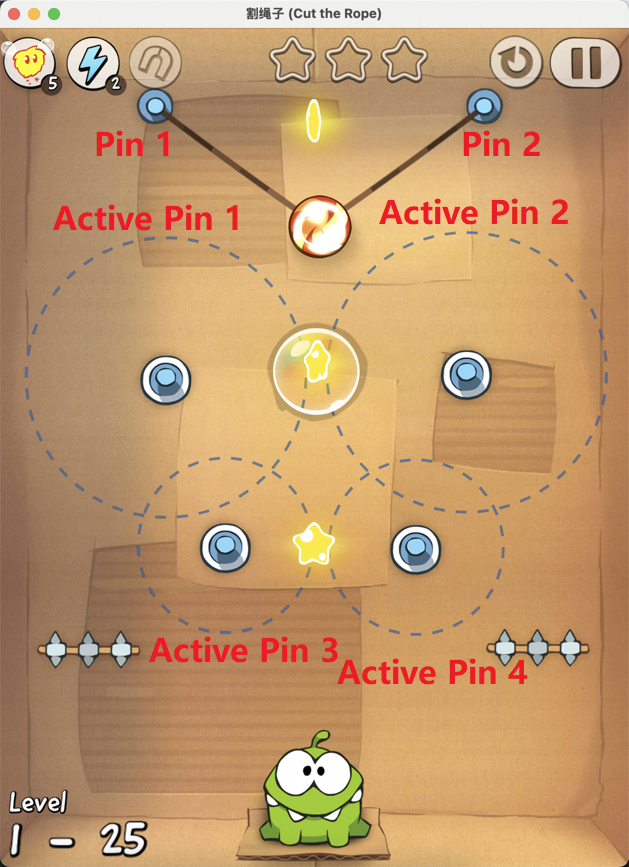}
    \end{subfigure}
    \begin{subfigure}[b]{0.19\textwidth}
        \centering
        \includegraphics[trim={0cm 0cm 0cm 1cm}, clip, width=\linewidth]{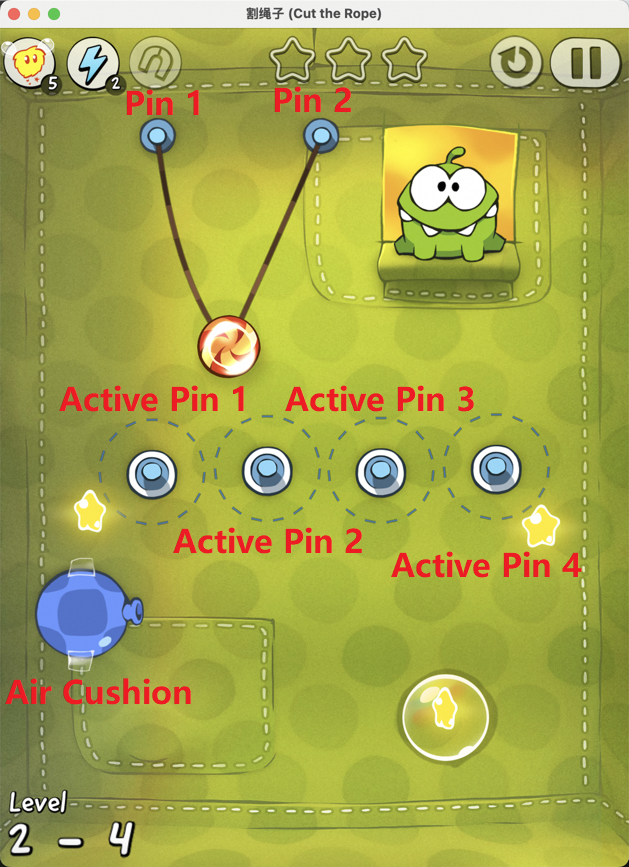}
    \end{subfigure}
    \begin{subfigure}[b]{0.19\textwidth}
        \centering
        \includegraphics[trim={0cm 0cm 0cm 1cm}, clip, width=\linewidth]{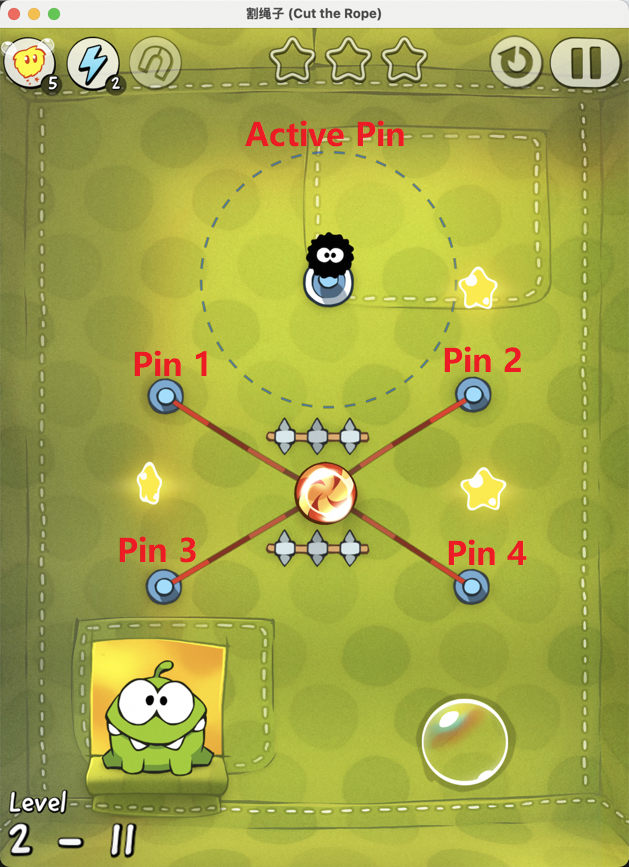}
    \end{subfigure}
    \begin{subfigure}[b]{0.19\textwidth}
        \centering
        \includegraphics[trim={0cm 0cm 0cm 1cm}, clip, width=\linewidth]{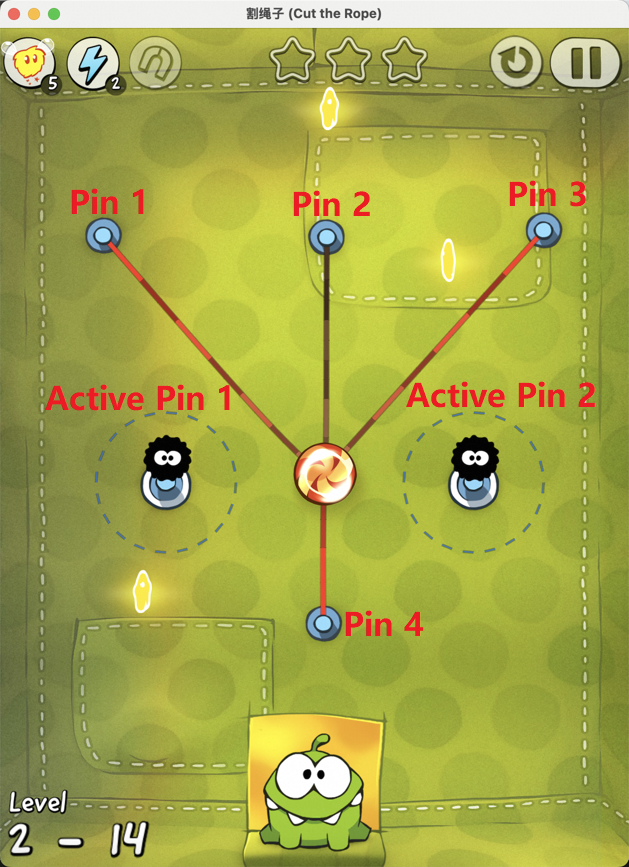}
    \end{subfigure}
    \begin{subfigure}[b]{0.19\textwidth}
        \centering
        \includegraphics[trim={0cm 0cm 0cm 1cm}, clip, width=\linewidth]{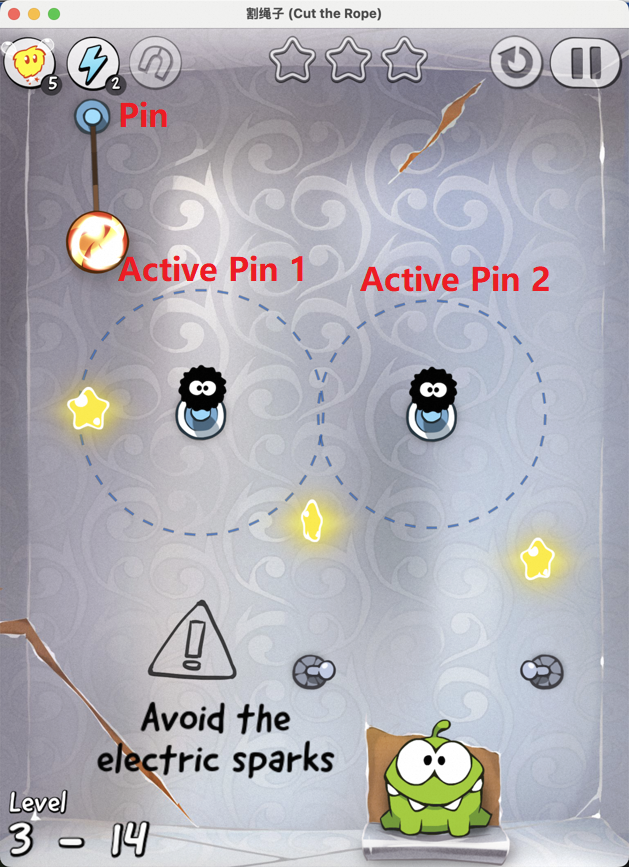}
    \end{subfigure}
    \begin{subfigure}[b]{0.19\textwidth}
        \centering
        \includegraphics[trim={0cm 0cm 0cm 1cm}, clip, width=\linewidth]{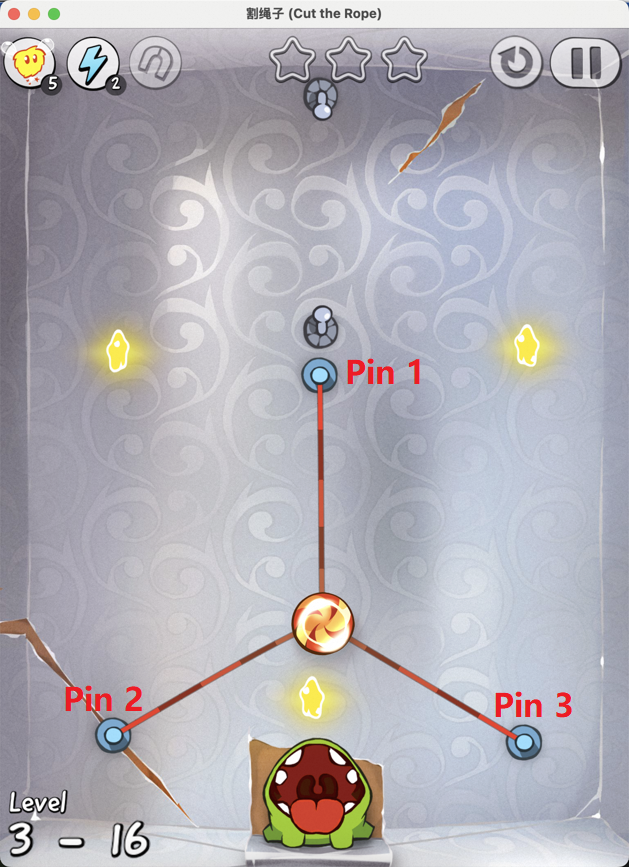}
    \end{subfigure}
    \begin{subfigure}[b]{0.19\textwidth}
        \centering
        \includegraphics[trim={0cm 0cm 0cm 1cm}, clip, width=\linewidth]{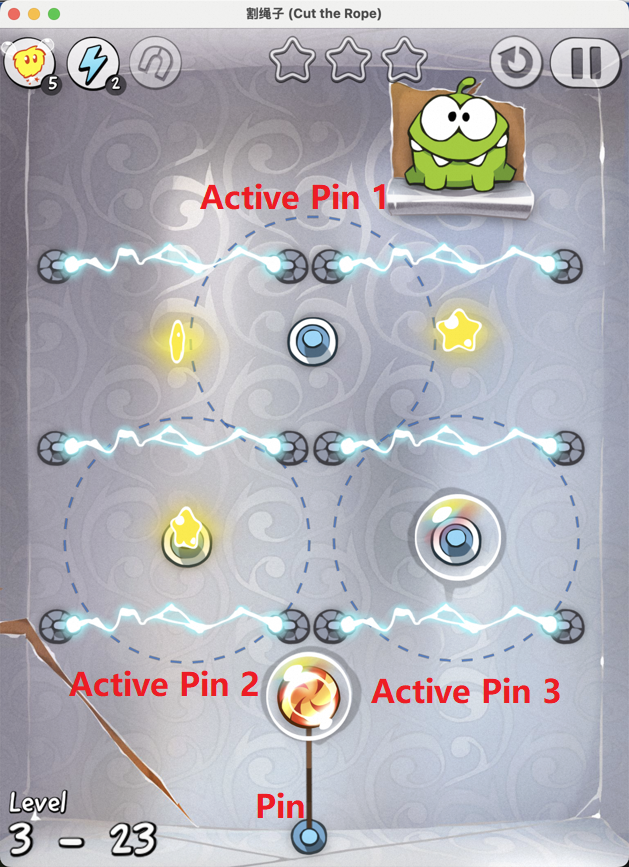}
    \end{subfigure}
    \begin{subfigure}[b]{0.19\textwidth}
        \centering
        \includegraphics[trim={0cm 0cm 0cm 1cm}, clip, width=\linewidth]{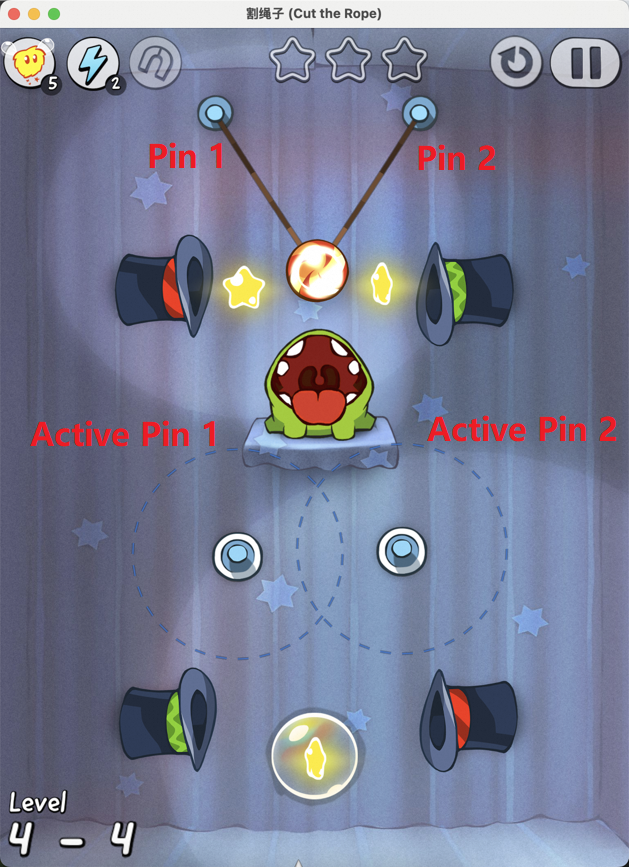}
    \end{subfigure}
    \begin{subfigure}[b]{0.19\textwidth}
        \centering
        \includegraphics[trim={0cm 0cm 0cm 1cm}, clip, width=\linewidth]{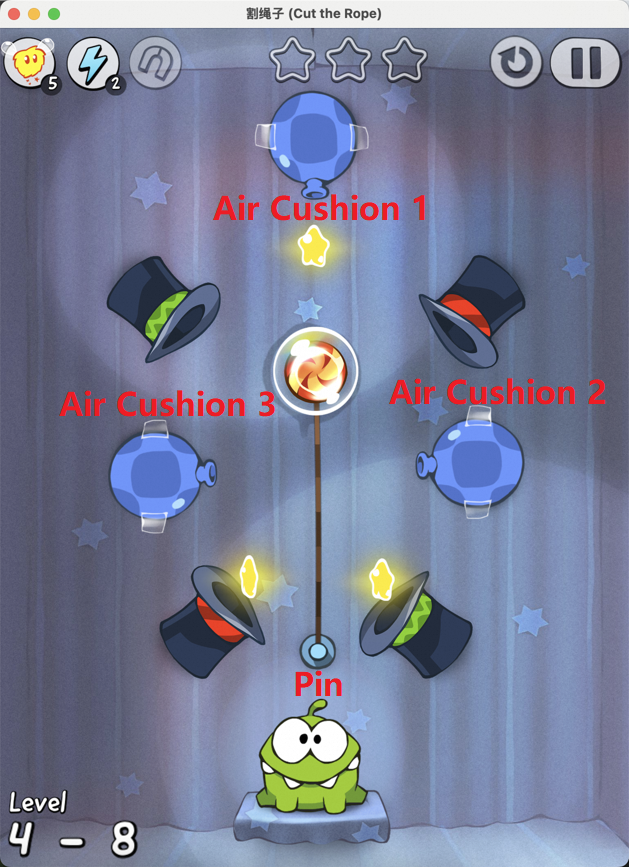}
    \end{subfigure}
    \begin{subfigure}[b]{0.19\textwidth}
        \centering
        \includegraphics[trim={0cm 0cm 0cm 1cm}, clip, width=\linewidth]{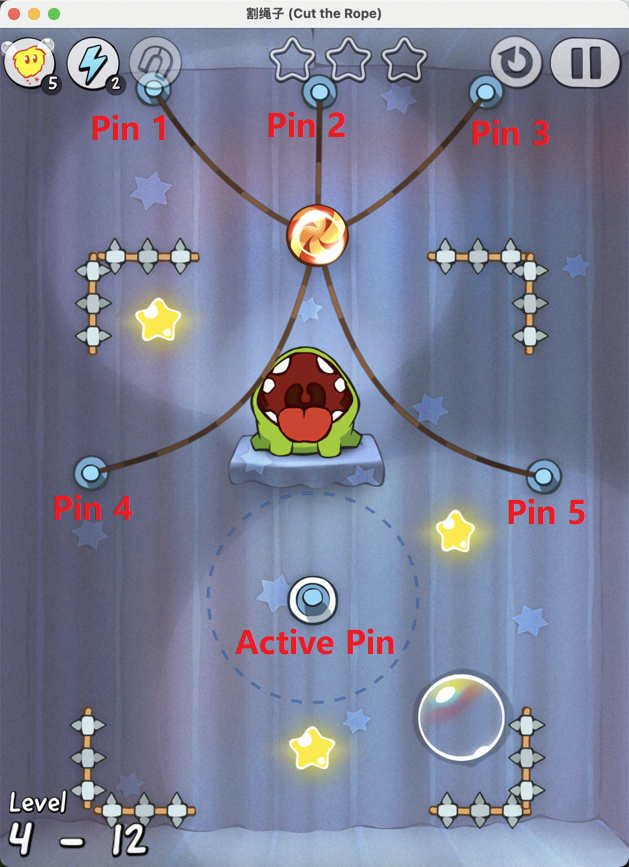}
    \end{subfigure}
    \begin{subfigure}[b]{0.19\textwidth}
        \centering
        \includegraphics[trim={0cm 0cm 0cm 1cm}, clip, width=\linewidth]{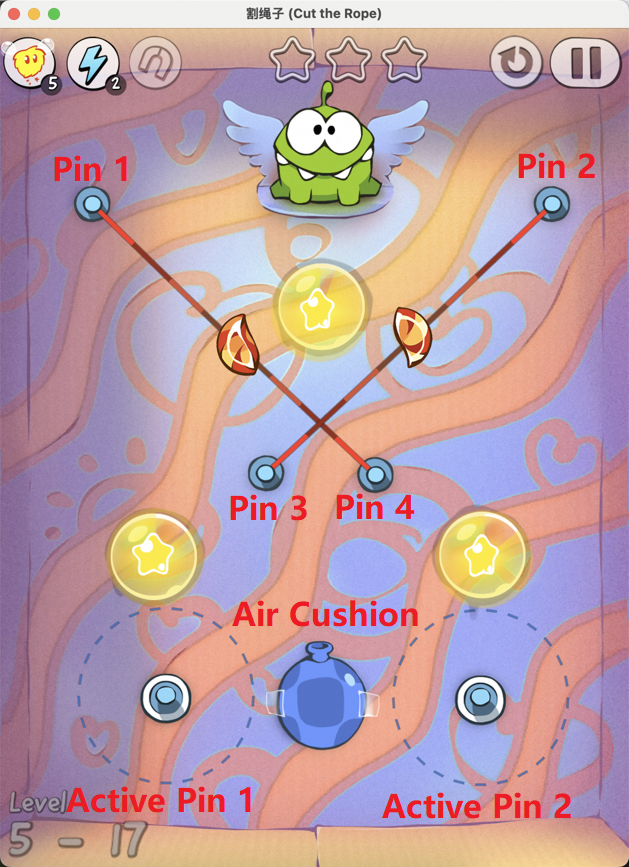}
    \end{subfigure}
    \begin{subfigure}[b]{0.19\textwidth}
        \centering
        \includegraphics[trim={0cm 0cm 0cm 1cm}, clip, width=\linewidth]{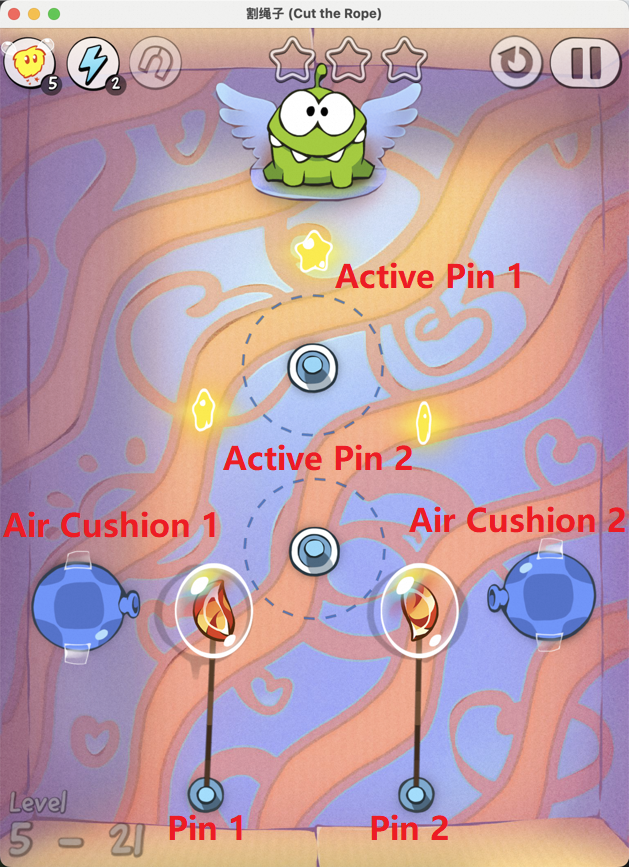}
    \end{subfigure}
    \begin{subfigure}[b]{0.19\textwidth}
        \centering
        \includegraphics[trim={0cm 0cm 0cm 1cm}, clip, width=\linewidth]{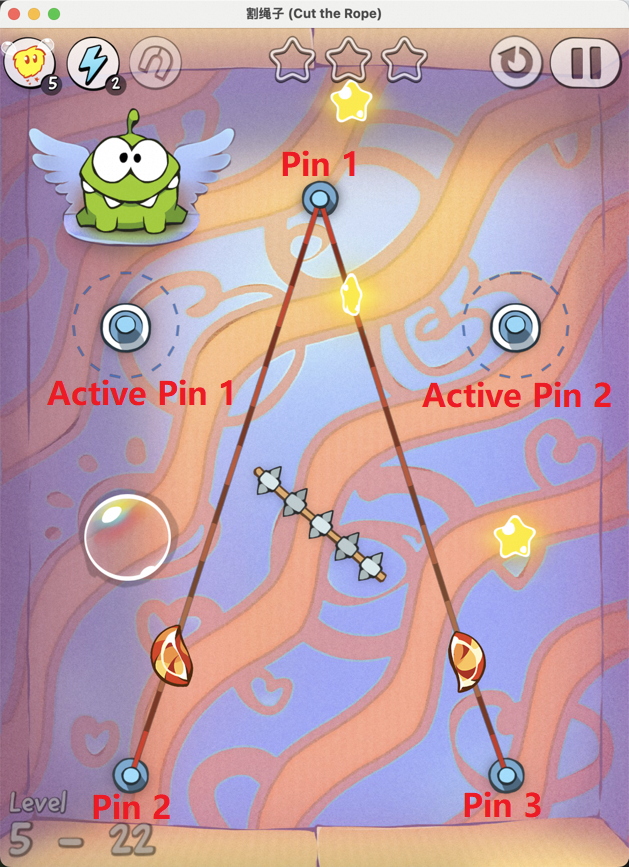}
    \end{subfigure}
    
    \caption{Examples of static element annotation in the \textit{Cut the Rope} game. The figure displays screenshots from various levels where key static props—such as Pins, Active Pins, and Air Cushions—are clearly marked with numerical IDs. This method converts pixel-level visual information, enabling Agent to accurately identify and locate each key object within the game world.}
\label{fig:cutrope_static_annotated}
\end{figure*}

\begin{figure*}[h]
    \centering
    \begin{subfigure}[b]{0.19\textwidth}
        \centering
        \includegraphics[trim={0cm 0cm 0cm 1cm}, clip, width=\linewidth]{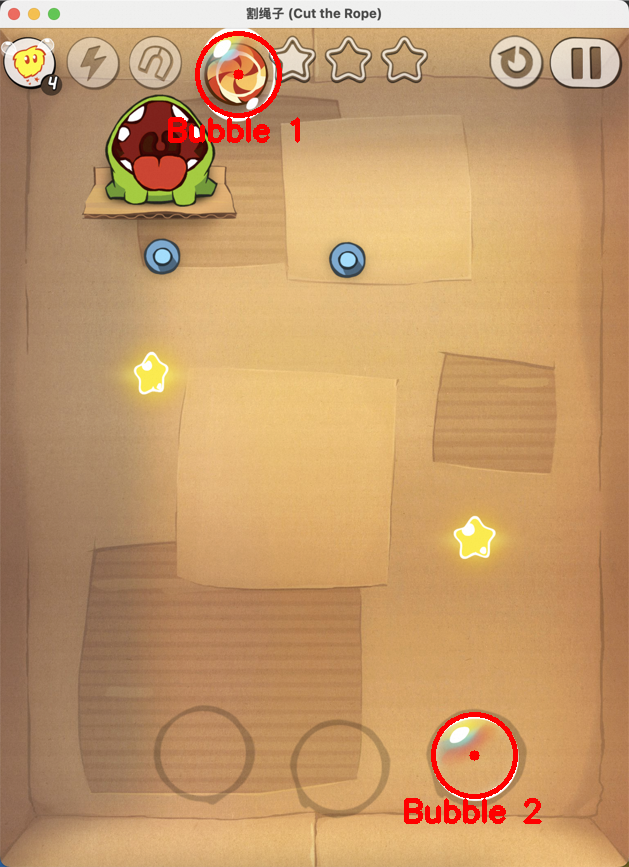}
    \end{subfigure}
    \begin{subfigure}[b]{0.19\textwidth}
        \centering
        \includegraphics[trim={0cm 0cm 0cm 1cm}, clip, width=\linewidth]{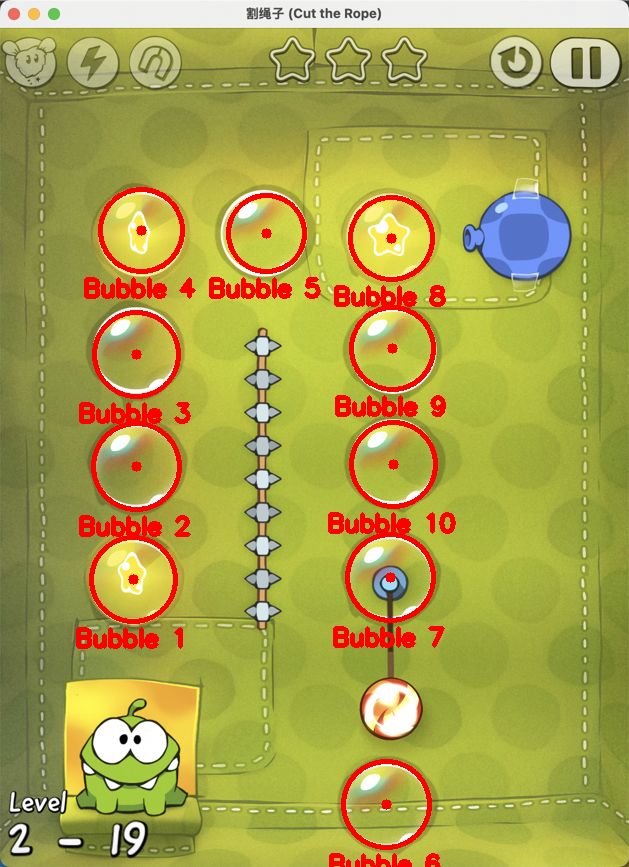}
    \end{subfigure}
    \begin{subfigure}[b]{0.19\textwidth}
        \centering
        \includegraphics[trim={0cm 0cm 0cm 1cm}, clip, width=\linewidth]{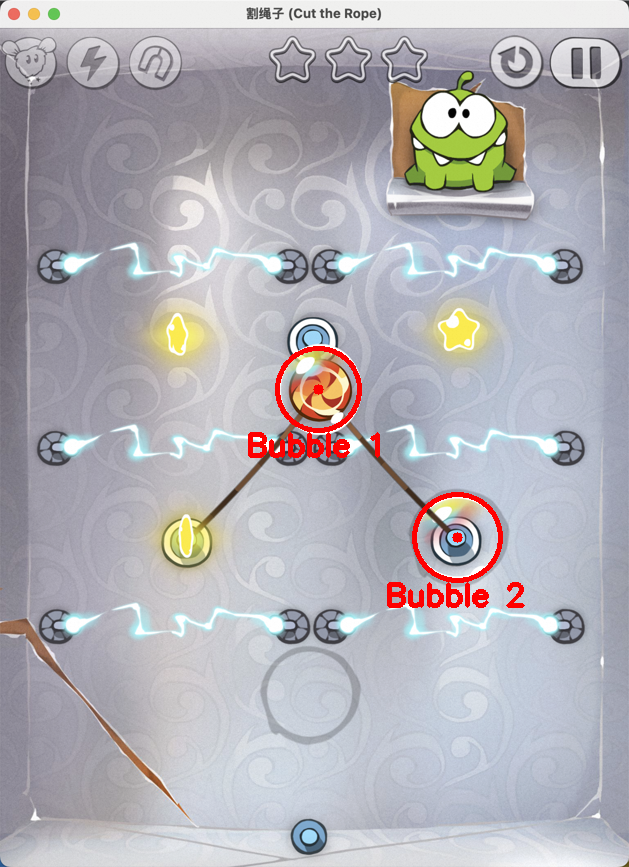}
    \end{subfigure}
    \begin{subfigure}[b]{0.19\textwidth}
        \centering
        \includegraphics[trim={0cm 0cm 0cm 1cm}, clip, width=\linewidth]{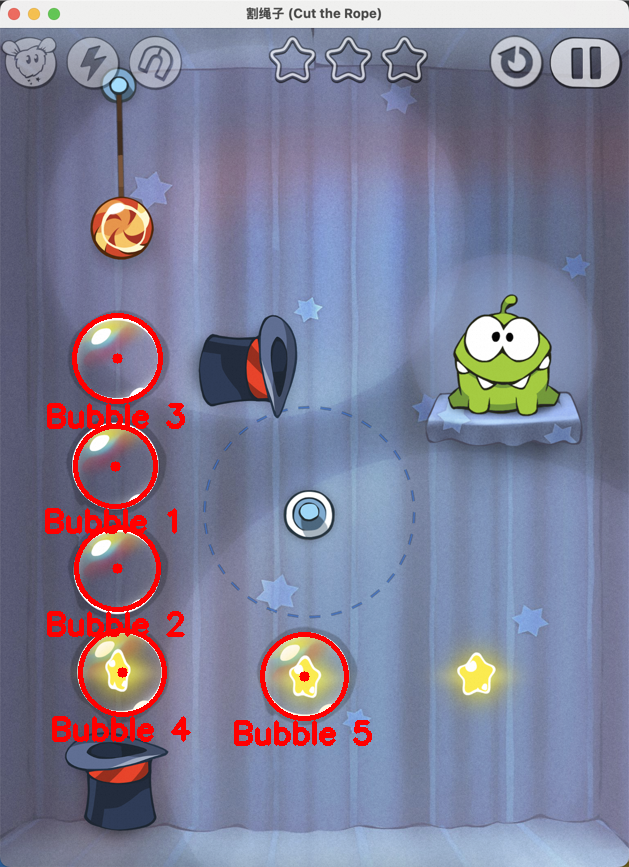}
    \end{subfigure}
    \begin{subfigure}[b]{0.19\textwidth}
        \centering
        \includegraphics[trim={0cm 0cm 0cm 1cm}, clip, width=\linewidth]{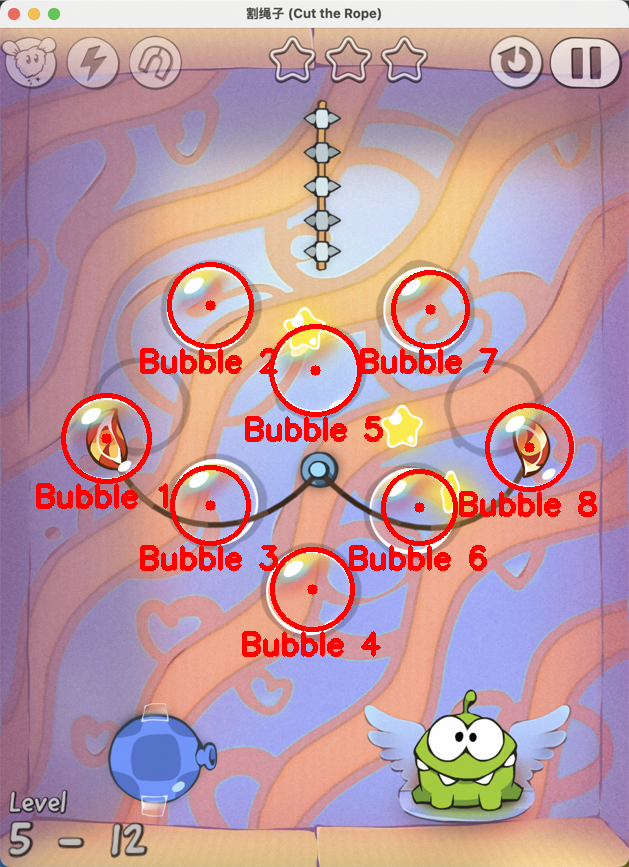}
    \end{subfigure}
    
    \caption{Examples of real-time annotation for the dynamic element (the bubble). The figure clearly shows that as the bubble moves across the scene.}
\label{fig:cutrope_bubble_annotated}
\end{figure*}

\begin{figure}[h]
\centering
\includegraphics[width=0.6\linewidth]{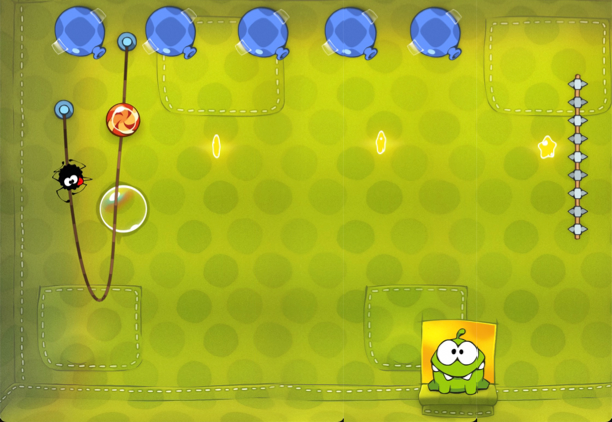}
\caption{An example of a level with multi-screen transitions. In these cases, the positions of game elements shift as the screen moves, making their locations inconsistent and difficult to track. As this prevents reliable element recognition across all levels, we have excluded such scenarios from our benchmark.}
\label{fig:cutropes_multi_screen}
\end{figure}

\textbf{Observation Space:} To enable agentic VLMs to understand and operate within the game environment, we have converted the in-game elements and interaction processes.
For static props, such as Pins, Active Pins, and Air Cushions, we enhance agent's spatial awareness of these key elements by annotating their physical locations on game screenshots (as shown in Figure \ref{fig:cutrope_static_annotated}). 
For dynamic elements, such as the Bubble enclosing the candy, we use an OpenCV-based method for real-time recognition to ensure agents can promptly capture its state and interact with it (as shown in Figure \ref{fig:cutrope_bubble_annotated}).
In terms of execution flow, after an agent completes an action, the game is paused, and a screenshot is taken. This screenshot serves as the visual input for the next round of decision-making. Once the agent issues a new command, the game resumes, thus forming a complete ``perception-decision-action" loop.
In this benchmark, we have selected 88 levels from the 125 levels across 5 boxes in the first session (Session 1) of \textit{Cut the Rope}, filtering out those that involve multi-screen transitions (as shown in Figure \ref{fig:cutropes_multi_screen}) and where the position of elements could not be obtained in real time.

\begin{figure}[h]
\centering
\includegraphics[trim={0cm 0cm 0cm 1cm}, clip, width=0.35\linewidth]{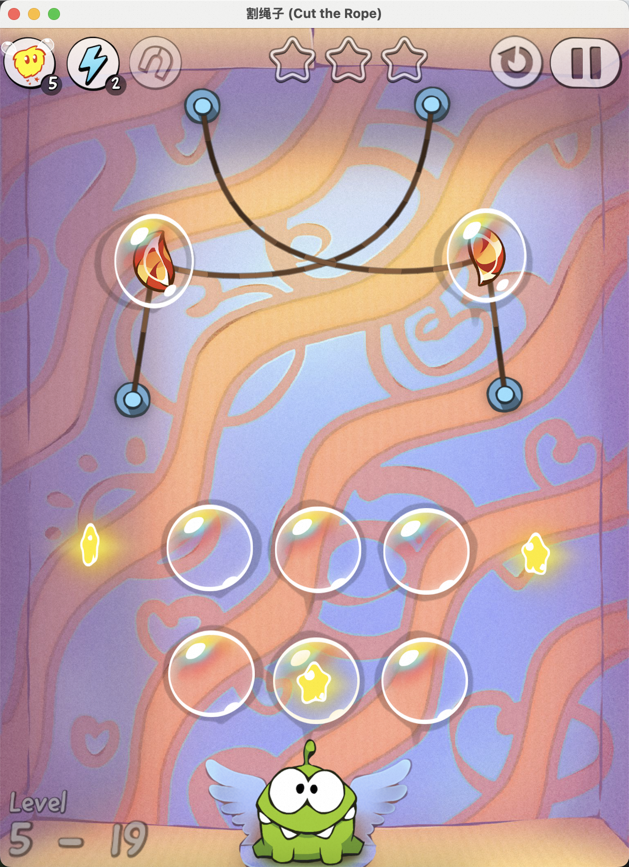}
\caption{An example of a level where VLMs fail to correctly identify the number of ropes.}
\label{fig:cutropes_case_study}
\end{figure}

The rationale for these detailed annotations is that current VLMs lack the ability to analyze the cartoon art style found in games like Cut the Rope. To illustrate this issue, we presented the image in Figure \ref{fig:cutropes_case_study} to high-performing models (Gemini-2.5-Pro and GPT-o3) and asked ``How many ropes are there in the picture? Describe how each rope is connected." As the following examples demonstrate, their responses reveal significant perceptual challenges.

\begin{examplebox}{Model: Gemini-2.5-Pro}
\textbf{Q:} How many ropes are there in the picture? Describe how each rope is connected.

\textbf{\xmark A:} Based on the image provided, there are three ropes in total. Here is a description of how each one is connected:

\begin{itemize}
    \item \textbf{Left Vertical Rope:} This rope is stretched vertically between the top-left blue anchor point and the bottom-left blue anchor point. A bubble containing a red and orange object is attached to this rope.
    \item \textbf{Right Vertical Rope:} This rope is stretched vertically between the top-right blue anchor point and the bottom-right blue anchor point. It also has a bubble with a red and orange object attached to it.
    \item \textbf{Top Crossed Rope:} This rope connects the top-left blue anchor point to the top-right blue anchor point. It hangs loosely in the middle, crossing over itself to form an ``X" shape.
\end{itemize}

\textbf{Q:} Usually, the blue pin and the candy are the endpoints of the ropes. Think about the intersection of the ropes and then answer me.

\textbf{\cmark A:} You are absolutely right! Thank you for the clarification. My initial description was based on a simple visual interpretation, but considering the game's mechanics and the intersection point, it's clear there are four ropes.

Here is the corrected description of how each rope is connected:

1. A rope connects the top-left blue pin to the candy bubble on the right.

2. A rope connects the top-right blue pin to the candy bubble on the left. These two ropes cross each other in the center, forming the intersection.

3. A rope connects the bottom-left blue pin to the candy bubble on the left.

4. A rope connects the bottom-right blue pin to the candy bubble on the right.
\end{examplebox}

\begin{examplebox}{Model: GPT-o3}
\textbf{Q:} How many ropes are there in the picture? Describe how each rope is connected.

\textbf{\xmark A: }There are 4 separate ropes in this level.

1. Upper-left rope – goes from the left-hand top peg to the centre top peg (it droops in the middle).

2. Upper-right rope – goes from the centre top peg to the right-hand top peg (it also droops, crossing the first one).

3. Left vertical rope – hangs straight down from the left flame-bubble to the lower left peg.

4. Right vertical rope – hangs straight down from the right flame-bubble to the lower right peg.\\

\textbf{Q:} Usually, the blue pin and the candy are the endpoints of the ropes. Think about the intersection of the ropes and then answer me.

\textbf{\xmark A:} There are 4 separate ropes in this level.

\begin{itemize}
    \item Two upper ropes: One runs from the top-left to the top-center pin, and the other from the top-center to the top-right pin. They droop and cross in the middle.
    \item Two vertical ropes: One hangs from the left flame-bubble down to the bottom-left pin. The other hangs from the right flame-bubble down to the bottom-right pin.
\end{itemize}

\end{examplebox}

Evidently, even the most advanced VLMs still struggle with problems that are trivial for humans to solve at a glance. This underscores the necessity of our meticulous annotations to enable a fairer and more accurate evaluation.

\noindent \textbf{Action Space:} Our action space is designed to abstract the game's core interactions into a set of discrete and structured commands, enabling agents to execute actions in a clear and unambiguous manner. These actions primarily include:
\begin{itemize}
    \item \textbf{Cutting:} This is the primary interaction. Using commands like \texttt{cut\_pin(id)}, and \texttt{cut\_active\_pin(id)}, agents can sever ropes connected to different types of anchor points.
    \item \textbf{Tool Interaction:} Agents can use \texttt{pop\_bubble(id)} to pop a bubble, causing the candy to fall, or use \texttt{tap\_air\_cushion(id, times)} to activate an air cushion and provide thrust to the candy.
    \item \textbf{Timing Control:} \texttt{sleep(seconds)} is a crucial command that allows agents to pause before executing the next action, waiting for in-game physical processes (like swinging or falling) to unfold to the optimal moment.
    \item \textbf{Termination:} The \texttt{success()} and \texttt{fail()} special commands are used to end the current level attempt when an agent determines that the task is either guaranteed to succeed or has irrevocably failed.
\end{itemize}

Through a strategic combination and precise timing of these actions, agents can tackle the various complex physics puzzles in the game and ultimately achieve its per-level objectives.

\subsection{Prompts}

\begin{systempromptbox}{\cutrepologo: System Prompt}
You are a \textbf{Master of Cut the Rope}, an expert physicist and strategist. Your sole purpose is to analyze each puzzle and devise the perfect plan to feed a delicious candy to a small green monster named Om Nom. You achieve this by applying a deep understanding of physics—gravity, momentum, inertia, and trajectory—to manipulate the environment.

Your analysis and instructions must be precise, logical, and focused on achieving a perfect score.\\

\textbf{Core Objectives}

You must adhere to these objectives in order of priority:

- \textbf{Primary Objective: Feed Om Nom}

\quad - The ultimate goal of every level is to successfully deliver the candy into Om Nom's mouth.

\quad - Combine two pieces of candy before feeding Om Norm!

\quad - Failure Conditions: The level is failed and must be restarted if the candy falls off-screen, is destroyed by hazards (like spikes), or is stolen by a spider.

- \textbf{Secondary Objective: Collect All Stars}

\quad - Each level contains three stars. To achieve a perfect score, the candy must physically touch all three stars before reaching Om Nom.

\quad - Your strategy should aim for a three-star collection.\\

\textbf{Core Mechanics \& Operations}

These are the fundamental actions you can command:

\quad - Cut the Rope: This is your primary action. Sever a rope by swiping across it. You can cut multiple ropes simultaneously.

\quad - Interact with Tools: Activate special objects by tapping them.

\textbf{The Arsenal: Game Elements \& Tools}

You must understand the function of every element to formulate a winning strategy.

- \textbf{Rope:} The basic element that suspends the candy. It can be cut.

\quad - Tension: Stretched-out ropes turn red, indicating higher tension and a stronger resulting swing or launch when cut.

- \textbf{Pin (Blue):} An anchor point for ropes.

- \textbf{Active Pin:} A pin with a dashed circle around it will automatically fire a new rope that attaches to the candy as soon as it enters the circle's range.

- \textbf{Bubble:} When attached, it makes the candy defy gravity and float upwards. Tap the bubble to pop it.

- \textbf{Air Cushion (Blue):} When tapped, it releases a directional puff of air, providing thrust to the candy.

- \textbf{Magic Hat:}  These come in pairs. When the candy enters one hat, it is instantly teleported to the other hat of the same color, conserving its entry momentum.

- \textbf{Hazards (Spikes, Electric Sparks):} Lethal obstacles. If the candy touches them, it is destroyed. Avoid them at all costs.

- \textbf{Spider:} An enemy that will climb along a rope towards the candy. You must cut the rope it is on before it reaches the candy.\\

\textbf{Strategic Principles: The Master's Mindset}

To succeed, you must think like a physicist and a `Cut the Rope` master.

- \textbf{Sequence (Order of Operations):} The order in which you cut ropes and activate tools is paramount. A wrong sequence will lead to failure. Analyze the entire system before making the first move.

- \textbf{Timing (Precision in Action):} *When* you act is as important as *what* you do. Cutting a rope at the peak of a swing maximizes horizontal distance by converting potential energy into kinetic energy. Popping a bubble at the right moment can drop the candy perfectly onto a moving platform.

- \textbf{Prediction (Physics-Based Foresight):} You must constantly predict the candy's trajectory. Before every action, mentally simulate the outcome based on the laws of physics:

\begin{itemize}
    \item \textbf{Gravity:} The constant downward pull.
    \item \textbf{Inertia \& Momentum:} An object in motion stays in motion. Use swings to build momentum.
    \item \textbf{Buoyancy:} The upward force from a bubble.
    \item \textbf{Combination (Tool Synergy):} The most complex puzzles require combining tools. A bubble might be needed to lift the candy into the jet stream of an air cushion, which then pushes it past spikes and toward the final star.
\end{itemize}

Your task is to analyze the initial state of each level and output a clear, step-by-step plan of cuts and interactions, specifying the precise timing and sequence required to collect all three stars and safely deliver the candy to Om Nom.

\end{systempromptbox}

\begin{table*}[h!]
\centering
\caption{Benchmark Cut the Rope. Model performance comparison. Prompt Format: VLA.}
\label{tab:model_performance_comparison_cutrope_vla}
\resizebox{\linewidth}{!}{
\begin{tabular}{l | lllllll}
\toprule
\textbf{Model} & \textbf{SuccRate} & \textbf{Total\goldstar} & \textbf{Mean\goldstar\ (Overall)} & \textbf{Mean\goldstar\ (Completed)} & \textbf{\# Num. 1\goldstar} & \textbf{\# Num. 2\goldstar} & \textbf{\# Num. 3\goldstar} \\
\midrule

\texttt{\textit{MOCK}}& 11.36\% & 14.00 & 0.16 & 1.40 & 3.00 & 1.00 & 3.00\\

\rowcolor{lightgray}\multicolumn{8}{c}{\textbf{Open-Source Models}} \\ \midrule

\rowcolor{mylightgray}\multicolumn{8}{c}{\textbf{Qwen2.5-VL Series}} \\
\textbf{Qwen-3B} & 7.95\% & 11.00 & 0.12 & 1.57 & 1.00 & 2.00 & 2.00\\
\textbf{Qwen-7B} & 9.09\% & 9.00 & 0.10 & 1.12 & 1.00 & 1.00 & 2.00\\
\textbf{Qwen-32B} & 6.82\% & 10.00 & 0.11 & 1.67 & 2.00 & 1.00 & 2.00\\
\textbf{Qwen-72B} & 13.64\% & 22.00 & 0.25 & 1.83 & 6.00 & 2.00 & 4.00\\

\rowcolor{lightgray}\multicolumn{8}{c}{\textbf{Closed-Source Models}} \\ \midrule

\rowcolor{mylightgray}\multicolumn{8}{c}{\textbf{Claude Series}} \\
\textbf{Claude 3.5 Sonnet} & 21.59\% & 29.00 & 0.33 & 1.53 & 3.00 & 4.00 & 6.00\\
\textbf{Claude 3.7 Sonnet} & 20.45\% & 24.00 & 0.27 & 1.33 & 4.00 & 4.00 & 4.00\\
\textbf{Claude 4.0 Sonnet} & 22.73\% \secondlogo & 29.00 & 0.33 & 1.45 & 6.00 & 4.00 & 5.00\\
\textbf{Claude 4.0 Opus} & 26.14\% \firstlogo & 33.00 & 0.38 & 1.43 & 6.00 & 6.00 & 5.00\\
\midrule

\rowcolor{mylightgray}\multicolumn{8}{c}{\textbf{Gemini Series}} \\
\textbf{Gemini-2.0-Flash} & 18.18\% & 25.00 & 0.28 & 1.56 & 4.00 & 3.00 & 5.00\\
\textbf{Gemini-2.5-Flash} & 12.50\% & 15.00 & 0.17 & 1.36 & 4.00 & 1.00 & 3.00\\
\textbf{Gemini-2.5-Pro} & 22.73\% \secondlogo & 30.00 & 0.34 & 1.50 & 4.00 & 4.00 & 6.00\\
\midrule

\rowcolor{mylightgray}\multicolumn{8}{c}{\textbf{GPT Series}} \\
\textbf{GPT-4V} & 15.91\% & 15.00 & 0.17 & 1.07 & 7.00 & 1.00 & 2.00\\
\textbf{GPT-4o-mini} & 7.95\% & 8.00 & 0.09 & 1.14 & 2.00 & 0.00 & 2.00\\
\textbf{GPT-4o} & 17.05\% & 19.00 & 0.22 & 1.27 & 5.00 & 4.00 & 2.00\\
\textbf{GPT-o3} & 18.18\% & 30.00 & 0.34 & 1.88 & 3.00 & 3.00 & 7.00\\
\textbf{GPT-o4-mini} & 17.05\% & 26.00 & 0.30 & 1.73 & 6.00 & 4.00 & 4.00\\
\midrule
\rowcolor{mylightgray}\textit{\textbf{Human}} & 41.36\% & 91.60 & 1.03 & 2.51 & 5.20 & 7.80 & 23.40 \\
\bottomrule
\end{tabular}
}
\end{table*}

\begin{userpromptbox}{\cutrepologo: User Prompt}
Please analyze the current game screen and plan a single action to ultimately deliver the candy to the green monster, Om Nom.
You will receive an annotated game screenshot, along with a more detailed text-based state description below it.\\

\textless\ initial\_image\_list \textgreater 

\textless\ annotated\_image\_path \textgreater 

\textless\ current\_image\_path \textgreater (or if has bubble \textless\ bubble\_annotated\_current\_image\_path \textgreater)\\

\textbf{Current Level State Analysis:}

\begin{itemize}
    \item \textbf{Pin:} \{len(pins)\} present, with IDs from 1 to \{len(pins)\}.
    \item \textbf{Active Pin:}  \{len(active\_pins)\} present, with IDs from 1 to \{len(active\_pins)\}.
    \item \textbf{Bubble:} \{len(bubbles)\} present, with IDs from 1 to \{len(bubbles)\}
    \item \textbf{Air Cushion:} \{len(air\_cushions)\} present, with IDs from 1 to \{len(air\_cushions)\}.
\end{itemize}

\textbf{History of Actions:}

\{history\_prompt\_text\}\\

\textbf{Task Instructions:}

Please carefully analyze the spatial relationships and physical possibilities of the game elements. Choose the single action from the list below that best helps achieve the objective.

\begin{itemize}
    \item cut\_pin(id=pin\_index) \# Effect: Cuts the rope attached to pin\_index.
    \item cut\_active\_pin(id=active\_pin\_index) \# Effect: Cuts the rope attached to active\_pin\_index.
    \item pop\_bubble(id=bubble\_index) \# Effect: Pops the bubble with the specified bubble\_index. The candy inside will lose its buoyancy and begin to fall vertically. Only works if there is candy inside the bubble.
    \item tap\_air\_cushion(id=air\_cushion\_index, times=[1, 3, 5]) \# Effect: Taps the air cushion with the specified air\_cushion\_index a number of times, to make it release a puff of air, pushing the candy in a specific direction.
    \item sleep(seconds=x) \# Effect: Waits for x seconds, allowing in-game physics to play out. Use this when you need to wait for a physical process (like a swing) to reach a specific state before executing the next action. After calling this, the system will provide a new game state and request the next action after x seconds.
    \item success() \# Effect: Declares the mission successful. Call this when you determine the candy is on an inevitable trajectory to enter Om Nom's mouth, requiring no further actions.
    \item fail() \# Effect: Declares the mission failed. Call this when you determine the candy is lost and no further actions can succeed.
\end{itemize}

\textbf{Output Format:}

Please strictly adhere to the following format for your chosen action. Do not add any extra explanations. ONLY output the action code in `[' and `]' brackets, without any additional text or comments.

Example: [ACTION\_CODE(parameters)]

\end{userpromptbox}

\subsection{Experiment Results}

The Cut the Rope environment represents what might be considered a ``last mile" challenge for agentic VLMs, exposing a confluence of their deepest limitations in perception, physical intuition, and dynamic control. As shown in Tables \ref{tab:model_performance_master_summary} \& \ref{tab:model_performance_comparison_cutrope_vla}, the performance of all tested models is dramatically lower than the human benchmark. The top-performing model, Claude 4.0 Opus, only achieves a 26.14\% success rate, trailing significantly behind the human players' 41.36\% average. The disparity is even more pronounced in the quality of solutions: the human players collected an average total of 91.6 stars, nearly triple that of the best model (33.00), demonstrating a superior ability to find optimal, multi-objective solutions.

Analysis of failure patterns reveals a critical weakness in spatiotemporal reasoning that begins at the most fundamental level of perception. The game's vibrant, cartoonish GUI, while trivial for a five-year-old child to interpret, presents a significant obstacle for even state-of-the-art models, which often fail at basic tasks like correctly counting the number of ropes in a scene. 
This highlights why our meticulous annotations are necessary to even begin evaluating reasoning. Beyond perception, the models fail to grasp the intuitive physics of the game—the kind of implicit, ``unspoken" understanding of momentum and trajectory that cannot be simply conveyed through textual prompts or game guides. Success often hinges on cutting a rope at the peak of a swing or precisely controlling the number of air cushion puffs. The models' inability to master this is evident in their frequent errors: they either resort to ``brute-force" actions immediately or become paralyzed by indecision, failing to wait for the candy to swing into an optimal position. This spatiotemporal deficit becomes even more acute in levels requiring screen transitions, where models exhibit a form of ``amnesia," unable to maintain a coherent state representation as the viewpoint shifts—a primary reason such levels were excluded from our final benchmark as they are currently beyond the reach of any VLM.

Furthermore, the quality of successful attempts underscores this limitation. The Mean\goldstar (Completed) metric shows that even when models manage to succesfully solve a level, their solutions lack finesse. The human player averages 2.51 stars per completed level, whereas the best models like GPT-o3 and Claude 4.0 Opus only manage 1.88 and 1.43 stars, respectively. This indicates that the models' successful attempts are often simplistic or ``brute-force" solutions that achieve the primary objective (feeding Om Nom) but fail to execute the elegant, precisely-timed maneuvers required to collect all three stars. They struggle with multi-objective planning under dynamic physical constraints.

In conclusion, the Cut the Rope benchmark effectively exposes the fragility of current VLM physical reasoning. By requiring the integration of nuanced perception, implicit physical intuition, and precise timing, it moves beyond static puzzles and tests the core of dynamic interaction. The models' widespread failure confirms a fundamental disconnect between recognizing physical elements and procedurally controlling them within a dynamic, time-sensitive system.

\section{Human performance}
\label{appendix:human_perf}

The reported \textit{human} performance for benchmarks Pooltool, Angry Birds, and Cut the Rope is not intended as representation of performance goal for human experts. The performance results are calculated only to serve as ballpark results for comparison purposes with agentic VLM results.

For Pooltool and Angry Birds, human results represent the average results of two non-expert co-authors playing the games. For Cut the Rope, human performance is the average result of 5 non-expert crowdsourced workers playing the game.

\end{document}